\documentclass{article}

\PassOptionsToPackage{sort,numbers}{natbib}

\usepackage[final]{neurips_2025}

\usepackage[utf8]{inputenc} %
\usepackage[T1]{fontenc}    %
\usepackage{hyperref}       %
\usepackage{url}            %
\usepackage{booktabs}       %
\usepackage{amsfonts}       %
\usepackage{nicefrac}       %
\usepackage{microtype}      %
\usepackage{xcolor}         %
\usepackage{amsmath,amssymb}
\usepackage{algpseudocode}
\usepackage{algorithm}
\usepackage{color}
\usepackage{float}
\usepackage{subcaption}
\usepackage{amsthm}
\usepackage{wrapfig}
\usepackage{graphicx}
\usepackage{siunitx}
\usepackage{comment}
\usepackage[export]{adjustbox}
\usepackage{multirow}
\usepackage{afterpage}

\DeclareMathOperator{\D}{\mathcal{D}}

\DeclareMathOperator{\E}{\mathbb{E}}
\DeclareMathOperator{\Var}{\mathrm{Var}}
\DeclareMathOperator{\Cov}{\mathrm{Cov}}
\DeclareMathOperator{\alp}{\bar{\alpha}}
\DeclareMathOperator{\al}{\tilde{\alpha}}
\DeclareMathOperator{\bet}{\bar{\beta}}

\newcommand{\smwidehat}[1]{\smash{\widehat{#1}}}

\newtheorem{proposition}{Proposition}
\newtheorem{lemma}{Lemma}
\newtheorem{corollary}{Corollary}
\newtheorem{proposition*}{Proposition}

\title{Active Measurement: Efficient Estimation at Scale}

\author{%
  Max Hamilton$^*$~~
   Jinlin Lai$^*$~~
   Wenlong Zhao~~
   Subhransu Maji$^\dagger$~~ 
   Daniel Sheldon$^\dagger$\\
   Manning College of Information \& Computer Sciences\\
University of Massachusetts, Amherst\\
\texttt{\{jmhamilton,jinlinlai,wenlongzhao,smaji,sheldon\}@cs.umass.edu}
}

\begin{document}

\maketitle

\renewcommand\thefootnote{}%
\footnotetext{$^{*\dagger}$ Denotes authors with equal contribution}
\addtocounter{footnote}{-1}

\begin{abstract}
    AI has the potential to transform scientific discovery by analyzing vast datasets with little human effort. 
    However, current workflows often do not provide the accuracy or statistical guarantees that are needed.
    We introduce \emph{active measurement}, a human-in-the-loop AI framework for scientific measurement.
    An AI model is used to predict measurements for individual units, which are then sampled for human labeling using importance sampling.
    With each new set of human labels, the AI model is improved and an unbiased Monte Carlo estimate of the total measurement is refined.
    Active measurement can provide precise estimates even with an imperfect AI model, and requires little human effort when the AI model is very accurate.
    We derive novel estimators, weighting schemes, and confidence intervals, and show that active measurement reduces estimation error compared to alternatives in several measurement tasks.
\end{abstract}

\section{Introduction}

AI offers a transformative approach to scientific discovery, empowering scientists to analyze vast datasets in ways that traditional methods cannot achieve~\cite{wang2023scientific,lang2023high, tuia2022perspectives}. 
Applications include species identification from images and audio for biodiversity monitoring~\cite{MerlinSoundID,van2018inaturalist,belotti2023longterm,kay2022caltech}, disease diagnosis in medical imaging~\cite{thirunavukarasu2023large,aggarwal2021diagnostic}, classifying galaxies in astronomy~\cite{lintott2008galaxy}, assessing crop health in agriculture~\cite{fuentes2024transformative,gao2020framework}, and myriad other applications in fields such as remote sensing, microscopy, and neuroscience.
In these applications, the typical goal is to make measurements to answer science or policy questions, which therefore must be precise. But AI models are far from perfect: they may introduce bias or have unacceptable error rates, and do not offer the statistical guarantees that scientists desire.

\afterpage{
\begin{figure}[t]
    \vspace{-0.1in}
    \includegraphics[width=\textwidth]{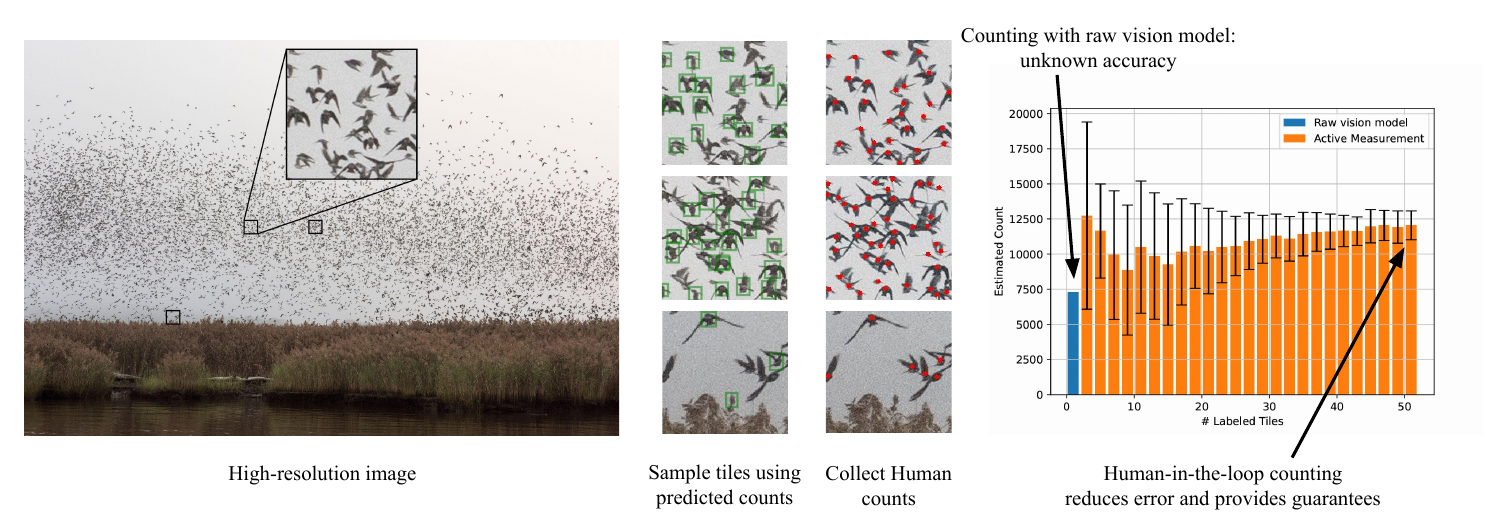}
    \vspace{-0.25in}
    \caption{\textbf{Active measurement.} An AI model predicts counts for each image tile, which are then used to form a proposal distribution to selectively sample tiles for human labeling. Ground-truth counts are used both to improve the AI model and produce a Monte Carlo estimate of the final measurement. %
    }
    \label{fig:overview}
    \vspace{-0.14in}
\end{figure}}

As a concrete example, consider counting birds in the high-resolution photograph in Fig.~\ref{fig:overview} of a large flock of Tree Swallows in Old Lyme, CT in September, 2018.
Scientists would like to estimate the population size and track it over time from photographs like this \cite[e.g.,][]{arteta2016counting,horn2008analyzing,buler2012mapping}.
Computer vision is an excellent tool for this task, but current workflows are poorly suited to scientific measurement.
A scientist might be able to estimate the count with relatively low effort using a pre-trained object detector adapted with a few labeled examples. 
However, substantial additional effort is needed to obtain a model that is accurate enough for scientific measurement and gain confidence in its performance.
A typical scenario would be to first tune and validate the detector's performance using labeled image tiles. 
To obtain bounds on count performance, more validation is needed.
After substantial effort, the scientist may find that the count is \num{20358} $\pm$ \num{8000} birds, which is not accurate enough to track populations over time.
What can be done? Under current practice, the scientist returns to model development. 
For challenging tasks, it may be impossible to achieve precise enough measurements.

We introduce a new paradigm of \emph{active measurement} where the scientist uses AI to \emph{interactively} solve the measurement task (Fig.~\ref{fig:overview}).
Active measurement is based on Monte Carlo estimation.
In each step, the scientist receives an unbiased estimate of the true count \emph{and} an estimate of its error and is asked to provide new labels.
These labels are used \emph{both} to reduce Monte Carlo error of the estimated measurement and to improve the AI model for future rounds of estimation.
The process continues until the error is sufficiently low --- in this example the estimates are 11,977 $\pm$ 1,076 after labeling 50 tiles in the image, closely aligning with the true count of 12,486 birds. 
Unlike existing workflows, arbitrarily low error is possible with enough labels even with an imperfect model.
On the other hand, if the model approaches perfect performance, very accurate estimates are possible with a small amount of labeled data.

Active measurement builds on adaptive importance sampling~\cite{owen2013monte}, active testing~\cite{kossen2021active,nguyen2018active}, and prior work on covariate and detector-based counting~\cite{meng2022count,perez2024discount}, but extends each of these works in different directions.
We introduce the active measurement framework, derive new estimators, and show that they are unbiased and consistent.
We contribute novel approaches for confidence interval construction that are tailored to the active measurement setting, and derive novel weighting schemes to combine estimates obtained in each step to minimize the overall estimation variance.
We show empirically that our techniques provide accurate confidence intervals and reduce estimation error compared to prior methods on several scientific measurement tasks. 

Code for this paper is available at: \url{https://github.com/cvl-umass/active-measurement}.
\vspace{-0.04in}
\section{Active measurement}
\vspace{-0.04in}
We consider the following scientific measurement task. 
There is a set $\Omega$ of $N=|\Omega|$ individual units (e.g., image tiles), with unknown ground-truth measurement $f(s) \geq 0$ for each $s \in \Omega$. 
For $S \subseteq \Omega$, define $F(S) = \sum_{s\in S}f(s)$.
We seek to estimate the total measurement $F(\Omega)$ across all units.

Active measurement combines human labeling with AI predictions.
Suppose that, at time $t$, the scientist has labeled units $\D_t\subseteq \Omega$ and therefore knows the ground-truth value $F(\D_t)=\sum_{s\in \D_t}f(s)$.
Further, suppose an AI model is available (e.g., trained on $\D_t$) to predict $g(s) \approx f(s)$ for other units. 
Let $q_t$ be the probability distribution proportional to $g$ on $\Omega\backslash \D_t$.
The base estimator for active measurement is:
\begin{equation}
    \label{eq:base_estimator}
\hat F_t = F(\D_t) + \frac{f(s_t)}{q_t(s_t)}, \quad s_t \sim q_t.
\end{equation}
The second term is an importance sampling estimate satisfying $\E_{q_t}[f(s_t)/q_t(s_t)] = F(\Omega\backslash \D_t)$, from which it follows that $\E[\hat F_t] = F(\Omega)$.
This estimator is inspired by active testing~\cite{kossen2021active}, which we discuss more below.
It is unbiased for any proposal distribution $q_t$, but has minimum variance (of zero) when $q_t \propto f$, which motivates the choice $q_t \propto g \approx f$.

\begin{figure}[t]
    \centering
    \vspace{-0.1in}
    \begin{minipage}[t]{0.515\textwidth}
        \begin{algorithm}[H]
            \small
        \caption{Active measurement}
        \label{alg:active_measurement}
		\begin{algorithmic}[1]
            \Require Initially labeled units $\D_1\subseteq \Omega$, acquisition distribution $q_1$, weight sequences $\alpha_\tau$, $\beta_\tau$
			\For{$t= 1,2,...,T$} 
            \State Sample $s_{t}\sim q_t(\cdot)$ and obtain $f(s_t)$
			\State Form IS estimate \begin{align}
			    \hat{F}_t=F(\D_{t})+\frac{f(s_t)}{q_t(s_t)}
			\end{align}
			\State Combine estimates as $\hat{F}_{1:t}=\sum_{\tau=1}^t\alp_\tau \hat{F}_\tau$
            \State Get variance estimates $\{\smwidehat{\Var}_\tau\}_{\tau=1}^t$ using Alg. \ref{alg:variance_update}
            \State Combine variances as $\widehat{\Var}_{1:t}=\sum_{\tau=1}^t\alp_\tau^2 \smwidehat{\Var}_\tau$
			\State Update $\D_{t+1}=\D_{t}\cup \{s_t\}$
                \State Update acquisition distribution $q_{t+1}$ over $\Omega\backslash \D_{t+1}$, e.g., by updating an AI model using $\D_{t+1}$
			\EndFor 
		\end{algorithmic}
\end{algorithm}
    \end{minipage}
    \hfill
    \begin{minipage}[t]{0.46\textwidth}
        \begin{algorithm}[H]
            \small
            \caption{Variance estimation} 
            \label{alg:variance_update}
            \begin{algorithmic}[1]
                \Require All variables from Alg. 1
                \For{$\tau= 1,2,...,t$}
                \State Get mean estimate $\hat{G}_{\tau,t}=\hat{F}_{1:t-1}-F(\D_\tau)$
                \For{$r=\tau,\ldots,t$}
                \State Form single variance estimator \begin{align}   \smwidehat{\Var}_{\tau,r}=&\sum_{s\in\D_{r}\backslash\D_{\tau}}q_{\tau}(s)\left(\frac{f(s)}{q_\tau(s)}-\hat{G}_{\tau,t}\right)^2\notag\\
                &+\frac{q_\tau(s_{r})}{q_r(s_{r})} \left(\frac{f(s_{r})}{q_\tau(s_{r})}-\hat{G}_{\tau,t}\right)^2
                \end{align}
                \EndFor 
                \State $\smwidehat{\Var}_\tau=\sum_{r=\tau}^t\bet_r\smwidehat{\Var}_{\tau,r}$
                \EndFor
            \end{algorithmic}
        \end{algorithm}
    \end{minipage}
    \vspace{-0.1in}
\end{figure}

The active measurement framework, shown in Alg. \ref{alg:active_measurement}, uses this estimator in each step of a human-in-the-loop process.
The set $\D_1$ contains initially labeled units, which may be empty. At the start of step $t$, units in $\D_{t}$ have been labeled, and an \textit{acquisition distribution} $q_t$ is used to sample a new unit for labeling. 
Our typical acquisition strategy is to train an AI model on $\D_t$ to generate predictions $g(s)$ for all units $s \in \Omega\backslash \D_t$ and then choose $q_t \propto g$.
In Line 3, the newly labeled unit is used to form an importance sampling estimate with $\E[\hat{F}_t]=F(\Omega)$ using Eq.~\eqref{eq:base_estimator}.
Line 4 combines the estimators from each step into the estimator $\hat{F}_{1:t}$ with normalized weights that satisfy $\sum_{\tau=1}^t\alp_\tau=1$, which incorporates all of the labeled samples and represents the model's overall estimate at time $t$. 
Hereafter, for any weighting scheme, we use $\alpha_\tau$ to denote a sequence of unnormalized weights for $1 \leq \tau \leq N$ and $\alp_\tau$ for their normalized counterparts $\alp_\tau=\alpha_\tau/\sum_{r=1}^t\alpha_r$, with $t$ implicit from context.
\begin{proposition}
    The combined estimator $\hat{F}_{1:t}=\sum_{\tau=1}^t\alp_\tau \hat{F}_\tau$ is unbiased: $\E[\hat{F}_{1:t}]=F(\Omega)$. 
\end{proposition}
At step $t$, we also generate a variance estimate from Alg.~\ref{alg:variance_update}, which is derived in \S~\ref{sec:variance}.
Alg.~\ref{alg:variance_update} takes $\mathcal O(t^2)$ time naively, but can be improved to $\mathcal{O}(t)$ using the streaming algorithm in \S~\ref{sec:streaming}, so that step $t$ of active measurement takes $\mathcal O(t)$ time overall.
In practical settings we expect the time to be dominated by labeling and updating an AI model. 
We next discuss related work before returning to weighting schemes and variance estimation.

\section{Related work}
\label{sec:related_work}
Active measurement is closely related to three existing lines of work. Active testing~\cite{FarquharGR21,kossen2021active} interactively estimates the test loss of an AI model by sampling unlabeled points according to predicted losses from a surrogate model and then using the newly acquired labels to form an importance-sampling based estimator of the loss \emph{and} update the surrogate model.
The sampling strategy and form of the estimator in Eq.~\eqref{eq:base_estimator} is identical to that used by active testing, with the significant difference that the estimand is different: active measurement seeks to directly estimate a scientific measurement while iteratively refining the AI model itself, while active testing estimates a test loss and refines a surrogate model. Compared to active testing, we also contribute novel weighting schemes (\S~\ref{sec:weighting}) and show that these reduce error, and novel approaches for variance estimation and confidence interval construction (\S~\ref{sec:variance}), which was not considered in active testing. 

Active testing and active measurement are both versions of adaptive importance sampling (AIS)~\cite{oh1992adaptive, owen2013monte, bugallo2017adaptive}. AIS improves an importance-sampling estimator $f(s)/q(s)$ by iteratively refining the proposal $q$ and forming a weighted combination of the estimators made with different proposal distributions. Active measurement modifies AIS to sample \emph{without} replacement from a finite sample space, which leads to the estimator in Eq.~\eqref{eq:base_estimator}, compared to the usual importance-sampling estimator supported on all of $\Omega$. Sampling without replacement leads to novel considerations in the selection of weights (\S~\ref{sec:weighting}) and in variance estimation (\S~\ref{sec:variance}).
The DISCount (detector-based importance sampling) method of \citet{perez2024discount}, which builds on~\cite{meng2022count}, estimates the total counts in scientific collections using AI model predictions for the importance sampling proposal distribution.
Active measurement shares the same goal, and advances on DISCount by interactively refining the AI model with acquired labels and by sampling without replacement, which both can significantly reduce estimation error.

The concept of combining an AI model with limited human effort is related to semi-supervised learning~\citep{4787647}, where the model is trained with using a combination of labeled and unlabeled data. Most recently, prediction-powered inference (PPI)~\citep{angelopoulos2023prediction} was proposed to give valid statistical inference using a small amount of labeled data. 
Unlike PPI, which assumes data is iid and seeks to estimate a population parameter, active measurement samples data interactively and non-uniformly as aided by AI and seeks to estimate a scientific measurement on a finite dataset. We show in the experiments that active measurement outperforms a PPI-motivated baseline.
In a similar spirit, active machine learning has been used to construct large datasets with crowd-sourcing~\citep{patterson2015tropel,kovashka2016crowdsourcing,Walmsley_2021}. We have a different objective of achieving accurate estimation with a small amount of labeled data. 

\section{Weighting schemes} \label{sec:weighting}

What is an appropriate weight sequence $\alpha_\tau$ to combine the estimators in active measurement?
An important observation is that the estimators in each step, though not independent, are uncorrelated: 
\begin{proposition}
    For any $1\le\tau<r\le t$, $\Cov(\hat{F}_\tau,\hat{F}_r)=0$.
\end{proposition}
Thus, the combined variance is $\Var[\hat F_{1:t}] = \sum_{\tau=1}^t \bar \alpha_\tau^2 \Var[\hat F_\tau]$, and the weighting that achieves minimum variance uses inverse-variance weighting, i.e., $\alpha_\tau=1/\Var[\hat{F}_\tau]$. 
However, in practice we don't know $\Var[\hat{F}_\tau]$ and it is difficult to estimate (see \S~\ref{sec:variance}), so we first consider fixed weighting approaches that approximate this principle under assumptions about how $\Var[\hat F_\tau]$ changes due to: (1) adapting the model, and (2) the shrinking sample space.

\textbf{Square root law. } In AIS, a simple and near-optimal alternative is the square root law $\alpha_\tau^{\mathrm{SQRT}}= \sqrt{\tau}$~\citep{owen2020square}. To derive this law, it is assumed that variance reduces at the rate $\Var[\hat{F}_\tau]\propto \tau^{-y}$ due to adaption of the proposal distribution, but the rate $y\in [0,1]$ is unknown. 
The square root law works as a conservative strategy that is within a factor of $9/8$ of the optimal variance even without knowing the rate $y$: if the variance with an optimal weighting strategy for rate $y$ at step $t$ is $\Var_\mathrm{opt}(y,t)$ and $\hat F_{1:t}^\mathrm{SQRT}$ is the estimator using weights $\alpha_\tau^{\mathrm{SQRT}}$ (we will use similar notation for other weighting schemes below), then $\sup_{t\ge 1}\sup_{y\in[0,1]}\frac{\Var[\hat{F}^{\mathrm{SQRT}}_{1:t}]}{\Var_\mathrm{opt}(y,t)}\le \frac{9}{8}$.
However, in our context, the model $\Var[\hat F_\tau] \propto \tau^{-y}$ does not account for shrinking variance due to sampling without replacement, which is not optimal.

\textbf{LURE weights. } In active testing, a weighting scheme known as LURE was introduced for the pool-based setting with sampling without replacement~\citep{FarquharGR21}. If the number of units is $|\Omega|=N$, the LURE weights are $\alpha_\tau^{\mathrm{LURE}}=\frac{1}{(N-\tau)(N-\tau+1)}$. The original motivation was to treat samples equally in the combined estimate, but we can reinterpret these weights as accounting for variance reduction due to the shrinking sample space: 

\begin{proposition}
\label{prop:var}
    If there exist constants $0<A\le B$ such that for any $s\in \Omega$, $A\le f(s)\le B$ and $A\le g(s)\le B$, then there exists a constant $C>0$ such that $\Var [\hat{F}_\tau]\le C(N-\tau)(N-\tau+1)$. 
\end{proposition} 

For brevity, define $w_\tau=\frac{1}{(N-\tau)(N-\tau+1)}$. This proposition indicates that the variance is order $1/w_\tau$ when only considering the sample space reduction, so the LURE weights of $w_\tau$ are well justified by inverse variance weighting. 
However, in our setting, LURE weights neglect variance reduction due to adaptation of the AI model, which is also not optimal.

\begin{wrapfigure}{r}{0.30\textwidth}
  \vspace{-15pt}
  \begin{center}
    \includegraphics[width=0.25\textwidth]{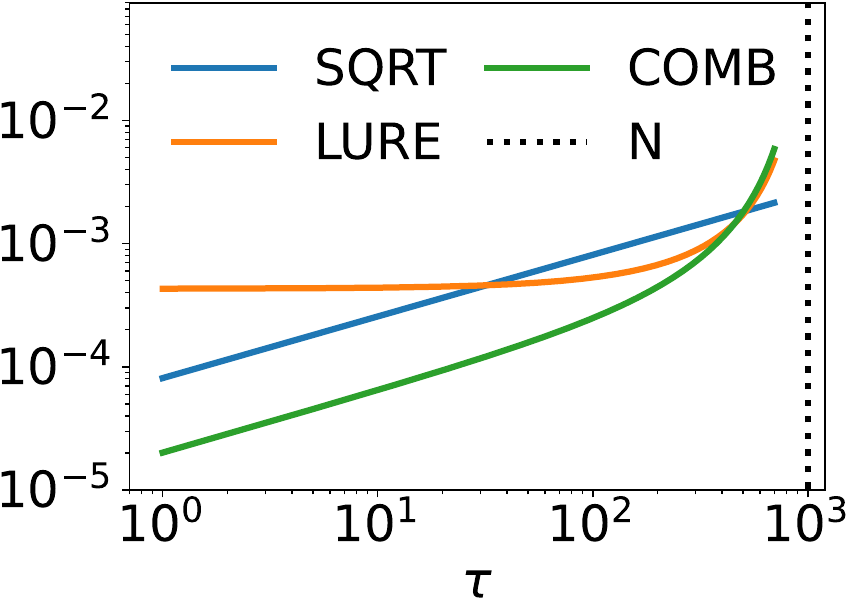}
  \end{center}
  \caption{Normalized weights as functions of $\tau$ for $t=700$ and $N=1000$. }
  \vspace{-10pt}
  \label{fig:weights}
\end{wrapfigure}

\textbf{Combined weights. } 
We propose to combine the above two weighting schemes and use $\alpha_\tau^{\mathrm{COMB}}= w_\tau\sqrt{\tau}$. Fig. \ref{fig:weights} demonstrates the growth of different weighting schemes. When $\tau$ is small, the detector has not been well-tuned so weighting schemes that discourage early estimators, like the square root law, are preferred. As $\tau$ becomes closer to $N$, the sample space reduction becomes more significant, which is addressed by the LURE weights. Our combined weights unify the best of the two, accounting for the two sources of variance reduction. 
The following result bounds our worst-case estimation error under our variance reduction model, in a way similar to the square root law for AIS.

\begin{proposition}
    If $\Var[\hat{F}_\tau]\propto \tau^{-y}/w_\tau$, where $y\in[0,1]$ and $w_\tau$ are non-decreasing, then the estimate $\hat{F}_{1:t}^{\mathrm{COMB}}=\sum_{\tau=1}^t\alp_\tau^{\mathrm{COMB}}\hat{F}_\tau$, where $\alpha_\tau^{\mathrm{COMB}}= w_\tau\sqrt{\tau}$, satisfies 
    \vspace{-0.05in}
    \begin{align}
        \sup_{t\ge 1}\sup_{y\in[0,1]}\frac{\Var[\hat{F}^{\mathrm{COMB}}_{1:t}]}{\Var_\mathrm{opt}(y,t)}\le \frac{9}{8},
    \end{align}
    where $\Var_\mathrm{opt}(y,t)$ is the estimation variance when the estimators are weighted proportionally to the inverse of the ground-truth variances.
\end{proposition}
In \S \ref{exp:weighting}, we show empirically that the combined weighting scheme produces lower estimation error than either of the two components alone. 

\textbf{Inverse variance weighting. } In AIS, it has been advised not to approximate inverse variance weighting using \emph{estimated} variances~\cite{owen2020square}, due to the unreliability of variance estimates with only a finite number of samples in each round of AIS and the potentially large bias this could introduce.
In the next section, we introduce consistent estimates of (conditional) variances and show that these can make inverse variance weighting practical and achieve even better estimation error than combined weighting scheme in some settings.

\section{Variance estimation and confidence intervals} \label{sec:variance}

Along with an unbiased and consistent estimator $\hat F_{1:t}$ we typically want a consistent estimate of $\Var[\hat F_{1:t}]$ to understand the error and construct confidence intervals. However, this is complicated by the fact that the estimators $\hat F_\tau$ are not iid. Rather, each depends on the samples drawn in the previous step, and we only have a \emph{single} random draw for each $\tau$.
As a result, we will be able to form an unbiased but not a consistent estimate of $\Var[\hat F_{1:t}]$, unlike in the iid case (but similar to AIS~\cite{zhou1998adaptive}).

\textbf{Martingale convergence.}
Fortunately, we can turn to martingale arguments to instead estimate a conditional variance that is appropriate for confidence intervals. 
For this argument, consider the centered sequence of estimators $\tilde F_{1:t} = \sum_{\tau=1}^t \alpha_\tau (\hat F_\tau - F(\Omega))$ where the weights are unnormalized.
Because each $\hat F_t$ is unbiased, the displacement $\tilde F_{1:t} - \tilde F_{1:t-1} = \alpha_t (\hat F_t - F(\Omega))$ has zero mean, making the sequence $\tilde F_{1:t}$ a martingale. 
A martingale central limit theorem can be used to show that averaging over enough of these zero-mean displacements gives a limiting behavior similar to iid averaging. 
Stated informally, we have the following:
\begin{proposition}[informal]
\label{prop:martingale_clt}
Let $V^2_{1:t} = \sum_{\tau=1}^t \bar\alpha_\tau^2 \Var[F_\tau | \D_\tau]$ be the conditional variance of $\hat F_{1:t}$. Under suitable regularity conditions, $\frac{\hat F_{1:t} - F(\Omega)}{V_{1:t}} \stackrel{D}{\to} \mathcal N(0, 1)$.
\end{proposition}
We provide a formal statement and proof in \S~\ref{sec:martingale_clt}, which requires constructing a triangular array and letting the number of active measurement steps $T$ and domain size $N$ go to infinity jointly. 
See~\cite{zhou1998adaptive} and~\cite{portier2018asymptotic} for similar results for AIS, i.e., sampling \emph{with} replacement.
This result motivates forming estimators of the \emph{conditional} variance for use in confidence intervals. 

\textbf{Novel conditional variance estimators.}
We first focus on the conditional variance of a single estimator $\hat F_\tau$, which is the variance of the importance weights $f(s)/q_\tau(s)$ where $s \sim q_\tau$:
\begin{align}
    \label{eq:var}
    \Var[\hat{F}_\tau|\D_\tau]=\sum_{s\in\Omega\backslash \D_{\tau}}q_\tau(s)\left(\frac{f(s)}{q_\tau(s)}-F(\Omega\backslash \D_{\tau})\right)^2
\end{align}
In importance sampling, we typically estimate this variance over iid samples from $q_\tau$, but in active measurement we have only a single sample $s_\tau \sim q_\tau$.
We therefore derive a novel variance estimator that uses samples $s_\tau, s_{\tau+1},\ldots s_t$ to estimate $\Var[\hat{F}_\tau|\D_\tau]$ following an importance sampling approach similar to that of the base estimator in Eq.~\eqref{eq:base_estimator}. For $r\ge\tau$, let
\begin{align}
\label{eq:var_estimate}
    \smwidehat{\Var}_{\tau,r}=&\sum_{s\in\D_{r}\backslash\D_{\tau}}q_{\tau}(s)\left(\frac{f(s)}{q_\tau(s)}-F(\Omega\backslash \D_{\tau})\right)^2+\frac{q_\tau(s_{r})}{q_r(s_{r})} \left(\frac{f(s_{r})}{q_\tau(s_{r})}-F(\Omega\backslash \D_{\tau})\right)^2.
\end{align}
The first term of Eq.~\ref{eq:var_estimate} is the exact sum of the terms from Eq.~\ref{eq:var} for $s \in \{s_\tau, \ldots, s_{r-1}\}$, and the second term is an importance sampling estimate using $s_r \sim q_r$ for the sum of the remaining terms. 
We then mix the variance estimates $\smwidehat{\Var}_{\tau,r}$ for each $r$ to get the combined estimator $\smwidehat{\Var}_{\tau}:=\sum_{r=\tau}^t\bet_{r}\smwidehat{\Var}_{\tau,r}$, where $\beta_r$ is another sequence of weights; for simplicity, we will always use the LURE weights $\beta_{r}^{\mathrm{LURE}}=w_r$ to model the shrinking sample space.

\begin{proposition}
\label{prop:var_unbiased}
    The estimators $\smwidehat \Var_\tau$ and $\smwidehat \Var_{\tau, r}$ for $\tau \leq r \leq t$ satisfy
    $\E[\smwidehat \Var_{\tau} | \D_\tau] = \E[\widehat \Var_{\tau, r} | \D_\tau] = \Var[\hat{F}_\tau | \D_\tau]$ and $\E[\widehat \Var_{\tau}] = \E[\smwidehat \Var_{\tau, r}] = \Var[\hat{F}_\tau]$.
\end{proposition}
Further, beacuse it averages over many individual estimates, the error of $\smwidehat{\Var}_{\tau}$ for estimating the conditional variance converges as $t \to N$:
\begin{proposition}
\label{prop:LURE_var}
    With the same settings as Prop. \ref{prop:var}, the variance estimate for $\hat{F}_\tau$ weighted by the LURE scheme $\smwidehat{\Var}_{\tau}^{\mathrm{LURE}}=\sum_{r=\tau}^t\bet_{r}^{\mathrm{LURE}}\smwidehat{\Var}_{\tau,r}$ satisfies that $\Var\left[\smwidehat{\Var}_{\tau}^{\mathrm{LURE}}|\D_\tau\right]\lesssim \frac{1}{t-\tau+1}\cdot\min\big(1,\frac{(N-t)^2}{t-\tau+1}\big)$.
\end{proposition}
The notation $\lesssim$ hides multiplicative constants with respect to $t$. 
When $t \ll N$, we have the usual Monte Carlo rate of $(t-\tau+1)^{-1}$ (recall that $t - \tau + 1$ is the number of samples). On the other hand, when $t \to N$, a faster rate 
is achieved thanks to sampling without replacement. 
This result contrasts with the usual practice of variance estimation for AIS, where only the samples from one stage are used to estimate the conditional variance, and the stagewise estimators do not converge.

\textbf{Confidence intervals.}
Several practical steps remain to construct confidence intervals.
The full variance estimation procedure is shown include Alg. \ref{alg:variance_update}. 
Previously, we assumed knowledge of $F(\Omega\backslash \D_{\tau})$ for estimating the conditional variance, but in practice we use the plug-in estimator $\hat{G}_{\tau,t}=\hat{F}_{1:t-1}-F(\D_\tau)\approx F(\Omega\backslash \D_{\tau})$ from step $t$ of active measurement. 
The full conditional variance estimator is then
$\smwidehat{\Var}_{1:t}=\sum_{\tau=1}^t\alp_\tau^2\smwidehat{\Var}_{\tau} \approx \sum_{\tau=1}^t \alp_\tau^2 \Var[F_\tau | \D_\tau] = V^2_{1:t}$, which according to Prop.~\ref{prop:martingale_clt} is the appropriate quantity for the martingale central limit theorem.
The naive implementation to compute $\smwidehat{\Var}_{1:t}$ using Alg.~\ref{alg:variance_update} takes $\mathcal{O}(t^2)$ time but we can reduce the complexity to $\mathcal{O}(t)$ with a streaming algorithm; see \S \ref{sec:streaming} for details.
Finally, a $1-\alpha$ confidence interval is constructed as $\hat{F}_{1:t} \pm z_{\alpha/2}\smwidehat{\Var}_{1:t}^{1/2}$, where $z_{\alpha/2}$ is the $1-\alpha/2$ quantile of the standard normal distribution.

We will also consider confidence intervals formed using the ``simple'' conditional variance estimator $\smwidehat{\Var}_{1:t}^{\mathrm{simp}}=\sum_{\tau=1}^t\alp_\tau^2(\hat{F}_\tau-\hat F_{1:t})^2$, which is the motivated by the fact the sum of squared deviations $\sum_{\tau=1}^t \bar\alpha_\tau^2 (\hat F_\tau - F(\Omega))^2$ of the martingale sequence converges to the conditional variance $V^2_{1:t}$ under the same regularity conditions as Prop.~\ref{prop:martingale_clt}; see \S~\ref{sec:martingale_clt} for details.
When needed, we will denote the estimator $\smwidehat \Var_{1:t}$ from Alg.~\ref{alg:active_measurement} as $\smwidehat{\Var}_{1:t}^{\mathrm{cond}}$ to emphasize that it is based on estimating conditional variances for each $\tau$.

\textbf{Conditional inverse variance weighting.}
Prior AIS literature has advised against using inverse variance weighting with estimated variances~\cite{owen2020square}.
However, our analysis suggests that weighting with estimated \emph{conditional} variances may be appropriate. 
First, Prop.~\ref{prop:martingale_clt} can be read as $\hat F_{1:t} \approx \mathcal N(F(\Omega), V^2_{1:t})$, which motivates choosing weight sequences $\alpha_\tau$ to minimize the \emph{conditional} variance $V^2_{1:t} = \sum_{\tau=1}^t \bar\alpha_\tau^2 \Var[\hat F_\tau | \D_\tau]$, the optimal choice being $\alpha_\tau \propto 1/\Var[\hat F_\tau | D_\tau]$.
Second, Prop.~\ref{prop:LURE_var} controls the error of the conditional variance estimators $\widehat\Var_\tau \approx \Var[F_\tau | D_\tau]$. 
Because the error decays with the number of samples $t- \tau + 1$, estimates for $\tau \ll t$ will be more reliable than those for $\tau \approx t$.
This motivates our proposed scheme, which uses inverse estimated variances for $\tau \leq \gamma t$ and continues the sequence using the $\alpha_\tau^\mathrm{COMB}$ weights for $\tau > \gamma t$, where $\gamma \in (0, 1)$ is a hyperparameter:
\begin{align}
    \alpha_\tau^{\mathrm{INV}}=\begin{cases}
    1/\smwidehat{\Var}_{\tau}&1\le \tau\le \gamma t\\ \alpha^{\mathrm{COMB}}_\tau \cdot \frac{1/\smwidehat{\Var}_{\gamma t}}{\alpha^{\mathrm{COMB}}_{\gamma t}}&\gamma t\le \tau\le t\end{cases}.
\end{align} 
The constant factor $(1/\smwidehat{\Var}_{\gamma t})/(\alpha^{\mathrm{COMB}}_{\gamma t})$ for $\tau \geq \gamma t$ ensures that the weight sequence is continuous at $\tau = \gamma t$.
We show in our experiments that these weights can produce even better estimation results than the $\alpha^{\mathrm{COMB}}$ weights for suitable settings of $\gamma$.
Disadvantages are that $\hat F_{1:t}$ is no longer unbiased when using estimated weights, and that confidence intervals may have poorer coverage.

\section{Experiments}\label{sec:experiments}
We mainly experiment with two tasks: 1) counting birds in high-resolution images of Tree Swallow flocks, and 2) counting roosting birds in weather radar images across multiple radar stations and years.
The first task is analogous to common counting problems in microscopy or medical imaging, where an object detector trained on a few examples may perform reasonably well. In contrast, the second task is considerably more challenging, as it requires a custom detector---typically the result of substantial community effort involving dataset annotation, model training, and validation. To demonstrate domain generality, we also performed experiments on malaria-infected cell counting and damaged-building counting from satellite images. For all IS-based methods, we use proposal distributions proportional to predicted counts using an object detector.

\textbf{Counting birds in high-resolution images.}
We aim to estimate the total number of birds in two separate images---``sky" and ``reeds"---each with different difficulty levels. The reeds image is more challenging due to its higher bird density and more complex background (see Fig.~\ref{fig:overview} and Fig.~\ref{fig:sky-reeds-images}). We divide the sky and reeds images into tiles of size 200$\times$200 and 160$\times$160 pixels, respectively, and manually annotate the birds in each tile using the VGG annotator~\cite{dutta2016via}. This results in 925 tiles for sky and 1,426 tiles for reeds, which serve as the annotation units in our experiments. In total, the ground truth bird count is 5,682 for the sky image and 12,486 for the reeds image.

To detect birds, we train a Faster R-CNN \cite{ren2016fasterrcnnrealtimeobject} detector with a ResNet-50 \cite{he2015deepresiduallearningimage} backbone pre-trained on ImageNet \cite{5206848}, using the Detectron2 \cite{wu2019detectron2} library. The model is fine-tuned with a single A16 GPU for 400 iterations and a learning rate of 0.001 on the annotated tiles. The detector performs reasonably well, with average error rates of 9.5\% and 33.1\% when trained on 50 randomly selected tiles. We also experimented with few-shot detection models~\cite{countingdetr2022,ranjan2021learning} that do not require training, but found that a standard Faster R-CNN detector outperforms them when even a few labeled tiles are available. See \S~\ref{sec:fsmodels} for a comparison.

\paragraph{Counting roosting birds in radar.}
Another scientific application involves estimating bird counts from weather radar. Birds often congregate overnight in large numbers. Their mass departure from roosting sites in the morning can be detected by weather radar and the roosts leave visible signatures in radar-collected data channels such as reflectivity (Fig.~\ref{fig:roosts})~\cite{deng2023quantifying,belotti2023longterm}. Scientists can use these signals to estimate total bird counts, providing valuable data to analyze long-term migration trends.

We adapt the experimental setup of DISCount~\cite{perez2024discount}, leveraging expert annotated roosts from the Great Lakes analysis in~\cite{belotti2023longterm} and~\cite{deng2023quantifying} as ground truth. Our objective is to automatically estimate the total of the daily bird counts for each of 11 radar stations in the Great Lakes region, for dates between June 1st and October 31st over the five-year period from 2015 to 2019.

We follow \citet{Perez2022.10.28.513761} to pretrain a roost detector on a manually labeled training set from multiple radar stations that do not include the Great Lakes stations. It is based on a Faster R-CNN architecture with a ResNet-101 backbone and an adapter layer to handle radar channels across elevations and time steps. Detections from consecutive timestamps are assembled into tracks and bird counts are estimated from the tracks based on the radar geometry and reflectivity of the birds. We provide details in \S~\ref{appendix:roosts}. The detection task is challenging and the best model only achieves 56\% mean average precision. Thus, \cite{belotti2023longterm, deng2023quantifying} invested substantial manual screening time for scientific analyses that require high precision. Our method substantially reduces this screening effort in estimating the bird counts. 

We adapt the detector to each station with learning rate $10^{-4}$ for 3000 iterations on a single A16 GPU. To reduce overfitting, we use a mix of 80\% pretraining data and 20\% station-specific labeled data. %
The model is finetuned every 10 samples for the first 40 samples, after which performance saturates.

\paragraph{Counting malaria-infected cells.} The Malaria Cell dataset (image set BBBC041v1, available from the Broad Bioimage Benchmark Collection~\citep{ljosa2012annotated}) comprises 1,364 images (about 80,000 cells). This dataset contains microscopy images of blood smears used for detecting and classifying malaria parasites, aimed at enabling the development and evaluation of automated methods for parasite detection, quantification, and stage classification. For this evaluation, we focus on counting the number of infected cells, which are around 5\% of the dataset. We use the same settings as our sky and reeds image experiments. For our initial model we finetune the default Faster R-CNN network on three randomly selected cell images.

\paragraph{Counting damaged buildings.} For damaged building detection we focus on the Palu Tsunami subset of xBD~\citep{gupta2019creating}, which contains 113 satellite images of the shoreline before and after the Palu Tsunami. We count the number of damaged buildings, which corresponds to a label of “major-damage” or “destroyed”. Only the post disaster images are given to the model. We use the same hyperparameters as before. The initial model is trained on 5 randomly selected images.

\paragraph{Baselines.} We compare our work against DISCount \cite{perez2024discount}, which we will denote DIS. We also investigate the impact of the two major extensions individually: sampling without replacement (DIS+WOR) and fine-tuning the detector similar to adaptive importance sampling (DIS+AIS). Active measurement combines all of these, so we denote it as DIS+AIS+WOR. We also compare against simple Monte Carlo (MC),
i.e., DIS with uniform $q$~\cite{perez2024discount}, and MC without replacement (MC+WOR).
Lastly we show the performance of the raw detector predicted counts after each step of adaptation.  

Since averaging over many trials is computationally expensive---particularly when detector fine-tuning (+AIS) is involved---we approximate the effect of interactive model adaptation using a ``fixed checkpoint'' approach that uses a fixed sequence of detectors pre-trained on an increasing number of labels sampled according to a fixed strategy.
Specifically, we sample annotation units uniformly from those with non-zero counts for different sizes.
These fixed model checkpoints are then reused across all trials, allowing us to run a significantly larger number of evaluations. In a real-world deployment, a practitioner would typically fine-tune the detector using the same samples on which annotations are collected during active measurement. However, we find that the estimates produced using this approximation match those obtained via a full end-to-end setup, as discussed in \S~\ref{sec:endtoend}.

\paragraph{Evaluation metrics.} 
We evaluate our methods using fractional error and CI coverage. Specifically, we estimate them using the \emph{fractional error} $1/M\sum_{m=1}^M|\hat{F}_{1:t}^{(m)}-F(\Omega)|/F(\Omega)$ and \emph{coverage} $1/M\sum_{m=1}^M\mathbb{I}\big[|\hat{F}_{1:t}^{(m)}-F(\Omega)|\le z_{\alpha/2}\big(\smwidehat{\Var}_{1:t}^{(m)}\big)^{1/2}\big]$, 
where $M$ is the number of trials, $F(\Omega)$ is the ground truth measurement, and ($\hat{F}_{1:t}^{(m)}$, $\smwidehat{\Var}_{1:t}^{(m)}$) are the estimates from the $m$th trial.

\section{Results}
\begin{figure}
    \centering
    \includegraphics[width=\linewidth]{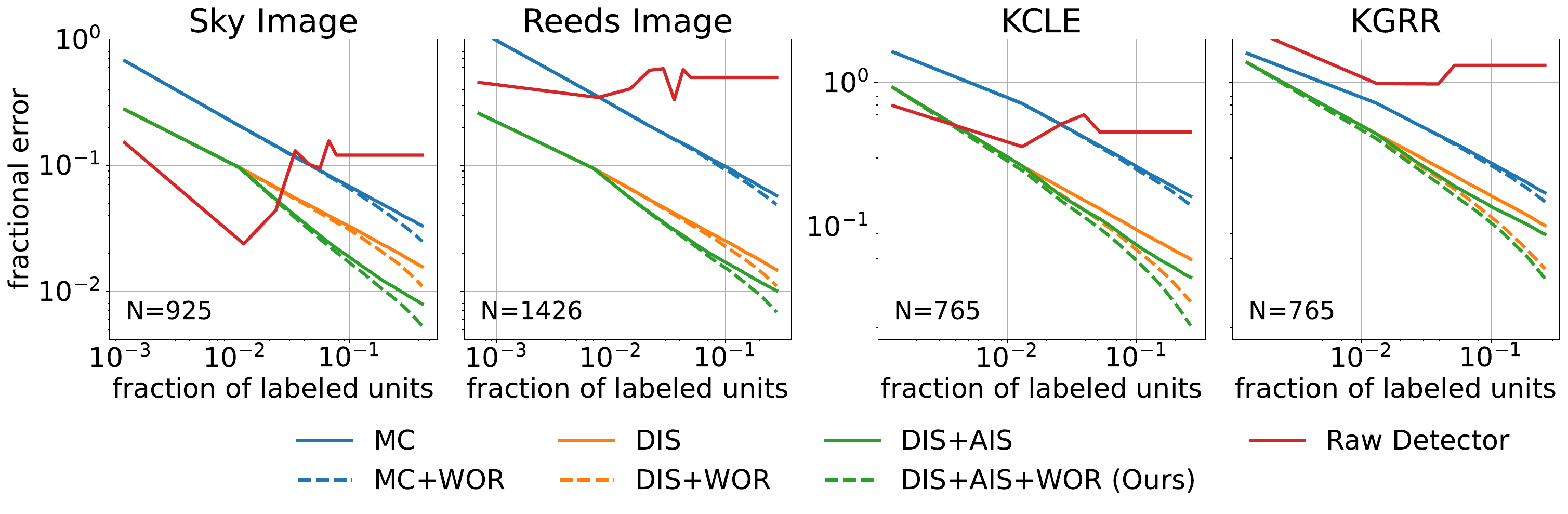}
    \begin{minipage}[b]{0.73\textwidth}
        \includegraphics[width=\linewidth,valign=m]{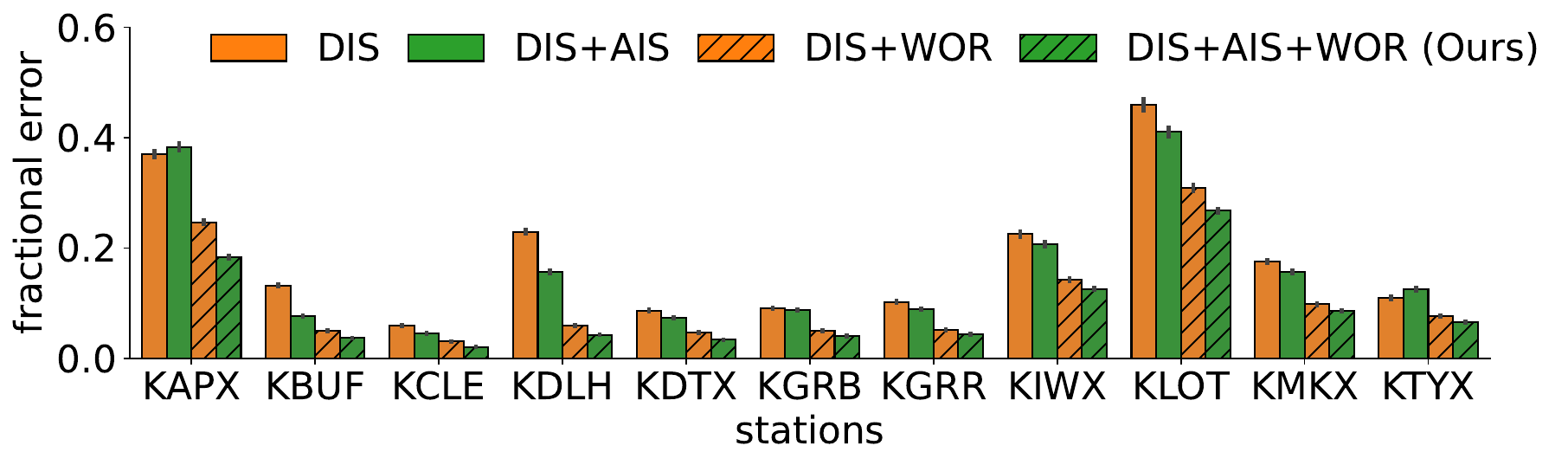}
    \end{minipage}\hfill
    \begin{minipage}[b]{0.25\textwidth}
        \centering
        \footnotesize
        \setlength{\tabcolsep}{2pt}
        \begin{tabular}{cc}
        \multicolumn{2}{c}{\textbf{Radar Task}} \\
         \textbf{Method} & \textbf{Frac. Err.} \\
         \hline
            Raw Det, \text{t=0~} & 3.78\\ 
            Raw Det, t=10 & 1.68\\ 
            Raw Det, t=20 & 1.74\\ 
            Raw Det, t=30 & 2.41\\ 
            Raw Det, t=40 & 2.79\\ 
            \hline 
            Act Mea, t=40 & 0.23 \\
            Act Mea, t=200 & 0.06 \\
        \vspace{0.1in}
    \end{tabular}
    \end{minipage}
    \vspace{-0.25in}
        \caption{\textbf{Estimation error on two measurement tasks.} Top: Fractional error of the estimated count as the percentage of labeled tiles increases, averaged over 10,000 runs for the counting birds in the "sky" and "reeds" images, and counting roosting birds in KCLE and KGRR radar stations. Bottom: Fractional error of the estimated count after 200 labeled days for the 11 radar stations, using different estimators. The bottom-right table shows that the geometric average fractional error across stations for the raw detector and active measurement across iterations. We see that both the adaptation and the sampling without replacement are beneficial, and quickly outperform the detector and baselines for both tasks.} 
    \label{fig:station_all}
\end{figure}
\vspace{-0.02in}
\paragraph{Main results.} Our main results for both tasks are presented in Fig.~\ref{fig:station_all}. The top row illustrates that simply using raw detector predictions (i.e., directly interpreting model outputs as estimates) can lead to suboptimal performance, particularly when the detector is biased or exhibits irreducible errors. In the sky and reeds images, the detector performance improves with additional training but typically saturates. In contrast, our method demonstrates greater robustness to biased predictions and continues to improve as detector performance improves. Compared to other unbiased estimators, active measurement (AIS+DIS+WOR) outperforms DISCount (DIS) with as little as 1\% of labeled tiles, and further gains are achieved by sampling without replacement once approximately 10\% of tiles have been labeled. These improvements are particularly notable in the challenging reeds image, where the detector’s baseline performance leaves considerable room for enhancement. 

On the radar task, shown in the top-right panels and the bottom row, the detector performance (summarized in the table) is substantially worse. Fine-tuning on station-specific data reduces the fractional error from 3.78 to approximately 2, averaged across stations, but performance again saturates. However, active measurement (AIS+DIS+WOR) continues to yield improvements. Improvements are consistent across all radar stations and are substantially better than DISCount (DIS).

While the fractional error of the raw detector saturates quickly, we observe that additional fine-tuning continues to improve the proposal distribution, which in turn leads to better AIS performance (see \S~\ref{sec:hellingerdistance} for details). In Fig.~\ref{fig:end_to_end}, we also compare true end-to-end training with the more computationally efficient fixed checkpoint scheme. We observe that fractional errors between the two methods are not significantly different, especially when fewer tiles are labeled. With a higher number of labeled tiles, we see greater variance arising from the choice of the specific checkpoint; however, the relative performance remains consistent.

\paragraph{Different weighting schemes.}
\label{exp:weighting}
We now examine the performance of different weighting schemes in active measurement. According to our theory, we expect that our proposed weights $\alpha^{\mathrm{COMB}}$ work consistently well for all $t$. In Fig. \ref{fig:weighting}, we find that this is the case for the counting problem in the high resolution images. When $t$ is small, the visual detector works poorly, so weighting schemes that assign equal weights early-on, like $\alpha^{\mathrm{LURE}}$, work worse than $\alpha^{\mathrm{COMB}}$. As $t$ increases, the reduction of the sample space is more significant, so weighting schemes that do not consider this, like the $\alpha^{\mathrm{SQRT}}$ weights, works increasingly worse. For large $t$, as the performance gains from fine-tuning saturate, the performance of LURE weights will eventually converge to COMB weights, but are never better. We also test the inverse variance weighting scheme with different hyperparameters $\gamma$. The full results can be found in Fig. \ref{fig:weighting_all}. In general, a conservative $\gamma = 0.3$ brings little benefit, while an aggressive $\gamma=0.9$ can be detrimental because each individual estimator may not be accurate enough. In the middle, we find that $\gamma=0.5$ performs well and can achieve even smaller error than $\alpha^{\mathrm{COMB}}$. 

\newcommand{\height}{0.42\linewidth}
\begin{figure}[t]
    \centering   
    \begin{minipage}{0.44\textwidth}
    \centering
    \includegraphics[height=\height]{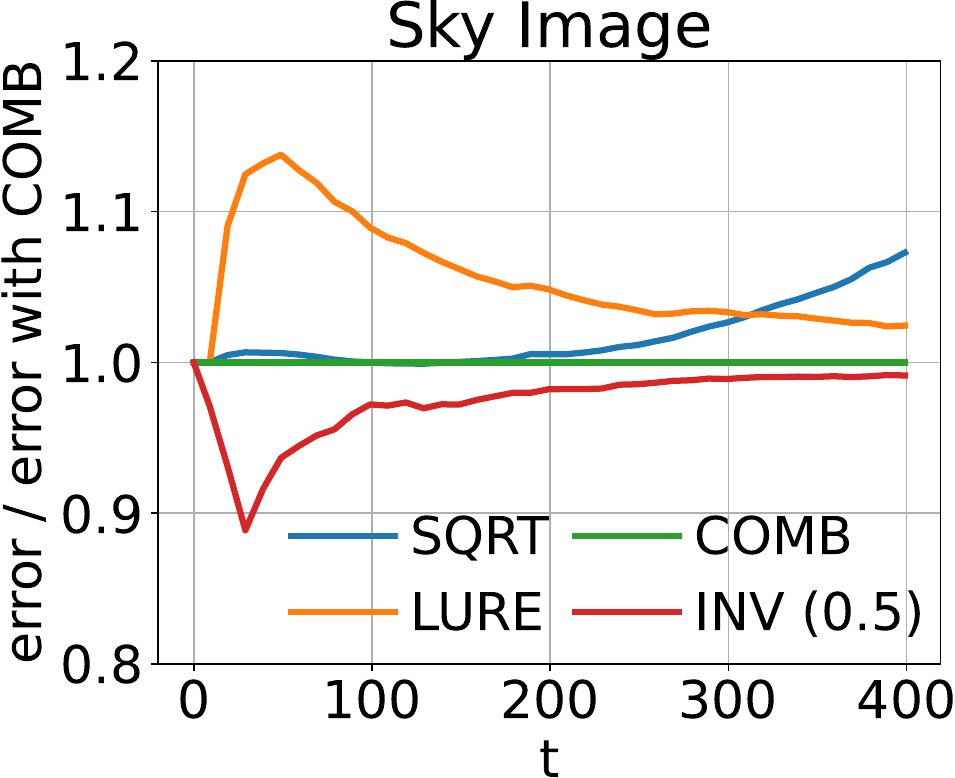}
    \includegraphics[height=\height]{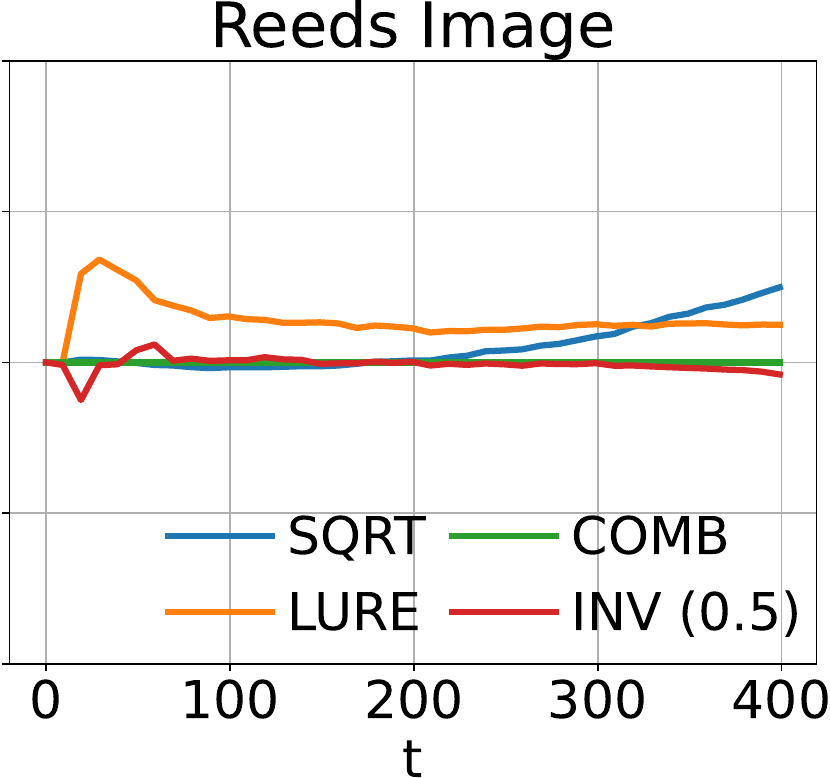}
    \caption{Relative errors compared to $\alpha^{\mathrm{COMB}}$ weighting. Other fixed weighting strategies ($\alpha^{\mathrm{SQRT}}$, $\alpha^{\mathrm{LURE}}$) are worse, but inverse variance weighting (denoted by INV ($\gamma=0.5$)) may achieve lower error. \phantom{which match our theories}
    }
    \label{fig:weighting}
    \end{minipage}
    \hspace{5pt}
\begin{minipage}{0.53\textwidth}
        \includegraphics[width=\linewidth]{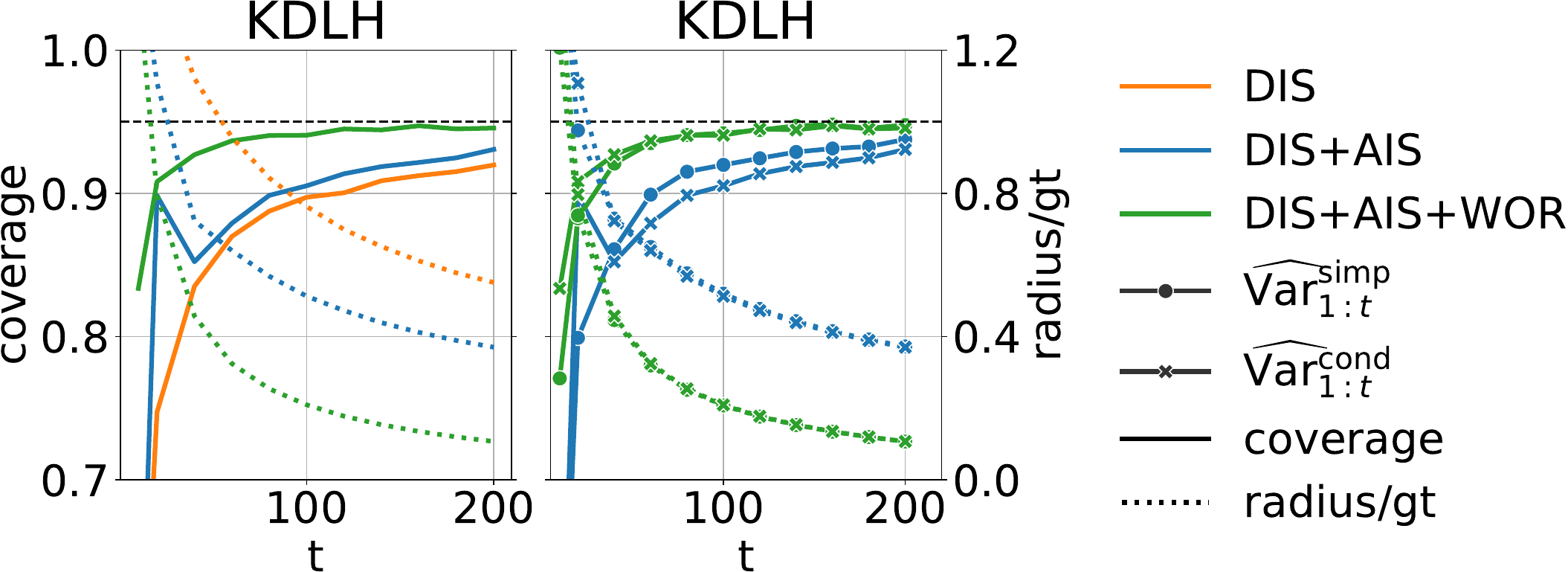}
    \caption{Coverage and radius (relative to the ground-truth) of CIs on the roost data for station KDLH as a function of $t$ (from 5,000 replications), built with either variance estimators from \S~\ref{sec:variance}. The left panel uses the $\smwidehat{\text{Var}}^{\text{cond}}_{1:t}$ estimator. We achieve the desired coverage with narrower CIs.}
    \label{fig:ci}
\end{minipage}
    \vspace{-0.1in}
\end{figure}

\paragraph{Confidence intervals.}
We evaluate variance estimators and confidence intervals by looking at CI coverage and width over 5,000 trials on the radar counting problem. Results from one station are shown in Fig. \ref{fig:ci}, and the full results can be found in Fig. \ref{fig:coverage_station_all}. Overall, we find that the DISCount baseline undercovers \emph{and} has wider confidence intervals, due to its higher estimation error---we expect width to be proportional to error if the coverage is correct. With adaptation, both metrics improve, but are different with different estimators. We compare the two variance estimators $\smwidehat{\Var}_{1:t}^{\mathrm{simp}}$ and $\smwidehat{\Var}_{1:t}^{\mathrm{cond}}$ in the second panel. In most of our experiments, the coverage with them both improves with more samples and converges to the desired confidence level.
\paragraph{Comparison with other baselines.}
\renewcommand{\height}{0.39\linewidth}

\begin{figure}[t]
    \centering   
    \begin{minipage}{0.44\textwidth}
  \begin{center}
    \includegraphics[height=\height]{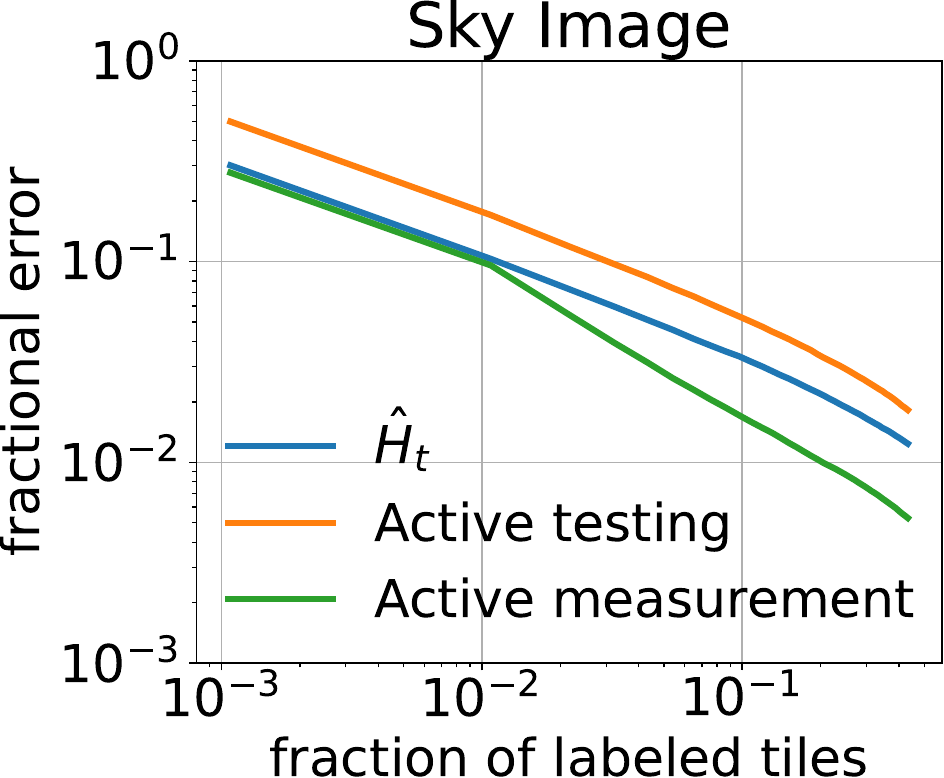}
      \includegraphics[height=\height]{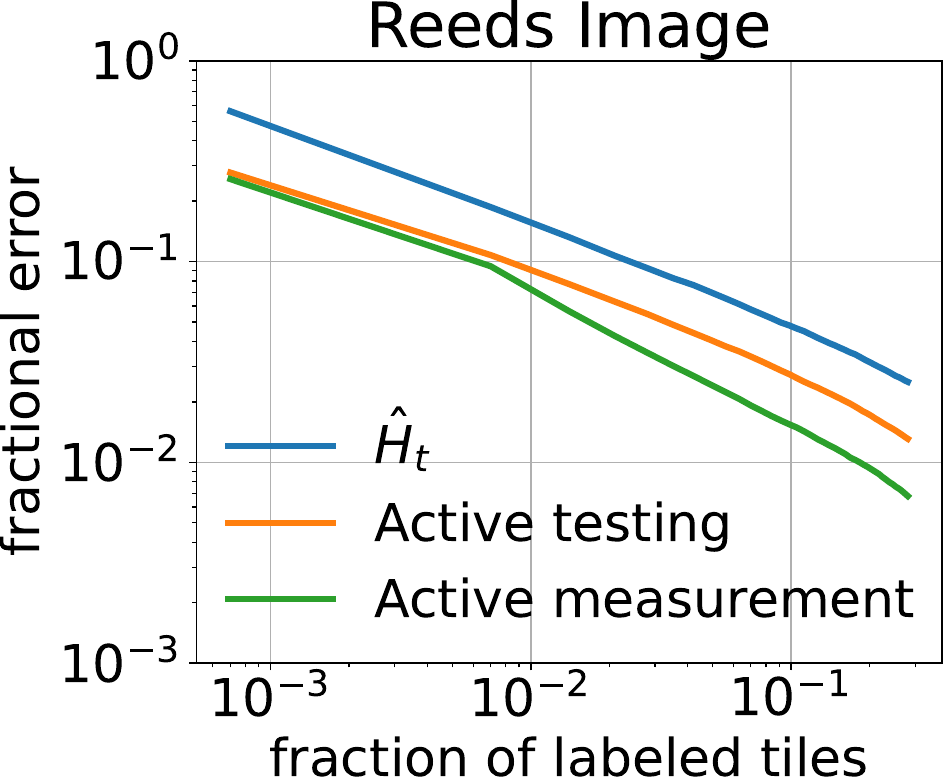}
  \end{center}
      \vspace{-5pt}
  \caption{Fractional error compared with other baselines ($\hat{H}$ is motivated by PPI).}
  \label{fig:active_testing}
    \end{minipage}
    \hspace{5pt}
\begin{minipage}{0.53\textwidth}
           \centering
    \includegraphics[width=\linewidth]{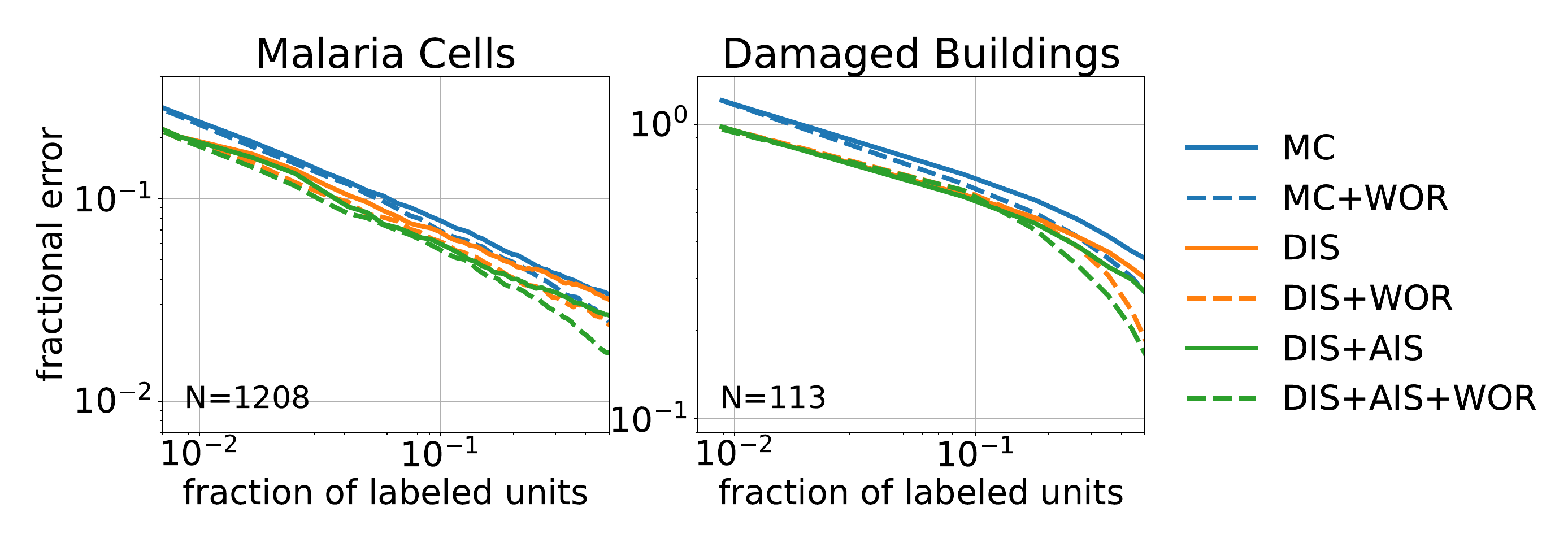}
    \caption{Estimation error on additional datasets, averaged over 1,000 end-to-end trials. }
        \label{fig:additional}
\end{minipage}
    \vspace{-0.18in}
\end{figure}

We use an experiment to illustrate the differences between active measurement and other baselines. Assume the loss function is $|f(s)-g(s)|$ for any $s\in\Omega$. Active testing will use an acquisition distribution proportional to the approximate loss to select units to label. Even if we know the ground-truth loss values, this acquisition distribution is worse than active measurement because of the mismatch between the proposal distribution and the estimand. Similarly, another estimator, motivated by PPI, is $\hat{H}_{t}=\sum_{s\in\Omega}g(s)-\frac{N}{t}\sum_{\tau=1}^t(g(s_\tau)-f(s_\tau))$  with uniformly sampled labeled data. This estimator is also worse than active measurement because it requires a fixed detector and does not select the best labels for estimation. This is shown in Fig. \ref{fig:active_testing}.

\paragraph{Additional results.} Additional experimental results on counting malaria cells and counting damaged buildings are in Fig. \ref{fig:additional}. Here we run active measurement with end-to-end settings and average over 1,000 trials. From the first figure, we observe trends similar to those seen in the image and radar experiments for malaria cell counting. DIS consistently outperforms MC, while AIS shows an early reduction in error as the model improves. WOR also demonstrates steady improvement, with its impact becoming more pronounced after roughly 10\% of the data is labeled. The results on the second dataset follow similar trends: WOR provides the most benefit when a larger fraction is labeled, while AIS is more helpful early on. We also note that our performance is lower than the numbers reported in DISCount~\citep{perez2024discount}, as they use a detector specifically trained for damaged building detection, whereas we use a simple ImageNet-pretrained backbone fine-tuned on just 5 satellite images. These results show that active measurement is capable of improving scientific estimation tasks across different domains.

\vspace{-0.1in}
\section{Conclusions, Limitations, Future Work} \label{sec:conclusion}
\vspace{-0.1in}
We introduced \emph{active measurement}, a framework that interactively leverages AI models to achieve precise scientific measurements. In contrast to traditional workflows, which are prone to errors from fully automated methods, active measurement integrates Monte Carlo estimation and model adaptation with iterative human labeling. This approach yields unbiased estimates with calibrated confidence intervals. We formally derive the unbiasedness and consistency of our estimators and propose novel techniques for sequential update weighting and uncertainty quantification. Empirical results on two measurement tasks show that active measurement not only reduces estimation error but also provides reliable uncertainty estimates, outperforming existing methods. 

One limitation of our approach is that the estimates may still not be precise enough for all applications; users should carefully validate the safety of any AI method in high-stakes scenarios. 
Another is that AI measurement has potential to be misused, e.g., to target vulnerable populations; our work does not enable new measurements but can make them more accurate.

There are several promising directions for future work. Beyond simple fine-tuning, unsupervised, semi-supervised, and transductive approaches could improve predictions across the full dataset. The current importance sampling strategy prioritizes units with high counts, but early-stage measurements could adopt more balanced sampling to better support detector training---for example, through active learning. Although model adaptation improves performance, there is a substantial computational overhead; in our experiments, we fine-tuned deep networks for detection tasks on GPUs. To enable interactivity, future systems must support fast and lightweight model updates. Approaches such as few-shot counting models and in-context learning architectures may offer practical solutions for efficient adaptation. Another promising opportunity is to consider correlation among images. As an example, in the reeds image we have observed spatial auto-correlation in the residuals of the AI model. Fitting a Gaussian process can exploit this spatial structure to improve predictions for unlabeled tiles.

\begin{ack}

We thank Maria C. T. D. Belotti for helpful discussions and contributing the algorithm to estimate bird counts from radar data. We also thank Jody Dole for kindly granting permission to use the sky and reeds photographs featured in our main experiment. This work is supported in part by NSF grants  \#2504073, \#2406687, \#2329927, and \#2210979.

\end{ack}

\bibliography{references}

\begin{thebibliography}{43}
\providecommand{\natexlab}[1]{#1}
\providecommand{\url}[1]{\texttt{#1}}
\expandafter\ifx\csname urlstyle\endcsname\relax
  \providecommand{\doi}[1]{doi: #1}\else
  \providecommand{\doi}{doi: \begingroup \urlstyle{rm}\Url}\fi

\bibitem[Mer()]{MerlinSoundID}
{Merlin Sound ID}.
\newblock \url{https://merlin.allaboutbirds.org/sound-id/}, accessed 13 October 2023.

\bibitem[Aggarwal et~al.(2021)Aggarwal, Sounderajah, Martin, Ting, Karthikesalingam, King, Ashrafian, and Darzi]{aggarwal2021diagnostic}
Ravi Aggarwal, Viknesh Sounderajah, Guy Martin, Daniel~SW Ting, Alan Karthikesalingam, Dominic King, Hutan Ashrafian, and Ara Darzi.
\newblock Diagnostic accuracy of deep learning in medical imaging: a systematic review and meta-analysis.
\newblock \emph{NPJ digital medicine}, 4\penalty0 (1):\penalty0 65, 2021.

\bibitem[Angelopoulos et~al.(2023)Angelopoulos, Bates, Fannjiang, Jordan, and Zrnic]{angelopoulos2023prediction}
Anastasios~N Angelopoulos, Stephen Bates, Clara Fannjiang, Michael~I Jordan, and Tijana Zrnic.
\newblock Prediction-powered inference.
\newblock \emph{Science}, 382\penalty0 (6671):\penalty0 669--674, 2023.

\bibitem[Arteta et~al.(2016)Arteta, Lempitsky, and Zisserman]{arteta2016counting}
Carlos Arteta, Victor Lempitsky, and Andrew Zisserman.
\newblock Counting in the wild.
\newblock In \emph{European conference on computer vision}, pages 483--498. Springer, 2016.

\bibitem[Belotti et~al.(2023)Belotti, Deng, Zhao, Simons, Cheng, Perez, Tielens, Maji, Sheldon, Kelly, et~al.]{belotti2023longterm}
Maria Carolina~TD Belotti, Yuting Deng, Wenlong Zhao, Victoria~F Simons, Zezhou Cheng, Gustavo Perez, Elske Tielens, Subhransu Maji, Daniel Sheldon, Jeffrey~F Kelly, et~al.
\newblock Long-term analysis of persistence and size of swallow and martin roosts in the {US Great Lakes}.
\newblock \emph{Remote Sensing in Ecology and Conservation}, 2023.

\bibitem[Bugallo et~al.(2017)Bugallo, Elvira, Martino, Luengo, Miguez, and Djuric]{bugallo2017adaptive}
Monica~F Bugallo, Victor Elvira, Luca Martino, David Luengo, Joaquin Miguez, and Petar~M Djuric.
\newblock Adaptive importance sampling: The past, the present, and the future.
\newblock \emph{IEEE Signal Processing Magazine}, 34\penalty0 (4):\penalty0 60--79, 2017.

\bibitem[Buler et~al.(2012)Buler, Randall, Fleskes, Barrow~Jr, Bogart, and Kluver]{buler2012mapping}
Jeffrey~J Buler, Lori~A Randall, Joseph~P Fleskes, Wylie~C Barrow~Jr, Tianna Bogart, and Daria Kluver.
\newblock Mapping wintering waterfowl distributions using weather surveillance radar.
\newblock \emph{PloS one}, 7\penalty0 (7):\penalty0 e41571, 2012.

\bibitem[Chapelle et~al.(2009)Chapelle, Scholkopf, and Zien]{4787647}
O.~Chapelle, B.~Scholkopf, and A.~Zien, Eds.
\newblock Semi-supervised learning [book reviews].
\newblock \emph{IEEE Transactions on Neural Networks}, 20\penalty0 (3):\penalty0 542--542, 2009.
\newblock \doi{10.1109/TNN.2009.2015974}.

\bibitem[Deng et~al.(2009)Deng, Dong, Socher, Li, Li, and Fei-Fei]{5206848}
Jia Deng, Wei Dong, Richard Socher, Li-Jia Li, Kai Li, and Li~Fei-Fei.
\newblock Imagenet: A large-scale hierarchical image database.
\newblock In \emph{2009 IEEE Conference on Computer Vision and Pattern Recognition}, pages 248--255, 2009.
\newblock \doi{10.1109/CVPR.2009.5206848}.

\bibitem[Deng et~al.(2023)Deng, Belotti, Zhao, Cheng, Perez, Tielens, Simons, Sheldon, Maji, Kelly, et~al.]{deng2023quantifying}
Yuting Deng, Maria Carolina~TD Belotti, Wenlong Zhao, Zezhou Cheng, Gustavo Perez, Elske Tielens, Victoria~F Simons, Daniel~R Sheldon, Subhransu Maji, Jeffrey~F Kelly, et~al.
\newblock Quantifying long-term phenological patterns of aerial insectivores roosting in the {Great Lakes} region using weather surveillance radar.
\newblock \emph{Global Change Biology}, 29\penalty0 (5):\penalty0 1407--1419, 2023.

\bibitem[Dutta et~al.(2016)Dutta, Gupta, and Zissermann]{dutta2016via}
A.~Dutta, A.~Gupta, and A.~Zissermann.
\newblock {VGG} image annotator ({VIA}), 2016.
\newblock Version: 2.0.12, Accessed: Mar. 2025.

\bibitem[Farquhar et~al.(2021)Farquhar, Gal, and Rainforth]{FarquharGR21}
Sebastian Farquhar, Yarin Gal, and Tom Rainforth.
\newblock On statistical bias in active learning: How and when to fix it.
\newblock In \emph{9th International Conference on Learning Representations, {ICLR} 2021, Virtual Event, Austria, May 3-7, 2021}, 2021.

\bibitem[Fuentes-Pe{\~n}ailillo et~al.(2024)Fuentes-Pe{\~n}ailillo, Gutter, Vega, and Silva]{fuentes2024transformative}
Fernando Fuentes-Pe{\~n}ailillo, Karen Gutter, Ricardo Vega, and Gilda~Carrasco Silva.
\newblock Transformative technologies in digital agriculture: Leveraging internet of things, remote sensing, and artificial intelligence for smart crop management.
\newblock \emph{Journal of Sensor and Actuator Networks}, 13\penalty0 (4):\penalty0 39, 2024.

\bibitem[Gao et~al.(2020)Gao, Sun, Hu, and Zhang]{gao2020framework}
Demin Gao, Quan Sun, Bin Hu, and Shuo Zhang.
\newblock A framework for agricultural pest and disease monitoring based on internet-of-things and unmanned aerial vehicles.
\newblock \emph{Sensors}, 20\penalty0 (5):\penalty0 1487, 2020.

\bibitem[Gupta et~al.(2019)Gupta, Goodman, Patel, Hosfelt, Sajeev, Heim, Doshi, Lucas, Choset, and Gaston]{gupta2019creating}
Ritwik Gupta, Bryce Goodman, Nirav Patel, Ricky Hosfelt, Sandra Sajeev, Eric Heim, Jigar Doshi, Keane Lucas, Howie Choset, and Matthew Gaston.
\newblock Creating {xBD}: A dataset for assessing building damage from satellite imagery.
\newblock In \emph{Proceedings of the IEEE/CVF conference on computer vision and pattern recognition workshops}, pages 10--17, 2019.

\bibitem[Hall and Heyde(2014)]{hall2014martingale}
Peter Hall and Christopher~C Heyde.
\newblock \emph{Martingale limit theory and its application}.
\newblock Academic press, 2014.

\bibitem[He et~al.(2015)He, Zhang, Ren, and Sun]{he2015deepresiduallearningimage}
Kaiming He, Xiangyu Zhang, Shaoqing Ren, and Jian Sun.
\newblock Deep residual learning for image recognition, 2015.
\newblock URL \url{https://arxiv.org/abs/1512.03385}.

\bibitem[Horn and Kunz(2008)]{horn2008analyzing}
Jason~W Horn and Thomas~H Kunz.
\newblock Analyzing {NEXRAD} doppler radar images to assess nightly dispersal patterns and population trends in {Brazilian} free-tailed bats (tadarida brasiliensis).
\newblock \emph{Integrative and Comparative Biology}, 48\penalty0 (1):\penalty0 24--39, 2008.

\bibitem[Kay et~al.(2022)Kay, Kulits, Stathatos, Deng, Young, Beery, Van~Horn, and Perona]{kay2022caltech}
Justin Kay, Peter Kulits, Suzanne Stathatos, Siqi Deng, Erik Young, Sara Beery, Grant Van~Horn, and Pietro Perona.
\newblock The {Caltech} fish counting dataset: A benchmark for multiple-object tracking and counting.
\newblock In \emph{European Conference on Computer Vision}, pages 290--311. Springer, 2022.

\bibitem[Kossen et~al.(2021)Kossen, Farquhar, Gal, and Rainforth]{kossen2021active}
Jannik Kossen, Sebastian Farquhar, Yarin Gal, and Tom Rainforth.
\newblock Active testing: Sample-efficient model evaluation.
\newblock In \emph{International Conference on Machine Learning}, pages 5753--5763. PMLR, 2021.

\bibitem[Kovashka et~al.(2016)Kovashka, Russakovsky, Fei-Fei, Grauman, et~al.]{kovashka2016crowdsourcing}
Adriana Kovashka, Olga Russakovsky, Li~Fei-Fei, Kristen Grauman, et~al.
\newblock Crowdsourcing in computer vision.
\newblock \emph{Foundations and Trends{\textregistered} in computer graphics and Vision}, 10\penalty0 (3):\penalty0 177--243, 2016.

\bibitem[Lang et~al.(2023)Lang, Jetz, Schindler, and Wegner]{lang2023high}
Nico Lang, Walter Jetz, Konrad Schindler, and Jan~Dirk Wegner.
\newblock A high-resolution canopy height model of the earth.
\newblock \emph{Nature Ecology \& Evolution}, pages 1--12, 2023.

\bibitem[Lintott et~al.(2008)Lintott, Schawinski, Slosar, Land, Bamford, Thomas, Raddick, Nichol, Szalay, Andreescu, et~al.]{lintott2008galaxy}
Chris~J Lintott, Kevin Schawinski, An{\v{z}}e Slosar, Kate Land, Steven Bamford, Daniel Thomas, M~Jordan Raddick, Robert~C Nichol, Alex Szalay, Dan Andreescu, et~al.
\newblock Galaxy {Zoo}: morphologies derived from visual inspection of galaxies from the {Sloan Digital Sky Survey}.
\newblock \emph{Monthly Notices of the Royal Astronomical Society}, 389\penalty0 (3):\penalty0 1179--1189, 2008.

\bibitem[Ljosa et~al.(2012)Ljosa, Sokolnicki, and Carpenter]{ljosa2012annotated}
Vebjorn Ljosa, Katherine~L Sokolnicki, and Anne~E Carpenter.
\newblock Annotated high-throughput microscopy image sets for validation.
\newblock \emph{Nature methods}, 9\penalty0 (7):\penalty0 637, 2012.

\bibitem[Meng et~al.(2022)Meng, Liu, Neiswanger, Song, Burke, Lobell, and Ermon]{meng2022count}
Chenlin Meng, Enci Liu, Willie Neiswanger, Jiaming Song, Marshall Burke, David Lobell, and Stefano Ermon.
\newblock {IS-Count}: Large-scale object counting from satellite images with covariate-based importance sampling.
\newblock In \emph{Proceedings of the AAAI Conference on Artificial Intelligence}, volume~36, pages 12034--12042, 2022.

\bibitem[Nguyen et~al.(2018)Nguyen, Ramanan, and Fowlkes]{nguyen2018active}
Phuc Nguyen, Deva Ramanan, and Charless Fowlkes.
\newblock Active testing: An efficient and robust framework for estimating accuracy.
\newblock In \emph{International Conference on Machine Learning}, pages 3759--3768. PMLR, 2018.

\bibitem[Oh and Berger(1992)]{oh1992adaptive}
Man-Suk Oh and James~O Berger.
\newblock Adaptive importance sampling in {M}onte {C}arlo integration.
\newblock \emph{Journal of statistical computation and simulation}, 41\penalty0 (3-4):\penalty0 143--168, 1992.

\bibitem[Owen(2013)]{owen2013monte}
Art~B Owen.
\newblock Monte {Carlo} theory, methods and examples, 2013.

\bibitem[Owen and Zhou(2020)]{owen2020square}
Art~B Owen and Yi~Zhou.
\newblock The square root rule for adaptive importance sampling.
\newblock \emph{ACM Transactions on Modeling and Computer Simulation (TOMACS)}, 30\penalty0 (2):\penalty0 1--12, 2020.

\bibitem[Patterson et~al.(2015)Patterson, Van~Horn, Belongie, Perona, and Hays]{patterson2015tropel}
Genevieve Patterson, Grant Van~Horn, Serge Belongie, Pietro Perona, and James Hays.
\newblock Tropel: Crowdsourcing detectors with minimal training.
\newblock In \emph{Proceedings of the AAAI Conference on Human Computation and Crowdsourcing}, volume~3, pages 150--159, 2015.

\bibitem[Perez et~al.(2022)Perez, Zhao, Cheng, T.~D.~Belotti, Deng, Simons, Tielens, Kelly, Horton, Maji, and Sheldon]{Perez2022.10.28.513761}
Gustavo Perez, Wenlong Zhao, Zezhou Cheng, Maria~Carolina T.~D.~Belotti, Yuting Deng, Victoria~F. Simons, Elske Tielens, Jeffrey~F. Kelly, Kyle~G. Horton, Subhransu Maji, and Daniel Sheldon.
\newblock Using spatio-temporal information in weather radar data to detect and track communal bird roosts.
\newblock \emph{bioRxiv}, 2022.
\newblock \doi{10.1101/2022.10.28.513761}.
\newblock URL \url{https://www.biorxiv.org/content/early/2022/10/31/2022.10.28.513761}.

\bibitem[Perez et~al.(2024)Perez, Maji, and Sheldon]{perez2024discount}
Gustavo Perez, Subhransu Maji, and Daniel Sheldon.
\newblock {DISCount}: counting in large image collections with detector-based importance sampling.
\newblock In \emph{Proceedings of the AAAI Conference on Artificial Intelligence}, volume~38, pages 22294--22302, 2024.

\bibitem[Portier and Delyon(2018)]{portier2018asymptotic}
Fran{\c{c}}ois Portier and Bernard Delyon.
\newblock Asymptotic optimality of adaptive importance sampling.
\newblock \emph{Advances in Neural Information Processing Systems}, 31, 2018.

\bibitem[Ranjan et~al.(2021)Ranjan, Sharma, Nguyen, and Hoai]{ranjan2021learning}
Viresh Ranjan, Udbhav Sharma, Thu Nguyen, and Minh Hoai.
\newblock Learning to count everything.
\newblock In \emph{Proceedings of the IEEE/CVF Conference on Computer Vision and Pattern Recognition}, pages 3394--3403, 2021.

\bibitem[Ren et~al.(2016)Ren, He, Girshick, and Sun]{ren2016fasterrcnnrealtimeobject}
Shaoqing Ren, Kaiming He, Ross Girshick, and Jian Sun.
\newblock Faster r-cnn: Towards real-time object detection with region proposal networks, 2016.
\newblock URL \url{https://arxiv.org/abs/1506.01497}.

\bibitem[Thanh~Nguyen and Hoai(2022)]{countingdetr2022}
Khoi~Nguyen Thanh~Nguyen, Chau~Pham and Minh Hoai.
\newblock {Few-shot Object Counting and Detection}.
\newblock In \emph{Proceedings of the European Conference on Computer Vision 2022}, 2022.

\bibitem[Thirunavukarasu et~al.(2023)Thirunavukarasu, Ting, Elangovan, Gutierrez, Tan, and Ting]{thirunavukarasu2023large}
Arun~James Thirunavukarasu, Darren Shu~Jeng Ting, Kabilan Elangovan, Laura Gutierrez, Ting~Fang Tan, and Daniel Shu~Wei Ting.
\newblock Large language models in medicine.
\newblock \emph{Nature medicine}, 29\penalty0 (8):\penalty0 1930--1940, 2023.

\bibitem[Tuia et~al.(2022)Tuia, Kellenberger, Beery, Costelloe, Zuffi, Risse, Mathis, Mathis, van Langevelde, Burghardt, et~al.]{tuia2022perspectives}
Devis Tuia, Benjamin Kellenberger, Sara Beery, Blair~R Costelloe, Silvia Zuffi, Benjamin Risse, Alexander Mathis, Mackenzie~W Mathis, Frank van Langevelde, Tilo Burghardt, et~al.
\newblock Perspectives in machine learning for wildlife conservation.
\newblock \emph{Nature communications}, 13\penalty0 (1):\penalty0 792, 2022.

\bibitem[Van~Horn et~al.(2018)Van~Horn, Mac~Aodha, Song, Cui, Sun, Shepard, Adam, Perona, and Belongie]{van2018inaturalist}
Grant Van~Horn, Oisin Mac~Aodha, Yang Song, Yin Cui, Chen Sun, Alex Shepard, Hartwig Adam, Pietro Perona, and Serge Belongie.
\newblock The {iNaturalist} species classification and detection dataset.
\newblock In \emph{Proceedings of the IEEE conference on computer vision and pattern recognition}, pages 8769--8778, 2018.

\bibitem[Walmsley et~al.(2021)Walmsley, Lintott, Géron, Kruk, Krawczyk, Willett, Bamford, Kelvin, Fortson, Gal, Keel, Masters, Mehta, Simmons, Smethurst, Smith, Baeten, and Macmillan]{Walmsley_2021}
Mike Walmsley, Chris Lintott, Tobias Géron, Sandor Kruk, Coleman Krawczyk, Kyle~W Willett, Steven Bamford, Lee~S Kelvin, Lucy Fortson, Yarin Gal, William Keel, Karen~L Masters, Vihang Mehta, Brooke~D Simmons, Rebecca Smethurst, Lewis Smith, Elisabeth~M Baeten, and Christine Macmillan.
\newblock Galaxy {Zoo DECaLS}: Detailed visual morphology measurements from volunteers and deep learning for 314,000 galaxies.
\newblock \emph{Monthly Notices of the Royal Astronomical Society}, 509\penalty0 (3):\penalty0 3966–3988, September 2021.
\newblock ISSN 1365-2966.
\newblock \doi{10.1093/mnras/stab2093}.

\bibitem[Wang et~al.(2023)Wang, Fu, Du, Gao, Huang, Liu, Chandak, Liu, Van~Katwyk, Deac, et~al.]{wang2023scientific}
Hanchen Wang, Tianfan Fu, Yuanqi Du, Wenhao Gao, Kexin Huang, Ziming Liu, Payal Chandak, Shengchao Liu, Peter Van~Katwyk, Andreea Deac, et~al.
\newblock Scientific discovery in the age of artificial intelligence.
\newblock \emph{Nature}, 620\penalty0 (7972):\penalty0 47--60, 2023.

\bibitem[Wu et~al.(2019)Wu, Kirillov, Massa, Lo, and Girshick]{wu2019detectron2}
Yuxin Wu, Alexander Kirillov, Francisco Massa, Wan-Yen Lo, and Ross Girshick.
\newblock Detectron2.
\newblock \url{https://github.com/facebookresearch/detectron2}, 2019.

\bibitem[Zhou(1998)]{zhou1998adaptive}
Yi~Zhou.
\newblock \emph{Adaptive importance sampling for integration}.
\newblock Phd thesis, Stanford University, 1998.

\end{thebibliography}

\clearpage
\newpage

\clearpage
\appendix
\setcounter{table}{0}   
\renewcommand{\thetable}{A\arabic{table}}
\setcounter{figure}{0}
\renewcommand{\thefigure}{A\arabic{figure}}

\label{sec:appendix}
\setcounter{proposition}{0}
\setcounter{lemma}{0}

\section{Proof of the theoretical results}
In this section, we restate each theoretical result and provide proofs.

\subsection{Proof of Proposition 1}
\begin{proposition}
    The combined estimator $\hat{F}_{1:t}=\sum_{\tau=1}^t\alp_\tau \hat{F}_\tau$ is unbiased: $\E[\hat{F}_{1:t}]=F(\Omega)$. 
\end{proposition}
\begin{proof}
    We first show that each individual estimator $\hat{F}_\tau$ is unbiased.
    \begin{align}
        \E_{\D_\tau,s_\tau}[\hat{F}_\tau]&=\E_{\D_\tau,s_\tau}\left[F(D_\tau)+\frac{f(s_\tau)}{q_\tau({s_\tau})}\right]\notag\\
        &=\E_{\D_\tau}\left[F(D_\tau)+\E_{s_\tau}\left[\frac{f(s_\tau)}{q_\tau({s_\tau})}\right]\right]\notag\\
        &=\E_{\D_\tau}\left[F(D_\tau)+\sum_{s_\tau\in\Omega\backslash\D_\tau}q_\tau(s_\tau)\frac{f(s_\tau)}{q_\tau({s_\tau})}\right]\notag\\
        &=\E_{\D_\tau}\left[F(D_\tau)+F(\Omega\backslash\D_\tau)\right]\notag\\
        &=F(\Omega).\notag
    \end{align}
    Therefore, for the combined estimator
    \begin{align}
        \E[\hat{F}_{1:t}]&=\E\left[\sum_{\tau=1}^t\alp_\tau \hat{F}_\tau\right]\notag\\
        &=\sum_{\tau=1}^t\alp_\tau \E[\hat{F}_\tau]\notag\\
        &=\sum_{\tau=1}^t\alp_\tau F(\Omega)\notag\\
        &=F(\Omega).\notag
    \end{align}
\end{proof}

\subsection{Proof of proposition 2}
\begin{proposition}
    For any $1\le\tau<r\le t$, $\Cov(\hat{F}_\tau,\hat{F}_r)=0$.
\end{proposition}
\begin{proof}
    \begin{align}
        \Cov[\hat{F}_{\tau},\hat{F}_r]&=\E[(\hat{F}_\tau-F(\Omega))(\hat{F}_r-F(\Omega))]\notag\\
        &=\E_{\D_r}[(\hat{F}_\tau-F(\Omega))\E_{s_r}[\hat{F}_r-F(\Omega)]]\notag\\
        &=\E_{\D_r}\left[(\hat{F}_\tau-F(\Omega))\sum_{s_r\in\Omega\backslash\D_r}q_r(s_r)\left(F(\D_r)+\frac{f(s_r)}{q_r(s_r)}-F(\Omega)\right)\right]\notag\\
        &=\E_{\D_r}\left[(\hat{F}_\tau-F(\Omega))\left(F(\D_r)+\sum_{s_r\in\Omega\backslash\D_r}q_r(s_r)\frac{f(s_r)}{q_r(s_r)}-F(\Omega)\right)\right]\notag\\
        &=\E_{\D_r}\left[(\hat{F}_\tau-F(\Omega))\cdot 0\right]\notag\\
        &=0.\notag
    \end{align}
\end{proof}

\subsection{Proof of proposition 3}
\begin{proposition}
    If there exists constants $0<A\le B$ such that for any $s\in \Omega$, $A\le f(s)\le B$ and $A\le g(s)\le B$, then there exists a constant $C>0$ such that $\Var [\hat{F}_\tau]\le C(N-\tau)(N-\tau+1)$. 
\end{proposition}
\begin{proof}
    To start, we first bound the sampling distribution implied by the detector measurements $g(s)$. For unit $s\in \Omega\backslash \D_\tau$, the sampling distribution is $q_\tau(s)=\frac{g(s)}{\sum_{s'\in\Omega\backslash \D_\tau}g(s')}$. By the lower and upper bounds, we can see that $\frac{A}{(N-\tau+1)B}\le q_\tau(s)\le \frac{B}{(N-\tau+1)A}$. Also, we can bound $F(\Omega)-F(\D_\tau)=\sum_{s\in\Omega\backslash\D_\tau}f(s)$ with the same technique, which is $(N-\tau+1)A\le F(\Omega)-F(\D_\tau)\le (N-\tau+1)B$.

    Next, we write the variance of $\hat{F}_\tau$ in terms of expectations.
    \begin{align}
        \Var[\hat{F}_\tau]&=\E_{\D_\tau,s_\tau}\left[\left(F(\D_\tau)+\frac{f(s_\tau)}{q_\tau(s_\tau)}-F(\Omega)\right)^2\right]\notag\\
        &=\E_{\D_\tau,s_\tau}\left[\frac{f(s_\tau)^2}{q_\tau(s_\tau)^2}+(F(\D_\tau)-F(\Omega))^2+2(F(\D_\tau)-F(\Omega))\frac{f(s_\tau)}{q_\tau(s_\tau)}\right]\notag\\
        &\le \E_{\D_\tau,s_\tau}\left[\frac{(N-\tau+1)^2B^4}{A^2}+(N-\tau+1)^2B^2-2(N-\tau+1)^2\frac{A^3}{B}\right]\notag\\
        &=(N-\tau+1)^2\frac{B^3(B^2+A^2)-2A^5}{A^2B}.\notag
    \end{align}
    When $\tau<N$, we further have
    \begin{align}
        \Var[\hat{F}_\tau]&=(N-\tau+1)^2\frac{B^3(B^2+A^2)-2A^5}{A^2B}\notag\\
        &= (N-\tau)(N-\tau+1)\frac{N-\tau+1}{N-\tau}\frac{B^3(B^2+A^2)-2A^5}{A^2B}\notag\\
        &\le  (N-\tau)(N-\tau+1)\frac{2B^3(B^2+A^2)-4A^5}{A^2B}.\notag
    \end{align}
    Also note that
    \begin{align}
        \Var[\hat{F}_N]&=\E_{\D_N,s_N}\left[\left(F(\D_N)+\frac{f(s_N)}{q_N(s_N)}-F(\Omega)\right)^2\right]\notag\\
        &=\E_{\D_N,s_N}\left[\left(F(\D_N)+\frac{f(s_N)}{1}-F(\Omega)\right)^2\right]\notag\\
        &=\E_{\D_N,s_N}\left[\left(F(\Omega)-F(\Omega)\right)^2\right]=0.\notag
    \end{align}
    This means that for all $\tau\le N$, $\Var[\hat{F}_\tau]\le C(N-\tau)(N-\tau+1)$, where $C=\frac{2B^3(B^2+A^2)-4A^5}{A^2B}$.
\end{proof}

\subsection{Proof of proposition 4}
\begin{proposition}
    If $\Var[\hat{F}_\tau]\propto \tau^{-y}/w_\tau$, where $y\in[0,1]$, and $w_\tau$ are non-decreasing, then the estimate $\hat{F}_{1:t}^{\mathrm{comb}}=\sum_{\tau=1}^t\alpha_\tau^{\mathrm{comb}}\hat{F}_\tau$, where $\alpha_\tau^{\mathrm{comb}}\propto w_\tau\sqrt{\tau}$ and $\sum_{\tau=1}^t\alpha_\tau^{\mathrm{comb}}=1$, satisfy 
    \begin{align}
        \sup_{t\ge 1}\sup_{y\in[0,1]}\frac{\Var[\hat{F}^{\mathrm{comb}}_{1:t}]}{\Var_\mathrm{opt}(y,t)} \le \frac{9}{8},\notag
    \end{align}
    where $\Var_\mathrm{opt}(y,t)$ is the estimation variance when the estimators are weighted proportional to inverse of ground-truth variances.
\end{proposition}
\begin{proof}
    Denote the variance of combined estimator weighted by $\alpha^{(x)}_{\tau}\propto w_\tau \tau^{x}$ by $\Var_{x}(y,t)$, where $x\in[0,1]$. An observation is that $\Var[\hat{F}^{\mathrm{comb}}_{1:t}]=\Var_{0.5}(y,t)$, and $\Var_{\mathrm{opt}}(y,t)=\Var_y(y,t)$. So we shall bound $\frac{\Var_{0.5}(y,t)}{\Var_{y}(y,t)}$. Note that for any $x\in[0,1]$,
    \begin{align}
        \Var_{x}(y,t)&\propto\frac{\sum_{\tau=1}^t w_\tau^2 \tau^{2x} \cdot \tau^{-y}/w_\tau}{\left(\sum_{\tau=1}^t w_\tau \tau^{x}\right)^2}\notag\\
        &=\frac{\sum_{\tau=1}^t w_\tau \tau^{2x-y}}{\left(\sum_{\tau=1}^t w_\tau \tau^{x}\right)^2}.\notag
    \end{align}
    Therefore, let
    \begin{align}
        l_x(y,t)&=\frac{\Var_{x}(y,t)}{\Var_{y}(y,t)}
        =\frac{\left(\sum_{\tau=1}^t w_\tau \tau^{2x-y}\right)\left(\sum_{\tau=1}^t w_\tau \tau^y\right)}{\left(\sum_{\tau=1}^t w_\tau \tau^{x} \right)^2},\notag
    \end{align}
    we have
    \begin{align}
        \frac{\partial^2 l_x(y,t)}{\partial y^2}&=\frac{\sum_{\tau=1}^t\sum_{\tau'=1}^t w_\tau w_\tau' \tau^{2x-y}\tau'^{y}(\log\tau-\log\tau')^2}{\left(\sum_{\tau=1}^t w_\tau \tau^{x}\right)^2}> 0.\notag
    \end{align}
    So $l_x(y,t)$ is strictly convex over $y$. Also note that $l_x(y,t)$ is symmetric around $y=x$. So $\sup_{y\in[0,1]}l_x(y,t)=\begin{cases}l_x(1,t)&x\le 1/2\\l_x(0,t)&x\ge1/2\end{cases}$. In the combined estimator we chose $x=0.5$, which leads to the supremum $l_{0.5}(1,t)$ for any $t$. Furthermore, $x=0.5$ is the best assumption for the mixing weights. The derivation is the same as \citet{owen2020square} and we omit it here. In such case
    \begin{align}
        \sup_{y\in[0,1]}l_{0.5}(y,t)=l_{0.5}(1,t) =
        \frac{{\left(\sum_{\tau=1}^t w_\tau \right)}{\left(\sum_{\tau=1}^t w_\tau \tau \right)}}{\left(\sum_{\tau=1}^t w_\tau \tau^{0.5}\right)^2} =
        \frac{{\sum_{\tau=1}^t \frac{w_\tau}{\left(\sum_{\tau'=1}^t w_{\tau'} \right)} \tau}}{\left(\sum_{\tau=1}^t \frac{w_\tau}{\left(\sum_{\tau'=1}^t w_{\tau'} \right)} \tau^{0.5}\right)^2}.\notag
    \end{align}
    We can plot this specific function for a few different choices of $w_\tau$ and $t$,
    \begin{align}
        \includegraphics[width=.7\textwidth]{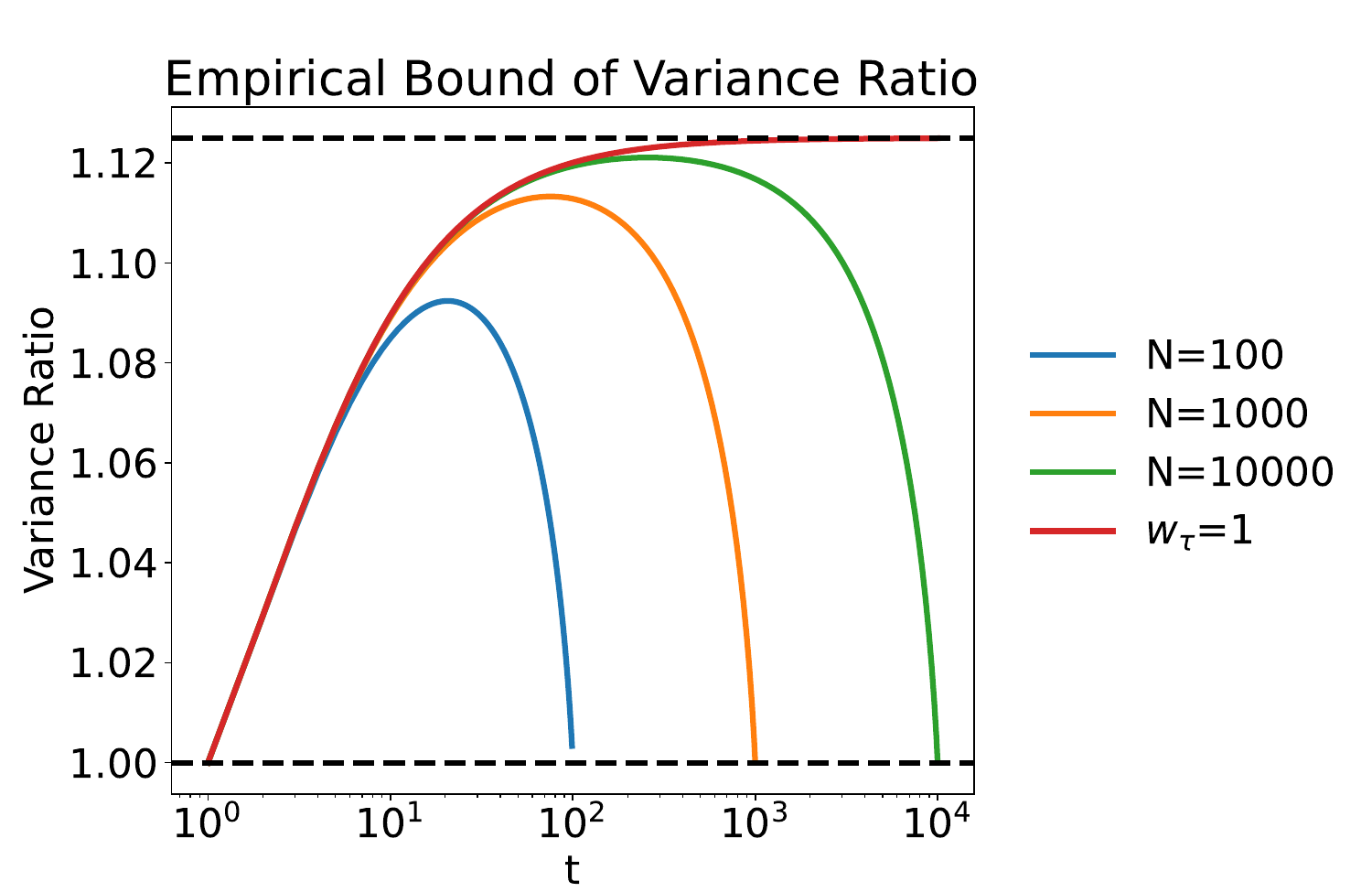}\notag
    \end{align}
    The lines labeled $N=x$ correspond to the combined weighting scheme proposed in our main method (see Section \ref{sec:weighting}), and $w_\tau = 1$ is a uniform weighting. We can see empirically that these are all bounded above by $\frac{9}{8}$.
    
    To prove this bound, we define the following weights,
    \begin{align}
        \hat{w}_\tau = \frac{w_\tau}{\left(\sum_{\tau'=1}^t w_{\tau'} \right)}, \quad \quad \beta_\tau = (\hat{w}_\tau-\hat{w}_{\tau-1})(t-\tau+1).\notag
    \end{align}
    Since $w_\tau$ are non-decreasing, $\hat{w}_\tau - \hat{w}_{\tau-1} \ge 0$ and thus $\beta_\tau \ge 0$. We can also cancel out terms of a telescoping series to see that $\sum_{\tau=1}^t \beta_\tau = \sum_{\tau=1}^t \hat{w}_\tau = 1$. Therefore both $\hat{w}_\tau$ and $\beta_\tau$ are non-negative weights that sum to 1. Note we define $\hat{w}_0 = 0$.
    
    Plugging these in, and borrowing a trick from Gabriel's Staircase, we can rewrite as,
    \begin{align}
        \frac{{\sum_{\tau=1}^t \frac{w_\tau}{\left(\sum_{\tau'=1}^t w_{\tau'} \right)} \tau}}{\left(\sum_{\tau=1}^t \frac{w_\tau}{\left(\sum_{\tau'=1}^t w_{\tau'} \right)} \tau^{0.5}\right)^2} = 
        \frac{{\sum_{\tau=1}^t \hat{w}_\tau \tau}}{\left(\sum_{\tau=1}^t \hat{w}_\tau \tau^{0.5}\right)^2} =
        \frac{{\sum_{\tau=1}^t \beta_{\tau} \frac{1}{t-\tau+1}\sum_{\tau'=\tau}^t \tau'}}{\left(\sum_{\tau=1}^t \beta_{\tau} \frac{1}{t-\tau+1} \sum_{\tau'=\tau}^t \tau'^{0.5}\right)^2}.\notag
    \end{align}

    We can bound this expression by considering the supremum over all possible positive weights $\lambda_\tau$,

    \begin{align}
        \frac{{\sum_{\tau=1}^t \beta_{\tau} \frac{1}{t-\tau+1}\sum_{\tau'=\tau}^t \tau'}}{\left(\sum_{\tau=1}^t \beta_{\tau} \frac{1}{t-\tau+1} \sum_{\tau'=\tau}^t \tau'^{0.5}\right)^2} \le \sup_{\lambda \in \mathbb{R}_{\ge 0}^t} \frac{{\sum_{\tau=1}^t \frac{\lambda_{\tau}}{\sum_{\tau'=1}^t \lambda_{\tau'}} \frac{1}{t-\tau+1}\sum_{\tau'=\tau}^t \tau'}}{\left(\sum_{\tau=1}^t \frac{\lambda_{\tau}}{\sum_{\tau'=1}^t \lambda_{\tau'}} \frac{1}{t-\tau+1} \sum_{\tau'=\tau}^t \tau'^{0.5}\right)^2}\notag\\=
        \sup_{\lambda \in \mathbb{R}_{\ge 0}^t} \frac{\left(\sum_{\tau=1}^t \lambda_{\tau}\right)\left(\sum_{\tau=1}^t \frac{\lambda_{\tau}}{t-\tau+1}\sum_{\tau'=\tau}^t \tau'\right)}{\left(\sum_{\tau=1}^t \frac{\lambda_{\tau}}{t-\tau+1} \sum_{\tau'=\tau}^t \tau'^{0.5}\right)^2}.\notag
    \end{align}
    
    We can simplify by using the closed form expression for sum of consecutive integers and a lower bound for the sum of consecutive square roots from \citet{owen2020square},
    \begin{align}
        \le \sup_{\lambda \in \mathbb{R}_{\ge 0}^t} \frac{\left(\sum_{\tau=1}^t \lambda_{\tau}\right)\left(\sum_{\tau=1}^t \frac{\lambda_{\tau}}{t-\tau+1}\frac{(t+\tau)(t-\tau+1)}{2}\right)}{\left(\sum_{\tau=1}^t  \frac{\lambda_{\tau}}{t-\tau+1} \frac{(t+0.5)^{1.5}-(\tau-0.5)^{1.5}}{1.5}\right)^2} \notag\\= 
        \sup_{\lambda \in \mathbb{R}_{\ge 0}^t} \frac{9}{8}\frac{\left(\sum_{\tau=1}^t \lambda_{\tau}\right)\left(\sum_{\tau=1}^t \lambda_{\tau}(t+\tau)\right)}{\left(\sum_{\tau=1}^t  \frac{\lambda_{\tau}}{t-\tau+1} \left((t+0.5)^{1.5}-(\tau-0.5)^{1.5}\right)\right)^2}.
        \label{line:34}
    \end{align}

    Define this last expression inside the supremum as $f(\lambda, t)$. Taking the derivative with respect to each $\lambda_n$,
    \begin{tiny}
    \AddToHook{normalfont}[normalsize]{\normalsize}
    \begin{align}
        \frac{\partial f(\lambda,t)}{\partial \lambda_n} = \frac{\left(\sum_{\tau=1}^{t}\lambda_{\tau}\left(t+\tau\right)\right)+\left(t+n\right)\left(\sum_{\tau=1}^{t}\lambda_{\tau}\right)}{\left(\sum_{\tau=1}^{t}\lambda_{\tau}\frac{1}{t-\tau+1}\left(\left(t+0.5\right)^{1.5}-\left(\tau-0.5\right)^{1.5}\right)\right)^{2}}\notag\\&\mkern-180mu -\frac{2\left(\sum_{\tau=1}^{t}\lambda_{\tau}\right)\left(\sum_{\tau=1}^{t}\lambda_{\tau}\left(t+\tau\right)\right)\left(\frac{1}{t-n+1}\left(\left(t+0.5\right)^{1.5}-\left(n-0.5\right)^{1.5}\right)\right)}{\left(\sum_{\tau=1}^{t}\lambda_{\tau}\frac{1}{t-\tau+1}\left(\left(t+0.5\right)^{1.5}-\left(\tau-0.5\right)^{1.5}\right)\right)^{3}}.\notag
    \end{align}
    \end{tiny}
    Setting this equal to 0 and rearranging,
    \begin{align}
        \frac{\partial f(\lambda,t)}{\partial \lambda_n} = 0 \notag
    \end{align}
    \begin{align}
        \frac{\left(\sum_{\tau=1}^{t}\frac{\lambda_{\tau}}{\sum_{\tau'=1}^{t}\lambda_{\tau'}}\tau\right)+\left(2t+n\right)}{\left(\frac{1}{t-n+1}\left(\left(t+0.5\right)^{1.5}-\left(n-0.5\right)^{1.5}\right)\right)}=\frac{2\left(\sum_{\tau=1}^{t}\lambda_{\tau}\left(t+\tau\right)\right)}{\left(\sum_{\tau=1}^{t}\frac{\lambda_{\tau}}{t-\tau+1}\left(\left(t+0.5\right)^{1.5}-\left(\tau-0.5\right)^{1.5}\right)\right)}.
        \label{line:36}
    \end{align}
    Note the right hand side does not depend on $n$, the index of the partial derivative. Thus at a critical point, the left hand side will equal the same value for all $n$. We can define this left side as a function $g(n, t, \lambda)$.

    To simplify, we make the following substitutions:
    \begin{align}
        c_1(t,\lambda) = \left(\sum_{\tau=1}^{t}\frac{\lambda_{\tau}}{\sum_{\tau'=1}^{t}\lambda_{\tau'}}\tau\right)-0.5\notag\\
        &\mkern-274mu u(n) = \sqrt{n-0.5}\notag\\
        &\mkern-274mu w(t) = \sqrt{t+0.5}\notag
    \end{align}
    Since $n \ge 1$, $t \ge 1$, and $n \le t$, we can note that $u(n) > 0$, $w(t) > 0$, $u(n) < w(t)$, and $0.5 \le c_1(t,\lambda) \le t-0.5$. This alleviates concerns over dividing by zero going forward.
    
    Applying these to $g(n, t, \lambda)$,
    \begin{align}
        g(n, t, \lambda) = \frac{c_1(t,\lambda)+2w(t)^2+u(n)^2}{\frac{w(t)^3-u(n)^3}{w(t)^2-u(n)^2}} = \frac{\left(w(t)+u(n)\right)\left(c_1(t,\lambda)+2w(t)^{2}+u(n)^{2}\right)}{w(t)^{2}+w(t)u(n)+u(n)^{2}}.\notag
    \end{align}
    Next we can consider the partial derivative over $n$ and set it equal to zero, noting that $u'(n) = \frac{1}{2u(n)}$,
    \begin{align}
        \frac{\partial g(n,t,\lambda)}{\partial n} = \frac{u(n)(u(n)^{3}+2w(t)u(n)^{2}+(2w(t)^{2}-c_1(t,\lambda))u(n)-2w(t)(w(t)^{2}+c_1(t,\lambda))}{2u\left(u(n)^2+w(t)u(n)+w(t)^2\right)^2}\notag\\
        u(n)^{3}+2w(t)u(n)^{2}+(2w(t)^{2}-c_1(t,\lambda))u(n)-2w(t)(w(t)^{2}+c_1(t,\lambda)) = 0.\notag
    \end{align}
    This is a cubic equation over $u(n)$. Looking at the coefficients, 1 and $2w(t)$ are both positive. Making a substitution and using a bound on $c_1$,
    $2w(t)^2-c_1(t,\lambda) \ge 2t+1-t+0.5 = t+1.5 > 0$. Clearly the last term is negative. By Descartes' rule of signs, the coefficients of this polynomial have 1 sign change and thus at most 1 positive real root over $u(n)$.
    Since $u(n)$ is monotonic over $n$ and positive for all $n\ge1$, we see that there exists at most one $n$ such that $\frac{\partial g(n,t,\lambda)}{\partial n} = 0$, for any given $t$ and $\lambda$. From this we can deduce that the equation $g(n,t,\lambda) = h(t,\lambda)$ for any function $h(t,\lambda)$ that does not depend on $n$ has at most 2 solutions over $n$. 

    Applying this to Eq. \ref{line:36} and noting that the left hand side is equal to $g(n,t,\lambda)$, we can conclude that this equation can have no more than 2 solutions over $n$. Thus the partial derivative of $f(\lambda, t)$ with respect to $\lambda$ can have at most 2 components with value 0 at the same time. We can also observe $f(c\lambda,t) = f(\lambda,t)$ for all $c \in \mathbb{R}^+$, thus the supremum must occur at some finite point. Since we're optimizing over the positive orthant, that finite point must have all non-zero components with gradient zero. Thus there can be at most 2 non-zero components of $\lambda$ at the supremum. Continuing from Eq. \ref{line:34}, we now have the upper bound,

    \begin{tiny}
    \AddToHook{normalfont}[normalsize]{\normalsize}
    \begin{align}
        &\mkern-800mu\sup_{\lambda \in \mathbb{R}_{\ge 0}^t} \frac{9}{8}\frac{\left(\sum_{\tau=1}^t \lambda_{\tau}\right)\left(\sum_{\tau=1}^t \lambda_{\tau}(t+\tau)\right)}{\left(\sum_{\tau=1}^t  \frac{\lambda_{\tau}}{t-\tau+1} \left((t+0.5)^{1.5}-(\tau-0.5)^{1.5}\right)\right)^2} \notag\\= 
        \sup_{\lambda \in \mathbb{R}_{\ge 0}^t, 1\le i < j \le t} \frac{9}{8}\frac{\left(\lambda_{i}+\lambda_j\right)\left(\lambda_i(t+i)+\lambda_j(t+j)\right)}{\left(\frac{\lambda_{i}}{t-i+1} ((t+0.5)^{1.5}-(i-0.5)^{1.5}) + \frac{\lambda_{j}}{t-j+1} ((t+0.5)^{1.5}-(j-0.5)^{1.5})\right)^2}.
        \label{line:43}
    \end{align}
    \end{tiny}
    We will denote this function inside the supremum as $z$. We can replace $\lambda_i$ and $\lambda_j$ with variables $x$, $y$ and make similar substitutions as before,
    \begin{align}
        u(i) = \sqrt{i-0.5}\notag\\
        v(j) = \sqrt{j-0.5}\notag\\
        w(t) = \sqrt{t+0.5}\notag
    \end{align}
    For more compact notation we will ignore arguments going forward. Making these substitutions yields,
    \begin{align}
        z(x,y,i,j,t) = \frac{9}{8}\frac{(x+y)(x(w^2+u^2)+y(w^2+v^2))}{\left(\frac{x(w^3-u^3)}{w^2-u^2} + \frac{y(w^3-v^3)}{w^2-v^2}\right)^2}.\notag
    \end{align}

    Using a computer algebra system, we can expand this as,
    \begin{align}
        &\mkern-390mu z(x,y,i,j,t) = \frac{9}{8} \left(1 - \frac{p}{q}\right)\notag\\
        p = -u^4 v^2 x y-2 u^4 v w x y-u^4 w^2 x y+2 u^3 v^3 x y+2 u^3 v^2 w x y-u^2 v^4 x y+2 u^2 v^3 w x y+\notag\\u^2 v^2 w^2 x^2+4 u^2 v^2 w^2 x y+u^2 v^2 w^2 y^2+2 u^2 v w^3 x^2+2 u^2 v w^3 x y+u^2 w^4 x^2+\notag\\u^2 w^4 x y-2 u v^4 w x y+2 u v^2 w^3 x y+2 u v^2 w^3 y^2-v^4 w^2 x y+v^2 w^4 x y+v^2 w^4 y^2
        \notag\\&\mkern-625mu
        q = \left(x(w^2+wu+u^2)(w+v) + y(w^2+wv+v^2)(w+u)\right)^2.\notag
    \end{align}
    By definition $x,y \ge 0$ and $1\le i < j \le t$, so $u,w,v > 0$, $u < w$, and $v < w$. Thus we can see that $q > 0$. Rearranging terms,
    \begin{align}
        p = u^{3}v^{2}\left(2w-u\right)xy+2u^{2}vw\left(w^{2}-u^{2}\right)xy+u^{2}w^{2}\left(w^{2}-u^{2}\right)xy\notag\\+u^{2}v^{2}\left(4w^{2}-v^{2}\right)xy+2uv^{2}w\left(w^{2}-v^{2}\right)xy+v^{2}w^{2}\left(w^{2}-v^{2}\right)xy\notag\\+2u^{3}v^{3}xy+2u^{2}v^{3}wxy+u^{2}v^{2}w^{2}x^{2}+u^{2}v^{2}w^{2}y^{2}+2u^{2}vw^{3}x^{2}\notag\\+u^{2}w^{4}x^{2}+2uv^{2}w^{3}y^{2}+v^{2}w^{4}y^{2}.\notag
    \end{align}
    Since $w>u$ and $w>v$ we can see that $p$ is a sum of positive terms so $p>0$. Putting this together we have $\frac{p}{q} > 0$. Therefore continuing from Eq. \ref{line:43},
    \begin{align}
        = \sup_{t\ge 1, x,y \ge 0, 1\le i < j \le t} z(x,y,i,j,t) = \sup_{t\ge 1, x,y \ge 0, 1\le i < j \le t} \frac{9}{8}(1-\frac{p}{q}) \le \frac{9}{8}.\notag
    \end{align}
    This proves our bound. In conclusion, we first show that the variance ratio is bounded by another ratio involving 2 weighted averages. We then use the fact that the weights are non-decreasing to transform the expressions into a weighted average of unweighted averages. Using closed forms and approximations we find a simpler upper bound. We consider the supremum of this expression over all possible weights, and observe that the maximum can only occur with at most 2 non-zero weights. Finally we analyze this 2 weight case and conclude it is bounded by $\frac{9}{8}$. Thus we have,
    \begin{align}
        \sup_{t\ge 1}\sup_{y\in[0,1]}\frac{\Var[\hat{F}^{\mathrm{comb}}_{1:t}]}{\Var_\mathrm{opt}(y,t)} \le \frac{9}{8},\notag
    \end{align}
\end{proof}

\subsection{Proof of Proposition 5 (martingale CLT)}
\label{sec:martingale_clt}
To construct the formal martingale CLT, we need to let the number of active measurement steps $T$ and the domain size $N$ go to infinity jointly. Suppose that for each $N$, we set the number of active measurement steps to be $T_N$ ($T_N<N$) where $T_N$ is non-decreasing with $N$. By letting $N\to\infty$ and $T_N\to\infty$, we have a martingale array $S_{N,t}=\sum_{\tau=1}^tX_{N,\tau}=\sum_{\tau=1}^t\al_\tau (\hat{F}_\tau-F(\Omega))$, for $t\le T_N$. Here $\al_\tau=\alpha_\tau/\sqrt{\sum_{\tau'=1}^{T_N}\alpha_{\tau'}^2\Var[\hat{F}_{\tau'}]}$ is normalized using total variances. It is direct to check that $\E[S_{N,t+1}-S_{N,t}|S_{N,t}]=\E[X_{N,t+1}|S_{N,t}]=0$. With addition assumptions, we are able to construct a limiting theorem for active measurement.

\begin{proposition}[formal]
     Assume that (1) $\sum_{\tau=1}^{T_N}\al_\tau^2\Var[\hat{F}_\tau|\D_\tau]\to \eta^2$ as $N\to\infty$; (2) $P(\eta^2<\infty)=1$ and $P(\eta^2>0)=1$; (3) $\sum_{\tau=1}^{T_N}\E[X_{N,\tau}^2\mathbb{I}(|X_{N,\tau}|>\epsilon)|\D_\tau]\xrightarrow{p} 0$ for any $\epsilon>0$. We have two central limit theorems
     \begin{gather}
     \label{eq:var_u}
         \frac{S_{N,T_N}}{\sqrt{\sum_{\tau=1}^{T_N}X_{N,\tau}^2}}\xrightarrow{D} \mathcal{N}(0,1),\\
         \label{eq:var_v}
         \frac{S_{N,T_N}}{\sqrt{\sum_{\tau=1}^{T_N}\al_\tau^2\Var[\hat{F}_\tau|\D_\tau]}}\xrightarrow{D} \mathcal{N}(0,1).
     \end{gather}
\end{proposition}
\begin{proof}
    We have met all the conditions for Theorem 3.3 and Corollary 3.2 in \citet{hall2014martingale}. Then we directly have
    \begin{align}
        \frac{S_{N,T_N}}{U_{N,T_N}}\xrightarrow{D} \mathcal{N}(0,1),\notag\\
         \frac{S_{N,T_N}}{V_{N,T_N}}\xrightarrow{D} \mathcal{N}(0,1),\notag
    \end{align}
    where
    \begin{align}
        U_{N,T_N}&=\sum_{\tau=1}^{T_N}X_{N,\tau}^2\notag\\
        V_{N,T_N}&=\sum_{\tau=1}^{T_N}\al_\tau^2\Var[\hat{F}_\tau|\D_\tau].\notag
    \end{align}
\end{proof}
The formal proposition employs a different normalizer from active measurement. It is useful for bounding the discrepancy between conditional and total variances. For active measurement, we have the following corollary. 
\begin{corollary}
    With the same conditions as Prop. \ref{prop:martingale_clt}, we also have that 
    \begin{gather}
    \label{eq:var_simp_mar}
         \frac{\hat{F}_{1:T_N}-F(\Omega)}{\sqrt{\sum_{\tau=1}^{T_N}\alp_\tau^2 (\hat{F}_\tau-F(\Omega))^2}}\xrightarrow{D} \mathcal{N}(0,1),\\
             \label{eq:var_cond_mar}
         \frac{\hat{F}_{1:T_N}-F(\Omega)}{\sqrt{\sum_{\tau=1}^{T_N}\alp_\tau^2\Var[\hat{F}_\tau|\D_\tau]}}\xrightarrow{D} \mathcal{N}(0,1).
     \end{gather}
\end{corollary}
\begin{proof}
    Basically the equations are the same as Eq. \ref{eq:var_u} and \ref{eq:var_v}, but with different normalizers. After multiplying $\frac{\sqrt{\sum_{\tau'=1}^{T_N}\alpha_{\tau'}^2\Var[\hat{F}_{\tau'}]}}{\sum_{\tau'=1}^{T_N}\alpha_{\tau'}}$ to both the numerator and denominator, we get Eq. \ref{eq:var_simp_mar} and \ref{eq:var_cond_mar}.
\end{proof}

This leads to our informal version of the martingale CLT. Both $\Var^{\mathrm{cond}}_{1:t}$ and $\Var^{\mathrm{simp}}_{1:t}$ can be constructed using the corollary.

We also demonstrate our assumptions in Prop. \ref{prop:martingale_clt} with a simple example. If $\Var[\hat{F}_\tau]=\sigma_\tau^2=\lambda_1\tau^{-y}/w_\tau$ and $\Var[\hat{F}_\tau|\D_\tau]=\lambda_2\tau^{-y}/w_\tau$ where $y\in[0,1]$ and $0<\lambda_2\le\lambda_1$, then with COMB weights
\begin{align}
    \sum_{\tau=1}^{T_N}\al_\tau^2\Var[\hat{F}_\tau|\D_\tau]&=\frac{\sum_{\tau=1}^{T_N}\lambda_2w_\tau\tau^{1-y}}{\sum_{\tau=1}^{T_N}\lambda_1w_\tau\tau^{1-y}}\notag\\
    &=\frac{\lambda_2}{\lambda_1}.\notag
\end{align}
It is clear that $0<\lambda_2/\lambda_1<\infty$, so the first two assumptions are met. For the third assumption, let $T_N=N/2$ and further assume for some $\delta>0$, $\sup_{\tau,N}\frac{\E[(\hat{F}_\tau-F(\Omega))^{2+\delta}|\D_\tau]}{\sigma_\tau^{2+\delta}}<\infty$. The additional assumption implies the uniform boundedness of $\frac{(\hat{F}_\tau-F(\Omega))^2}{\sigma_\tau^2}$, which is common in the literature~\citep{zhou1998adaptive}. Then note that
\begin{align}
\label{eq:clt_example1}
    \frac{\max_{\tau}\alpha_\tau^2\sigma_\tau^2}{\sum_{\tau=1}^{T_N}\alpha_{\tau}^2\sigma_{\tau}^2}&=\frac{\lambda_1w_{T_N}T_N^{1-y}}{\sum_{\tau=1}^{T_N}\lambda_1w_\tau\tau^{1-y}}\notag\\
    &\le \frac{\lambda_1w_{T_N}T_N^{1-y}}{\frac{\lambda_1}{T_N}\left(\sum_{\tau=1}^{T_N}w_\tau\right)\left(\sum_{\tau=1}^{T_N}\tau^{1-y}\right)}\notag\\
    &=\frac{NT_N^{1-y}}{(N-T_N+1)\left(\sum_{\tau=1}^{T_N}\tau^{1-y}\right)}\notag\\
    &\le \frac{NT_N^{1-y}}{(N-T_N+1)T_N^{2-y}/(2-y)}\to 0.
\end{align}
Therefore, for any $\tau$,
\begin{align}
    \E[X_{N,\tau}^2\mathbb{I}(|X_{N,\tau}|>\epsilon)|\D_\tau]&=\E\left[\frac{\alpha_{\tau}^2(\hat{F}_\tau-F(\Omega))^2}{\sum_{\tau'=1}^{T_N}\alpha_{\tau'}^2\sigma_{\tau'}^2}\mathbb{I}(|X_{N,\tau'}|>\epsilon)|\D_{\tau}\right]\notag\\
    &=\frac{\alpha_{\tau}^2\sigma_\tau^2}{\sum_{\tau'=1}^{T_N}\alpha_{\tau'}^2\sigma_{\tau'}^2}\E\left[\frac{(\hat{F}_\tau-F(\Omega))^2}{\sigma_\tau^2}\mathbb{I}(|X_{N,\tau'}|>\epsilon)|\D_{\tau}\right].\notag
\end{align}
Note that 
\begin{align}
    \E\left[\frac{(\hat{F}_\tau-F(\Omega))^2}{\sigma_\tau^2}\mathbb{I}(|X_{N,\tau'}|>\epsilon)|\D_{\tau}\right]&=\E\left[\frac{(\hat{F}_\tau-F(\Omega))^2}{\sigma_\tau^2}\mathbb{I}\left(\frac{|\hat{F}_\tau-F(\Omega)|}{\sigma_\tau}>\epsilon\frac{\sqrt{\sum_{\tau'=1}^{T_N}\alpha_{\tau'}^2\sigma_{\tau'}^2}}{\alpha_\tau\sigma_\tau}\right)|\D_{\tau}\right].\notag
\end{align}
Observe that $\frac{\E[(\hat{F}_\tau-F(\Omega))^2|\D_\tau]}{\sigma_\tau^2}=\frac{\lambda_2}{\lambda_1}$, but $\frac{\sqrt{\sum_{\tau'=1}^{T_N}\alpha_{\tau'}^2\sigma_{\tau'}^2}}{\alpha_\tau\sigma_\tau}\to\infty$ for any $\tau$ according to Eq. \ref{eq:clt_example1}. By the uniform boundedness, then for any $\epsilon_0>0$, there exists $N_0$ such that for any $N> N_0$, 
\begin{align}
    \E\left[\frac{(\hat{F}_\tau-F(\Omega))^2}{\sigma_\tau^2}\mathbb{I}(|X_{N,\tau'}|>\epsilon)|\D_{\tau}\right]<\epsilon_0.\notag
\end{align}
Therefore,
\begin{align}
    \sum_{\tau=1}^{T_N}\E[X_{N,\tau}^2\mathbb{I}(|X_{N,\tau}|>\epsilon)|\D_\tau]&<\sum_{\tau=1}^{T_N}\frac{\alpha_{\tau}^2\sigma_\tau^2}{\sum_{\tau'=1}^{T_N}\alpha_{\tau'}^2\sigma_{\tau'}^2}\epsilon_0\notag\\
    &=\epsilon_0.\notag
\end{align}
This shows that the third assumption is also met.

\subsection{Proof of proposition 6}
\begin{proposition}
    The estimators $\smwidehat \Var_\tau$ and $\smwidehat \Var_{\tau, r}$ for $\tau \leq r \leq t$ satisfy
    $\E[\smwidehat \Var_{\tau} | \D_\tau] = \E[\widehat \Var_{\tau, r} | \D_\tau] = \Var[\hat{F}_\tau | \D_\tau]$ and $\E[\widehat \Var_{\tau}] = \E[\smwidehat \Var_{\tau, r}] = \Var[\hat{F}_\tau]$.
\end{proposition}
\begin{proof}
We first show the unbiasedness of individual variance estimators $\smwidehat{\Var}_{\tau,r}$.
    \begin{align}
        &\E[\smwidehat{\Var}_{\tau,r}|\D_\tau]\notag\\
        =&\E_{\D_{r}\backslash\D_{\tau},s_r}\left[\sum_{s\in\D_{r}\backslash\D_{\tau}}q_{\tau}(s)\left(\frac{f(s)}{q_\tau(s)}-F(\Omega\backslash \D_{\tau})\right)^2+\frac{q_\tau(s_{r})}{q_r(s_{r})} \left(\frac{f(s_{r})}{q_\tau(s_{r})}-F(\Omega\backslash \D_{\tau})\right)^2|\D_\tau\right]\notag\\
        =&\E_{\D_{r}\backslash\D_{\tau}}\left[\sum_{s\in\D_{r}\backslash\D_{\tau}}q_{\tau}(s)\left(\frac{f(s)}{q_\tau(s)}-F(\Omega\backslash \D_{\tau})\right)^2+\E_{s_r}\left[\frac{q_\tau(s_{r})}{q_r(s_{r})} \left(\frac{f(s_{r})}{q_\tau(s_{r})}-F(\Omega\backslash \D_{\tau})\right)^2\right]|\D_\tau\right]\notag\\
        =&\E_{\D_{r}\backslash\D_{\tau}}\left[\sum_{s\in\D_{r}\backslash\D_{\tau}}q_{\tau}(s)\left(\frac{f(s)}{q_\tau(s)}-F(\Omega\backslash \D_{\tau})\right)^2+\sum_{s_r\in\Omega\backslash\D_r}q_\tau(s_{r}) \left(\frac{f(s_{r})}{q_\tau(s_{r})}-F(\Omega\backslash \D_{\tau})\right)^2|\D_\tau\right]\notag\\
        =&\E_{\D_{r}\backslash\D_{\tau}}\left[\sum_{s\in\Omega\backslash\D_{\tau}}q_{\tau}(s)\left(\frac{f(s)}{q_\tau(s)}-F(\Omega\backslash \D_{\tau})\right)^2|\D_\tau\right]\notag\\
        =&\Var[\hat{F}_\tau|\D_\tau].\notag
    \end{align}
    Also, note that
    \begin{align}
        \Var_{\D_\tau}[\E[\hat{F}_\tau|\D_\tau]]&=\Var_{\D_\tau}\left[\E\left[F(\D_\tau)+\frac{f(s_\tau)}{q(s_\tau)}|\D_\tau\right]\right]\notag\\
        &=\Var_{\D_\tau}\left[\sum_{s_\tau\in\Omega\backslash\D_\tau}q(s_\tau)\left(F(\D_\tau)+\frac{f(s_\tau)}{q(s_\tau)}\right)\right]\notag\\
        &=\Var_{\D_\tau}\left[F(\D_\tau)+\sum_{s_\tau\in\Omega\backslash\D_\tau}q(s_\tau)\frac{f(s_\tau)}{q(s_\tau)}\right]\notag\\
        &=\Var_{\D_\tau}\left[F(\Omega)\right]\notag\\
        &=0.\notag
    \end{align}
    Therefore, we further have
    \begin{align}
    \E[\E[\smwidehat{\Var}_{\tau,r}|\D_\tau]]&=
        \E_{\D_\tau}[\Var[\hat{F}_\tau|\D_\tau]]\notag\\
        &=\Var[\hat{F}_\tau]-\Var_{\D_\tau}[\E[\hat{F}_\tau|\D_\tau]]\notag\\
        &=\Var[\hat{F}_\tau].\notag
    \end{align}
    Since $\widehat \Var_{\tau}$ is convex on $\widehat \Var_{\tau,r}$ with the weights, we also have $\E[\smwidehat \Var_{\tau} | \D_\tau] =  \Var[\hat{F}_\tau | \D_\tau]$ and $\E[\widehat \Var_{\tau}] =  \Var[\hat{F}_\tau]$.
\end{proof}

\subsection{Proof of proposition 7}
\begin{proposition}
    With the same settings as Prop. \ref{prop:var}, the variance estimate for $\hat{F}_\tau$ weighted by the LURE scheme $\smwidehat{\Var}_{\tau}^{\mathrm{LURE}}=\sum_{r=\tau}^t\bet_{r}^{\mathrm{LURE}}\smwidehat{\Var}_{\tau,r}$ satisfies that $\Var\left[\smwidehat{\Var}_{\tau}^{\mathrm{LURE}}|\D_\tau\right]\lesssim \frac{1}{t-\tau+1}\cdot\min\left(1,\frac{(N-t)^2}{t-\tau+1}\right)$.
\end{proposition}

\begin{proof}
We first derive the normalized LURE weights.
    \begin{align}
        \sum_{r=\tau}^t\beta_{r}^{\mathrm{LURE}}&=\sum_{r=\tau}^t\frac{1}{(N-r)(N-r+1)}\notag\\
        &=\sum_{r=\tau}^t\frac{1}{N-r}-\frac{1}{N-r+1}\notag\\
        &=\frac{1}{N-t}-\frac{1}{N-\tau+1}\notag\\
        &=\frac{t-\tau+1}{(N-t)(N-\tau+1)}.\notag
    \end{align}

    So the normalized LURE weights satisfy
    \begin{align}
        \bet_r^{\mathrm{LURE}}=\frac{(N-t)(N-\tau+1)}{(t-\tau+1)(N-r)(N-r+1)}.\notag
    \end{align}
    We then bound the variance of individual variance estimators. Observe that
    \begin{align}
        \Var[\smwidehat{\Var}_{\tau,r}|\D_\tau]&= \Var[\E[\smwidehat{\Var}_{\tau,r}|D_r]|\D_\tau]+ \E[\Var[\smwidehat{\Var}_{\tau,r}|D_r]|\D_\tau]\notag\\
        &=\Var[\Var[\hat{F}_\tau]|\D_\tau]+ \E[\Var[\smwidehat{\Var}_{\tau,r}|D_r]|\D_\tau]\notag\\
        &= 0+\E\left[\Var\left[\frac{q_\tau(s_{r})}{q_r(s_{r})} \left(\frac{f(s_{r})}{q_\tau(s_{r})}-F(\Omega\backslash \D_{\tau})\right)^2|D_r\right]|\D_\tau\right]\notag\\
        &\le \E\left[\E\left[\frac{q_\tau(s_{r})^2}{q_r(s_{r})^2} \left(\frac{f(s_{r})}{q_\tau(s_{r})}-F(\Omega\backslash \D_{\tau})\right)^4|D_r\right]|\D_\tau\right]\notag\\
        &\le \E\left[\E\left[\frac{q_\tau(s_{r})^2}{q_r(s_{r})^2} \cdot 8\cdot\left(\frac{f(s_{r})^4}{q_\tau(s_{r})^4}+F(\Omega\backslash \D_{\tau})^4\right)|D_r\right]|\D_\tau\right]\notag\\
        &\le \E\left[\E\left[\frac{B^4(N-r+1)^2}{A^4(N-\tau+1)^2} \cdot 8\cdot\left(\frac{B^4(N-\tau+1)^4B^4}{A^4}+B^4(N-\tau+1)^4\right)|D_r\right]|\D_\tau\right]\notag\\
        &=\frac{8B^8(B^4+A^4)}{A^8}(N-r+1)^2(N-\tau+1)^2.\notag
    \end{align}
    Therefore, let $D=\frac{8B^8(B^4+A^4)}{A^8}$, we can say that $\Var[\smwidehat{\Var}_{\tau,r}|\D_\tau]\le D(N-\tau+1)^2(N-r+1)^2$.
 Next, we show that for every two variance estimators $\Cov(\smwidehat{\Var}_{\tau,r_1},\smwidehat{\Var}_{\tau,r_2}|\D_\tau)=0$ if $\tau\le r_1<r_2\le N$.
    \begin{align}
        \Cov(\smwidehat{\Var}_{\tau,r_1},\smwidehat{\Var}_{\tau,r_2}|\D_\tau)&=\E[(\smwidehat{\Var}_{\tau,r_1}-\Var_{s_\tau}[\hat{F}_\tau|\D_\tau])(\smwidehat{\Var}_{\tau,r_2}-\Var_{s_\tau}[\hat{F}_\tau|\D_\tau])|\D_\tau]\notag\\
        &=\E_{\D_{r_2}\backslash\D_\tau}[(\smwidehat{\Var}_{\tau,r_1}-\Var_{s_\tau}[\hat{F}_\tau|\D_\tau])\E_{s_{r_2}}[(\smwidehat{\Var}_{\tau,r_2}-\Var_{s_\tau}[\hat{F}_\tau|\D_\tau])]]\notag\\
        &=\E_{\D_{r_2}\backslash\D_\tau}[(\smwidehat{\Var}_{\tau,r_1}-\Var_{s_\tau}[\hat{F}_\tau|\D_\tau])\cdot 0]\notag\\
        &=0.\notag
    \end{align}
    
    Therefore (when $t<N$),
        \begin{align}
        \Var\left[\smwidehat{\Var}_{\tau}^{\mathrm{LURE}}|\D_\tau\right]&=\sum_{r=\tau}^t\frac{(N-t)^2(N-\tau+1)^2}{(t-\tau+1)^2(N-r)^2(N-r+1)^2}\Var[\smwidehat{\Var}_{\tau,r}]\notag\\
        &\le\sum_{r=\tau}^t\frac{(N-t)^2(N-\tau+1)^2}{(t-\tau+1)^2(N-r)^2(N-r+1)^2}D(N-\tau+1)^2(N-r+1)^2\notag\\
        &=\frac{D(N-t)^2(N-\tau+1)^4}{(t-\tau+1)^2}\sum_{r=\tau}^t\frac{1}{(N-r)^2}.\notag
    \end{align}
    There are two ways to bound the right hand side. When $t$ is far from $N$, we can bound with
    \begin{align}
        \Var\left[\smwidehat{\Var}_{\tau}^{\mathrm{LURE}}|\D_\tau\right]&\le \frac{D(N-t)^2(N-\tau+1)^4}{(t-\tau+1)^2}\sum_{r=\tau}^t\frac{1}{(N-r)^2}\notag\\
        &\le \frac{D(N-t)^2(N-\tau+1)^4}{(t-\tau+1)^2}\sum_{r=\tau}^t\frac{1}{(N-t)^2}\notag\\
        &=\frac{D(N-\tau+1)^4}{t-\tau+1}.\notag
    \end{align}
    When $t$ is close to $N$, by the fact that $\sum_{x=1}^\infty\frac{1}{x^2}=\frac{\pi^2}{6}$, we have
        \begin{align}
        \Var\left[\smwidehat{\Var}_{\tau}^{\mathrm{LURE}}|\D_\tau\right]&\le \frac{D(N-t)^2(N-\tau+1)^4}{(t-\tau+1)^2}\sum_{r=\tau}^t\frac{1}{(N-r)^2}\notag\\
        &\le \frac{D(N-\tau+1)^4}{t-\tau+1}\cdot\frac{\pi^2(N-t)^2}{6(t-\tau+1)}.\notag
    \end{align}
    The above two inequalities still hold when $t=N$, where LURE will assign infinite weights on the exact variance $\smwidehat{\Var}_{\tau,N}=\Var[\hat{F}_\tau|D_\tau]$, so $\Var[\smwidehat{\Var}^{\mathrm{LURE}}_\tau|\D_\tau]=0$. In summary, we see that $\Var\left[\smwidehat{\Var}_{\tau}^{\mathrm{LURE}}|\D_\tau\right]\le\frac{E}{t-\tau+1}\cdot\min\left(1,\frac{(N-t)^2}{(t-\tau+1)}\right)$, for $E=\frac{\pi^2D(N-\tau+1)^2}{6}$. Thus $\Var\left[\smwidehat{\Var}_{\tau}^{\mathrm{LURE}}|\D_\tau\right]\lesssim \frac{1}{t-\tau+1}\cdot\min\left(1,\frac{(N-t)^2}{(t-\tau+1)}\right)$.
    
\end{proof}
\section{Details of variance estimation} 
\subsection{The plug-in mean in variance estimation}
One practical gap is the incorporation of a plug-in mean instead of the ground-truth mean in Eq. \ref{eq:var_estimate}. We show that the error introduced by this trick is controlled. First of all, the variance of the plug-in mean could be controlled with the following lemma.
\begin{lemma}
\label{lemma:lure}
    For $\hat{F}_{1:t}^{\mathrm{LURE}}=\sum_{\tau=1}^t\alp^{\mathrm{LURE}}_{\tau}\hat{F}_{\tau}$, its variance satisfies that there exists a constant $E>0$, $\Var[\hat{F}_{1:t}^{\mathrm{LURE}}]\le \frac{E(N-t)}{t}$.
\end{lemma}
\begin{proof}
    By Prop. \ref{prop:var},
    \begin{align}
        \Var[\hat{F}_\tau]\le C(N-\tau)(N-\tau+1).\notag
    \end{align}
    Then 
    \begin{align}
        \Var[\hat{F}_{1:t}^{\mathrm{LURE}}]&=\sum_{\tau=1}^t\left(\alp^{\mathrm{LURE}}_{\tau}\right)^2\Var[\hat{F}_\tau]\notag\\
        &\le \sum_{\tau=1}^t\frac{N^2(N-t)^2}{t^2(N-\tau)^2(N-\tau+1)^2}\cdot C(N-\tau)(N-\tau+1)\notag\\
        &=\frac{CN^2(N-t)^2}{t^2}\sum_{\tau=1}^t\frac{1}{(N-\tau)(N-\tau+1)}\notag\\
        &=\frac{CN^2(N-t)^2}{t^2}\frac{t}{N(N-t)}\notag\\
        &=\frac{CN(N-t)}{t}.\notag
    \end{align}
    Let $E=CN$, then $\Var[\hat{F}_{1:t}^{\mathrm{LURE}}]\le \frac{E(N-t)}{t}$.
\end{proof}
This directly implies that the plug-in mean constructed from the LURE estimate will not be far from $F(\Omega)$. Denote the variance estimate with the plug-in mean by 
\begin{align}
    \widetilde{\Var}_{\tau,t}=&\sum_{s\in\D_{t}\backslash\D_{\tau}}q_{\tau}(s)\left(\frac{f(s)}{q_\tau(s)}-\hat{G}_{\tau,t}\right)^2+\frac{q_\tau(s_{t})}{q_t(s_{t})} \left(\frac{f(s_{t})}{q_\tau(s_{t})}-\hat{G}_{\tau,t}\right)^2.\notag
\end{align}
We control the error with the following proposition.
\begin{proposition}
\label{prop:error_plugin}
    Given the same assumptions as Prop. \ref{prop:var}, with at least probability $1-p$, there exists a constant $H$ dependent on $p$ such that $|\widetilde{\Var}_{\tau,t}-\smwidehat{\Var}_{\tau,t}|\le H(N-\tau+1)\sqrt{\frac{N-t+1}{t-1}}$.
\end{proposition}
\begin{proof}
First, note that
\begin{align}
    \hat{G}_{\tau,t}-F(\Omega\backslash\D_\tau)=\hat{F}_{1:t}^{\mathrm{LURE}}-F(\Omega).\notag
\end{align}
By Chebyshev's inequality, with at least probability $1-p$, 
\begin{align}
    |\hat{G}_{\tau,t}-F(\Omega\backslash\D_\tau)|=|\hat{F}_{1:t-1}^{\mathrm{LURE}}-F(\Omega)|\le \sqrt{\frac{E(N-t+1)}{p(t-1)}}.\notag
\end{align}
Also,
\begin{align}
     |\hat{G}_{\tau,t}+F(\Omega\backslash\D_\tau)|&\le |\hat{G}_{\tau,t}-F(\Omega\backslash\D_\tau)|+2F(\Omega\backslash\D_{\tau})\notag\\
     &\le \sqrt{\frac{E(N-t+1)}{p(t-1)}}+2B(N-\tau+1)\notag\\
     &\le I(N-\tau+1),\notag
\end{align}
where $I$ is a constant dependent on $p$.
    We expand the difference directly.
    \begin{align}
        |\widetilde{\Var}_{\tau,t}-\smwidehat{\Var}_{\tau,t}|=&\Bigg|\sum_{s\in\D_{t}\backslash\D_{\tau}}q_{\tau}(s)(\hat{G}_{\tau,t}^2-F(\Omega\backslash\D_\tau)^2)-2f(s)(\hat{G}_{\tau,t}-F(\Omega\backslash\D_\tau))\notag\\
        &+\frac{q_\tau(s_{t})}{q_t(s_{t})} \left(\hat{G}_{\tau,t}^2-F(\Omega\backslash\D_\tau)^2\right)-2\frac{f(s_t)}{q_t(s_t)}(\hat{G}_{\tau,t}-F(\Omega\backslash\D_\tau))\Bigg|\notag\\
        \le&\sum_{s\in\D_{t}\backslash\D_{\tau}}q_{\tau}(s)|\hat{G}_{\tau,t}^2-F(\Omega\backslash\D_\tau)^2|+2f(s)|\hat{G}_{\tau,t}+F(\Omega\backslash\D_\tau)|\notag\\
        &+\frac{q_\tau(s_{t})}{q_t(s_{t})} \left|\hat{G}_{\tau,t}^2-F(\Omega\backslash\D_\tau)^2\right|+2\frac{f(s_t)}{q_t(s_t)}|\hat{G}_{\tau,t}-F(\Omega\backslash\D_\tau)|\notag\\
        \le& \sqrt{\frac{E(N-t+1)}{p(t-1)}}\Bigg(\sum_{s\in\D_{t}\backslash\D_{\tau}}\frac{B}{(N-\tau+1)A}I(N-\tau+1)+2B\notag\\
        &+\frac{B^2(N-t+1)}{A^2(N-\tau+1)} I(N-\tau+1)+2\frac{B^2(N-t+1)}{A}\Bigg)\notag\\
        =&\left(\frac{IB}{A}+2B\right)\left(t-\tau+\frac{B}{A}(N-t+1)\right)\sqrt{\frac{E(N-t+1)}{p(t-1)}}\notag\\
        \le& \frac{(I+2A)B^2}{A^2}\sqrt{\frac{E}{p}}(N-\tau+1)\sqrt{\frac{N-t+1}{t-1}}.\notag
    \end{align}
    Let $H=\frac{(I+2A)B^2}{A^2}\sqrt{\frac{E}{p}}$, then $|\widetilde{\Var}_{\tau,t}-\smwidehat{\Var}_{\tau,t}|\le H(N-\tau+1)\sqrt{\frac{N-t+1}{t-1}}$.
\end{proof}
Comparing the derivation of Prop. \ref{prop:LURE_var} and Prop. \ref{prop:error_plugin}, the L1 error introduced by the plug-in mean goes to 0 at a faster rate than the standard deviation of $\smwidehat{\Var}_{\tau,t}$, indicating that the error is negligible in our evaluation when $t$ is large enough.

\subsection{The streaming algorithm of variance estimation}
\label{sec:streaming}

Alg. \ref{alg:variance_update} has a summation for each $1\le\tau\le t$, amounting to a complexity of $\mathcal{O}(t^2)$. In practice, we would reuse part of previous computation at step $t-1$ to accelerate the computation at step $t$. Observe that
\begin{align}
            \widetilde{\Var}_{\tau,t}&=\sum_{s\in\D_{t}\backslash\D_{\tau}}q_{\tau}(s)\left(\frac{f(s)}{q_\tau(s)}-\hat{G}_{\tau,t}\right)^2+\frac{q_\tau(s_{t})}{q_t(s_{t})} \left(\frac{f(s_{t})}{q_\tau(s_{t})}-\hat{G}_{\tau,t}\right)^2\notag\\
            &=\sum_{s\in\D_{t}\backslash\D_{\tau}}\left(\frac{f(s)^2}{q_\tau(s)}+q_\tau(s)\hat{G}_{\tau,t}^2-2f(s)\hat{G}_{\tau,t}\right)+\frac{q_\tau(s_{t})}{q_t(s_{t})} \left(\frac{f(s_{t})}{q_\tau(s_{t})}-\hat{G}_{\tau,t}\right)^2\notag\\
            &=\sum_{s\in\D_{t}\backslash\D_{\tau}}\frac{f(s)^2}{q_\tau(s)}+\hat{G}_{\tau,t}^2\sum_{s\in\D_{t}\backslash\D_{\tau}}q_\tau(s)-2\hat{G}_{\tau,t}\sum_{s\in\D_{t}\backslash\D_{\tau}}f(s)+\frac{q_\tau(s_{t})}{q_t(s_{t})} \left(\frac{f(s_{t})}{q_\tau(s_{t})}-\hat{G}_{\tau,t}\right)^2.\notag
            \end{align}
        This is a polynomial with respect to our plug-in mean $\hat{G}_{\tau,t}$. Therefore, we could maintain the three summations in the coefficients $\sum_{s\in\D_{t}\backslash\D_{\tau}}\frac{f(s)^2}{q_\tau(s)}$, $\sum_{s\in\D_{t}\backslash\D_{\tau}}q_\tau(s)$ and $\sum_{s\in\D_{t}\backslash\D_{\tau}}f(s)$ for each $\tau$ during the algorithm to speed-up the computation of each individual $\widetilde{\Var}_{\tau,t}$. In addition, our final estimate is
        \begin{align}
            \widetilde{\Var}_\tau=\sum_{r=\tau}^t\bet_r\widetilde{\Var}_{\tau,r}=\frac{\sum_{r=\tau}^t\beta_r\widetilde{\Var}_{\tau,r}}{\sum_{r=\tau}^t\beta_r}.\notag
        \end{align}
        
        The numerator is still a polynomial with respect to the plug-in mean whose coefficients can be written as summations. We would further maintain the three coefficients and $\sum_{r=\tau}^t\beta_r$ to avoid the loop to compute $\widehat{\Var}_\tau$. With the two techniques, our approach is summarized as Alg. \ref{alg:variance_update_streaming}.
        \begin{algorithm}[H]
            \small
        \caption{Streaming variance estimation update}
        \label{alg:variance_update_streaming}
		\begin{algorithmic}[1]
            \Require Sample $s_t$, label $f(s_t)$ and estimate $\hat{F}_{1:t-1}$
            \State Set $x_t=y_t=z_t=a_t=b_t=c_t=u_t=0$
		\For{$\tau= 1,2,...,t$} 
            \State Get estimated mean $\hat{G}_{\tau,t}=\hat{F}_{1:t-1}-F(\D_\tau)$
             \State $a_\tau \leftarrow a_\tau + \beta_t\left(x_\tau+\frac{f(s_t)^2}{q_t(s_t)q_\tau(s_t)}\right)$
             \State $b_\tau \leftarrow b_\tau + \beta_t\left(y_\tau+\frac{f(s_t)}{q_t(s_t)}\right)$
             \State $c_\tau \leftarrow c_\tau + \beta_t\left(z_\tau+\frac{q_\tau(s_t)}{q_t(s_t)}\right)$
            \State $u_\tau\leftarrow u_\tau + \beta_t$
            \State $\widetilde{\Var}_{\tau}=(a_{\tau}-2b_{\tau}\hat{G}_{\tau,t}+c_{\tau}\hat{G}^2_{\tau,t})/u_\tau$
            \State $x_{\tau}\leftarrow x_{\tau}+\frac{f(s_t)^2}{q_{\tau}(s_t)}$ 
            \State $y_{\tau}\leftarrow y_{\tau}+f(s_t)$
            \State $z_{\tau}\leftarrow z_{\tau}+q_\tau(s_t)$
			\EndFor 
		\end{algorithmic}
\end{algorithm}

\section{Experimental settings}
\subsection{Estimating birds in radar data}\label{appendix:roosts}
\paragraph{Weather radar preliminaries.} The NEXRAD radar network contains 159 high-resolution radars and covers most of the U.S. territories. Each radar station typically scans the surrounding atmosphere every 4-10 minutes to collect weather data. This is achieved by rotating the radar antenna around the vertical axis at multiple elevation angles. At each elevation, the radar performs ``sweeps'' to collect several radar products, such as, reflectivity and radial velocity. These signals also record objects like bird roosts. We render radar data for 300km $\times$ 300km regions into 600 $\times$ 600 pixel arrays, on which we train a detector model to automatically predict bounding boxes for roosts.

\begin{figure}[t]
    \centering
    \includegraphics[width=0.4\textwidth]{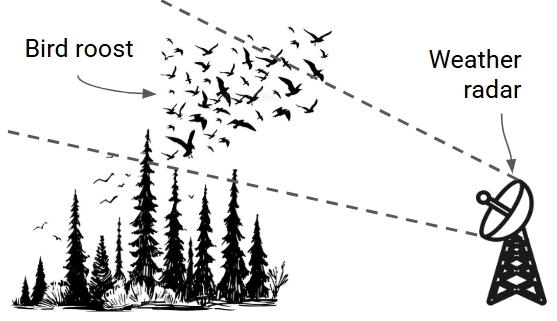}
    ~~~~~~~~~~
    \includegraphics[width=0.3\textwidth]{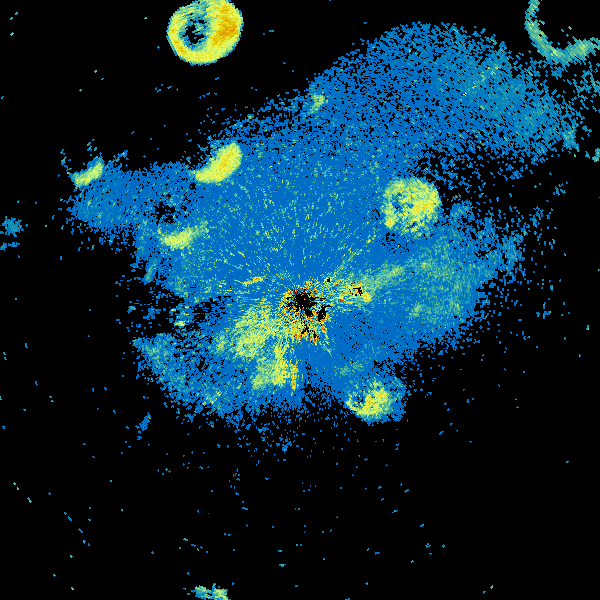}
    \caption{(Left) Weather radars can detect bird roosts departing from their nighttime roosting locations. (Right) Roosts appear as circular patterns in images rendered from radar products, such as the reflectivity at $0.5^\circ$ channel as visualized.}
    \label{fig:roosts}
\end{figure}

\paragraph{Visualization.} Fig.~\ref{fig:roosts} (left) illustrates weather radar scans capturing bird roosts. Fig.~\ref{fig:roosts} (right) visualizes the reflectivity data at $0.5\circ$ elevation of a radar scan that happened at the KCLE weather radar station on August 19 2015 at 10:31:22 UTC, which is a randomly sampled radar scan that captures bird roosts. The circular and semicircular patterns are massive amounts of birds departing from their overnight roosting locations.

\paragraph{Detector, tracking, and filtering.} We follow \citet{Perez2022.10.28.513761} to pretrain a station-agnostic spatiotemporal roost detector. We also follow their configuration for deploying the detector on the Great Lakes radar stations, which are excluded from the pretraining data, assembling model predicted boundings boxes into roost tracks, and filtering tracks with too few detected time frames or too low of a max or mean detection confidence score. 
Since each weather radar collects data every few minutes, the same roost is often captured by multiple consecutive scans of a radar station. To avoid double counting, it is necessary to assemble detections of the same roost into a track.

\citet{Perez2022.10.28.513761} deploys the roost predition system on radar data from 12 radar stations in the Great Lakes region and let human experts screen the system predictions. They focus on the time window from 30 minutes to 90 minutes after the local sunrise for every station-day. 

Following \citet{deng2023quantifying}, we focus on 11 stations, including KAPX, KBUF, KCLE, KDLH, KDTX, KGRB, KGRR, KLOT, KMKX, KTYX, KIWX, since KMQT does not clearly observe roosts in the human screened data.
Our goal is to estimate the number of birds in June to October in the 2015-2019 five-year period at these 11 stations. 
We use screened data from \citet{Perez2022.10.28.513761} as the ground truth bounding boxes for estimating bird counts. We let our station-specific finetuned checkpoints predict bounding boxes for the same time periods on the same station-day for our estimations.

\paragraph{Station-day bird count estimation.} 
\begin{itemize}
    \item Per-sweep counts. For each predicted bounding box by the detector model in a radar scan in a station-day, we enumerate over the radar sweeps at all elevation within 5000km height and follow \citet{belotti2023longterm} to estimate the bird count of this bounding box geographical region in this sweep.
    \item Per-bounding-box counts. We summarize, for each detection, radar sweeps taken across multiple elevations. In order to prevent double counting of birds in regions sampled twice by two consecutive beams, we bin the sweep elevations into $1\circ$ bins. We then take the average of sweeps that fall in the same bin. Lastly, we sum counts across all bins.
    \item Per-track counts. We obtain summaries across detections by finding the median count within each track and selecting two scans before and after the median. We then calculate the average count within this set of 5 scans.
    \item Per-day counts. We sum over per-track counts to obtain a per-day bird count at a station-day.
\end{itemize}

\subsection{Additional Finetuning And Sampling Details}
To fine-tune our models, we use the Detectron2 library with modified configurations. Specifically, we disable learning rate warmup, set the batch size to 8, and set FILTER\_EMPTY\_ANNOTATIONS to False. For the high-resolution image experiments, we use the faster\_rcnn\_R\_50\_FPN\_3x.yaml configuration with the default ImageNet-pretrained weights.

Fine-tuning the image detector takes approximately 20 minutes on a single NVIDIA A16 GPU. We perform fine-tuning 8 times, resulting in a total end-to-end runtime of just under 3 hours per image. For the roost detector, each fine-tuning run takes about 2 hours on a single A16 GPU. We fine-tune 4 times, totaling roughly 8 hours of compute per station for roost counting. In both cases, the time required to compute count and variance estimates is negligible.

In practice, the measurements from the detectors could be as low as $0$, making the labeling of some units impossible and introducing huge variances to the estimation. In our image counting tasks, we set $g(s)\leftarrow \max(g(s),1)$ to make sure that each tile can be sampled. In our radar counting tasks, we set $g(s)\to g(s)+1000$ to cover every day in every station. 

\subsection{Labeling Birds in Images}
\begin{figure}[t]
    \centering
    \includegraphics[width=0.49\textwidth]{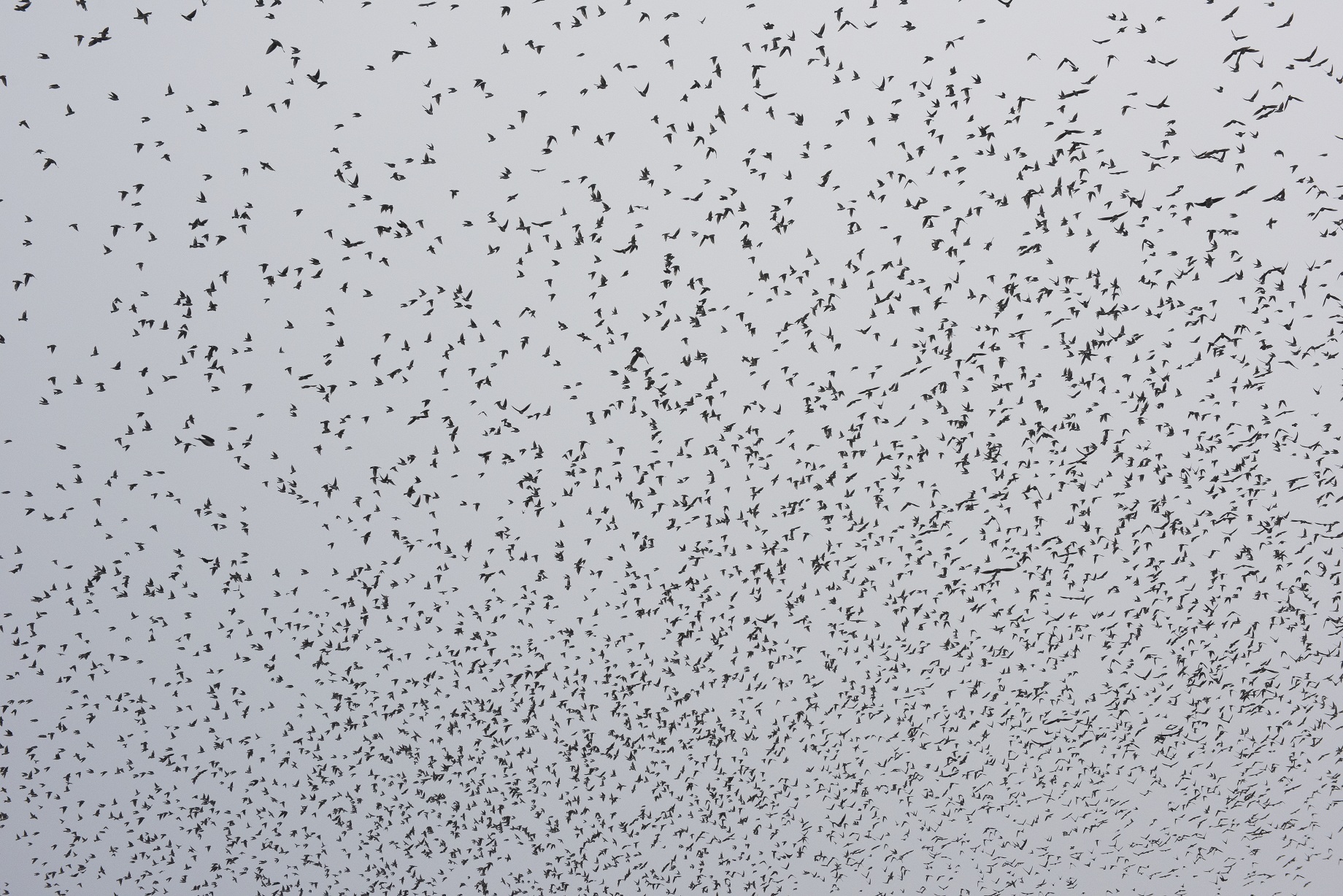}
    \includegraphics[width=0.49\textwidth]{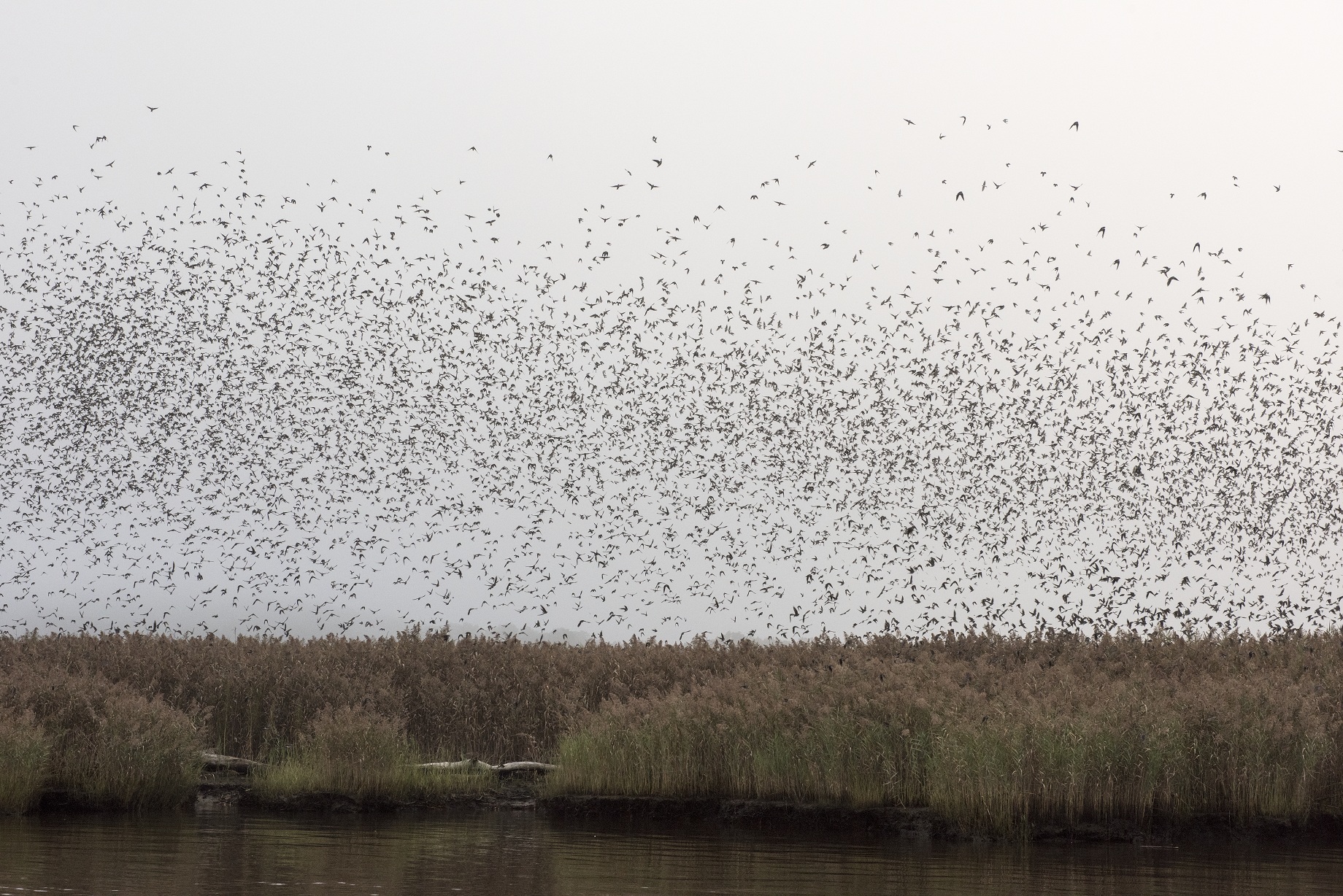}
    \caption{The Sky image (left) and Reeds image (right) used for our detection experiments. Note the resolution has been reduced for this visualization.}
    \label{fig:sky-reeds-images}
\end{figure}

 We visualize the Reeds and Sky images in figure \ref{fig:sky-reeds-images}. To collect ground truth, we manually label both images entirely. We use a tile size of 200 pixels for sky and 160 pixels for reeds, resulting in a total of 925 and 1426 tiles respectively. To label we randomly group the tiles into batches of 50, and have labelers annotate each batch using VGG annotator. To speed up the process, we label each bird with a single point at its center of mass. This results in about 1 second of human effort per labeled point. We only label a bird on the edge if a majority of it is visible in the tile. In total, our ground truth count for sky is 5682 birds and 12486 birds for reeds. 

To convert point annotations into approximate bounding boxes, we use a two-stage heuristic algorithm. First, we apply Otsu thresholding to each tile and compute the average size of a square bounding box by dividing the number of foreground pixels by the number of labeled points. In the second stage, we center these boxes at each labeled point, mask out foreground pixels outside the boxes, and recompute the average box size using the remaining foreground pixels.

\begin{figure}
    \centering
    \includegraphics[width=0.45\textwidth]{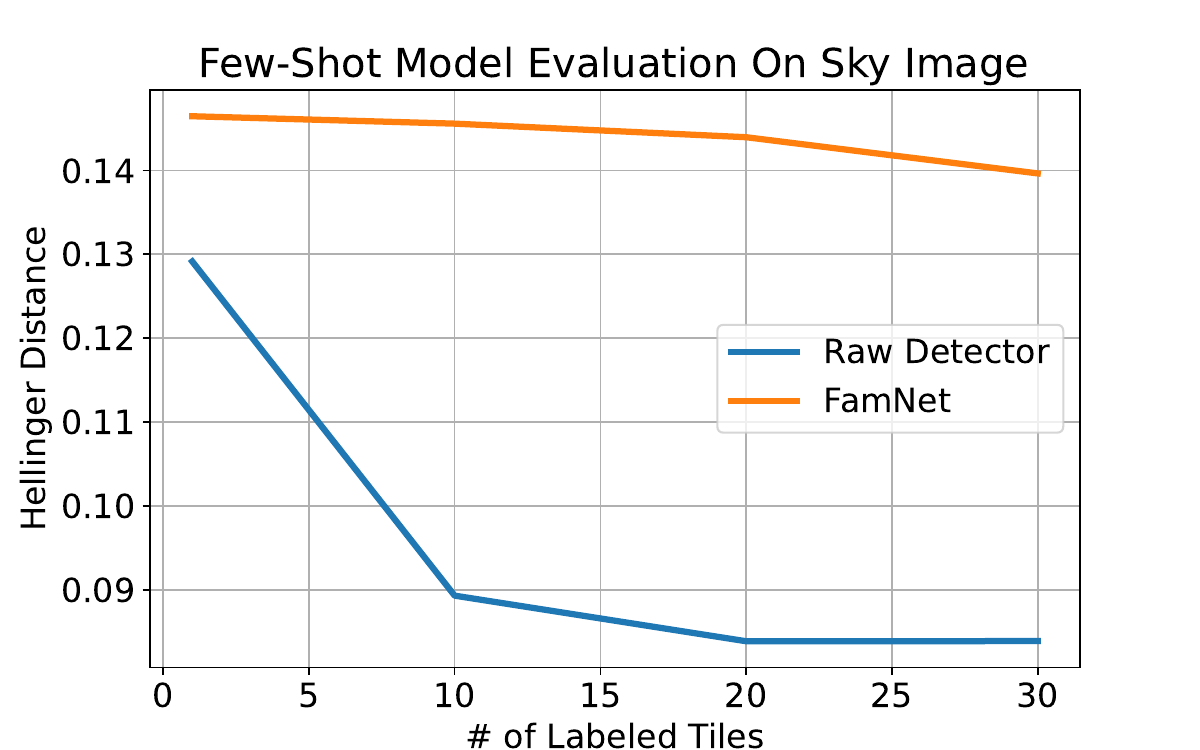}
    \includegraphics[width=0.45\textwidth]{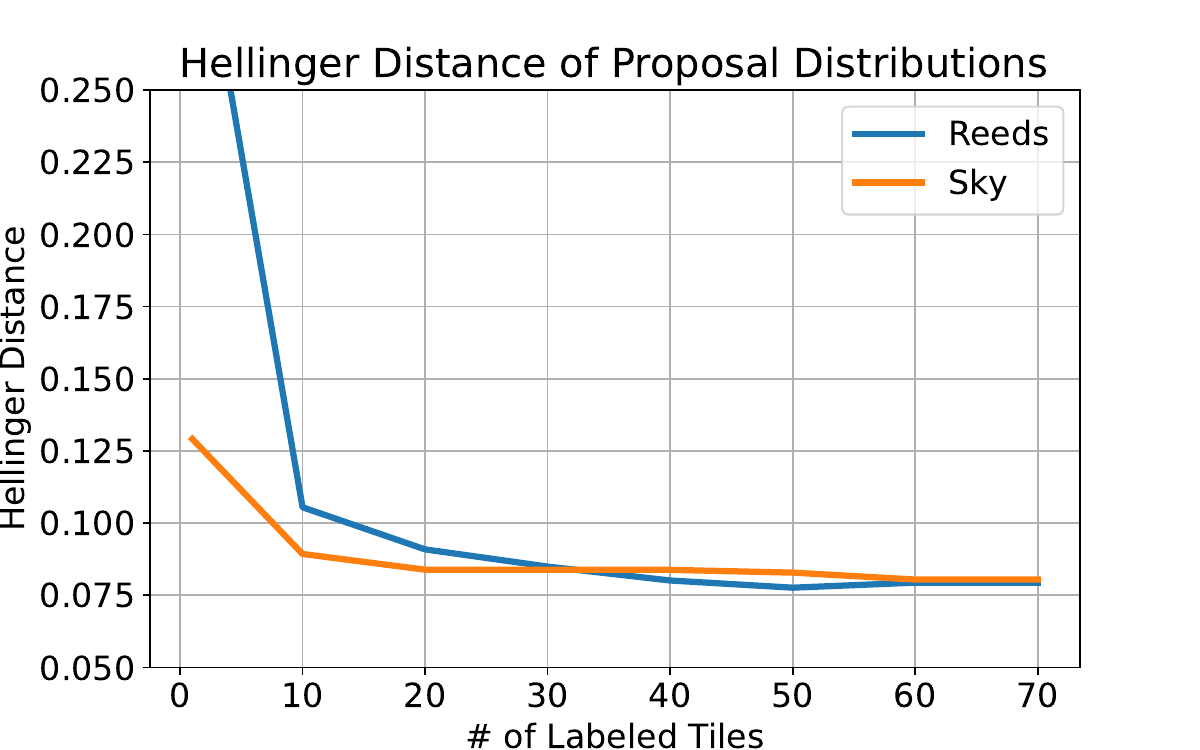} 
    \caption{Left: Comparing the performance of raw predictions from fine-tuned detectors vs. few-shot FamNet\cite{ranjan2021learning} model, based on Hellinger distance to ground truth. The FamNet model performs worse and does not improve as much with more labeled tiles.  Right: Hellinger distance between the raw predictions of the fine-tuned detector and the optimal sampling distribution for both images, averaged over 100 checkpoints. Unlike the raw counts which saturate quickly, additional training continues to improve the proposal distribution, especially on the harder Reeds image.}
    \label{fig:few_shot}
\end{figure}
\begin{figure}[t]
    \centering
    \includegraphics[width=1.00\linewidth]{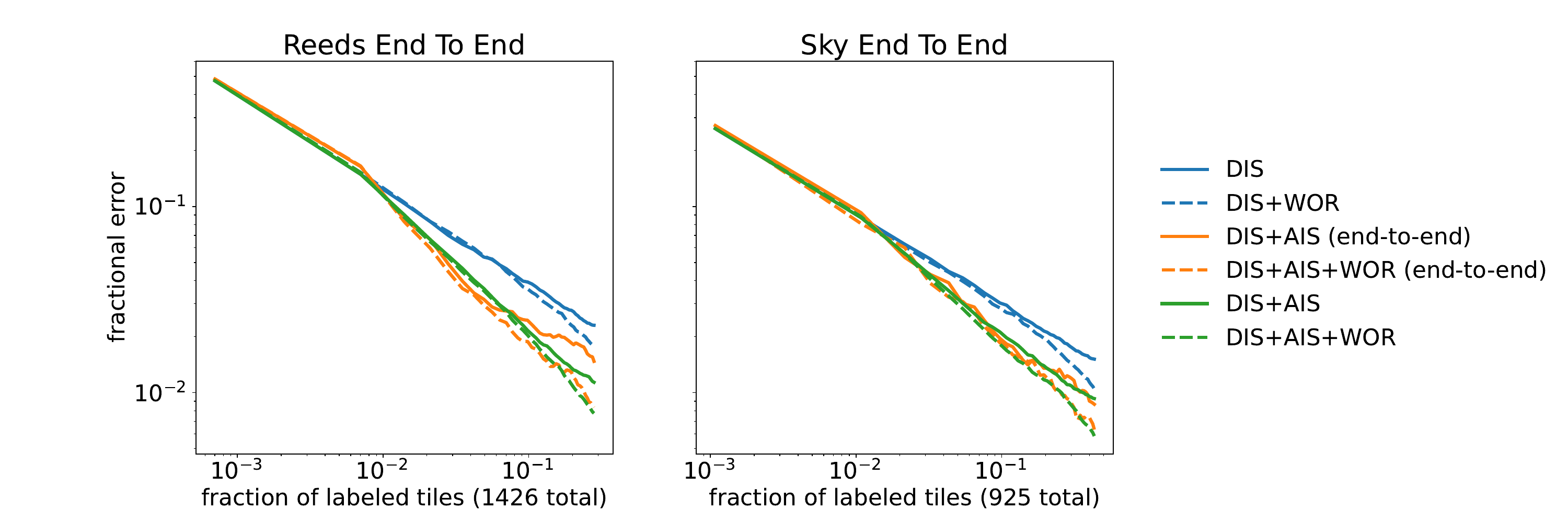}
    \caption{Comparing our fixed checkpoint scheme against true end-to-end training, averaged over 1000 trials. While the fixed checkpoint approach introduces slightly higher variance, particularly at higher label counts, it preserves relative trends between methods. This supports its use for our large-scale evaluation.}
    \label{fig:end_to_end}
\end{figure}

\section{Additional experimental results}
\subsection{Comparison to few-shot models} \label{sec:fsmodels}
Figure~\ref{fig:few_shot} presents results in terms of the Hellinger distance between the predicted proposal distribution $P$ and the optimal sampling distribution $Q$ (proportional to ground truth counts). We use Hellinger distance because it accommodates zero-probability entries. It is defined as:
\begin{align}
H(P,Q) = \frac{1}{\sqrt{2}}\sqrt{\sum_{i=1}^k(\sqrt{p_i}-\sqrt{q_i})^2}.\notag
\end{align}
Where $p_i$, $q_i$ are the probabilities assigned to unit $i$ for $P$ and $Q$, respectively. Unlike raw counts, the Hellinger distance is unaffected by constant-factor over or under prediction and thus gives a better idea of how good the model is for the estimation process.

On the left, we compare the proposal distributions of the sky image raw predictions from our fine-tuned detector and the few-shot FamNet model \cite{ranjan2021learning}, which estimates object counts from a few examples without additional training. FamNet struggles to adapt with more examples, while the fine-tuned model achieves lower Hellinger distance even with a single labeled tile.

\subsection{Fine-Tuning Improves Proposal Distribution}
\label{sec:hellingerdistance}
On the right of Figure~\ref{fig:few_shot}, we track the Hellinger distance over the course of finetuning, averaged over 100 checkpoints. While raw detector counts tend to saturate early, the Hellinger distance of the proposal distribution continues to improve, especially for the harder Reeds image, before eventually saturating.

\subsection{True End-to-End Finetuning} 
\label{sec:endtoend}
To approximate the effect of interactive model adaptation in our main results, we use a “fixed checkpoint” approach: a predefined sequence of detectors trained on a progressively larger number of labels. This allows us to conduct a significantly larger number of trials than would be feasible with full end-to-end training.

In practice, however, a practitioner would retrain the detector as new samples are labeled during active measurement. To assess how well our “fixed checkpoint” approach approximates this more realistic scenario, we compare it to end-to-end training averaged over 1,000 trials (Figure~\ref{fig:end_to_end}).

The results indicate that performance under both approaches is similar, particularly when using fewer tiles, and that the relative trends between methods are preserved. We do observe slightly higher variance in the fixed checkpoint setting, likely due to the specific checkpoint chosen.

\subsection{Different weighting schemes for all experiments}
We compare different weighting schemes in both image counting and radar counting experiments. The results are in Fig. \ref{fig:weighting_all} and \ref{fig:weighting_station}. From the first and the third columns, we conclude that the $\alpha_{\mathrm{COMB}}$ weights work consistently better than $\alpha_{\mathrm{SQRT}}$ and $\alpha_{\mathrm{LURE}}$. In the second and the fourth columns, the performances of inverse variance weighting with different $\gamma$ are compared. In general, a conservative $\gamma = 0.3$ brings little benefit, while an aggressive $\gamma = 0.9$ can be detrimental because each individual estimator may not be accurate enough. 
\renewcommand{\height}{0.22\linewidth}
\begin{figure}
    \centering
    \includegraphics[height=\height]{figure/weighting/sky_weight1.pdf}
    \includegraphics[height=\height]{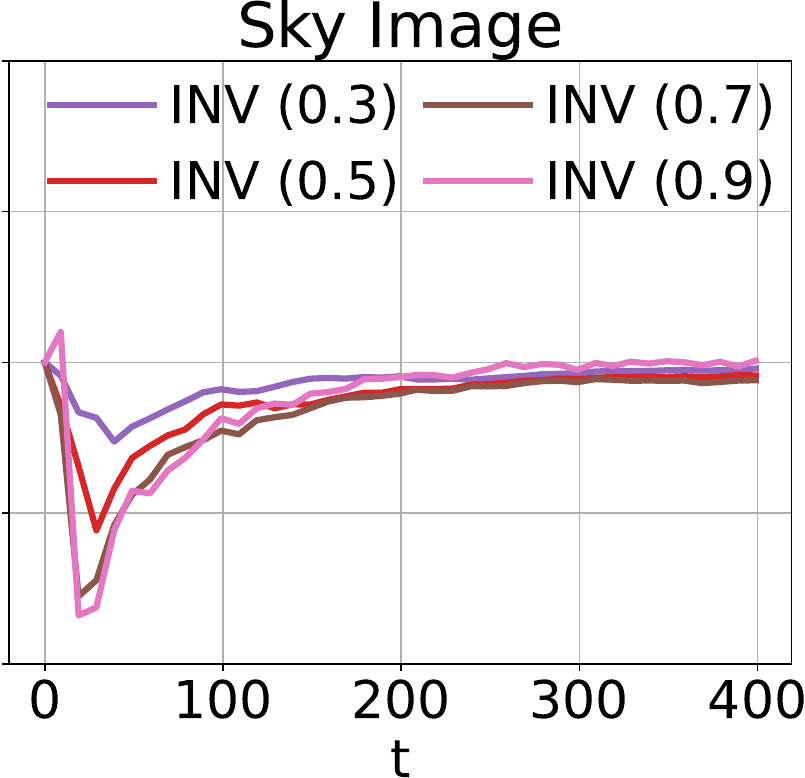} \hfill
    \includegraphics[height=\height]{figure/weighting/reeds_weight1.pdf}
    \includegraphics[height=\height]{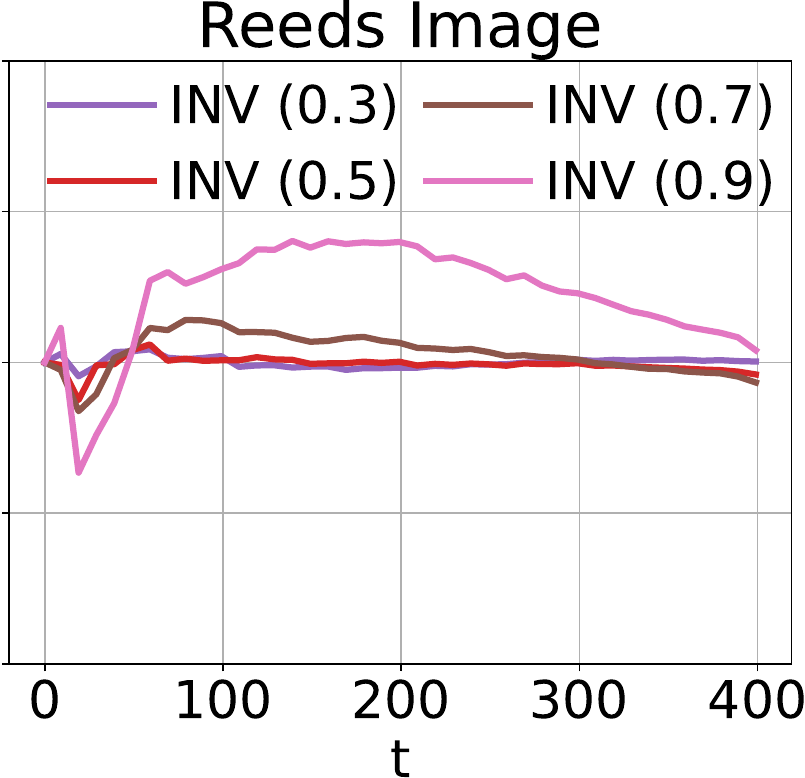}
                \vspace{-0.05in}
    \caption{Relative errors compared to $\alpha^{\mathrm{COMB}}$ weighting. Other fixed weighting strategies ($\alpha^{\mathrm{SQRT}}$, $\alpha^{\mathrm{LURE}}$) are worse, but inverse variance weighting (denoted by INV ($\gamma$)) may achieve lower error. 
    }
    \label{fig:weighting_all}
\end{figure}
\begin{figure}[t]
    \centering    
    \includegraphics[height=\height]{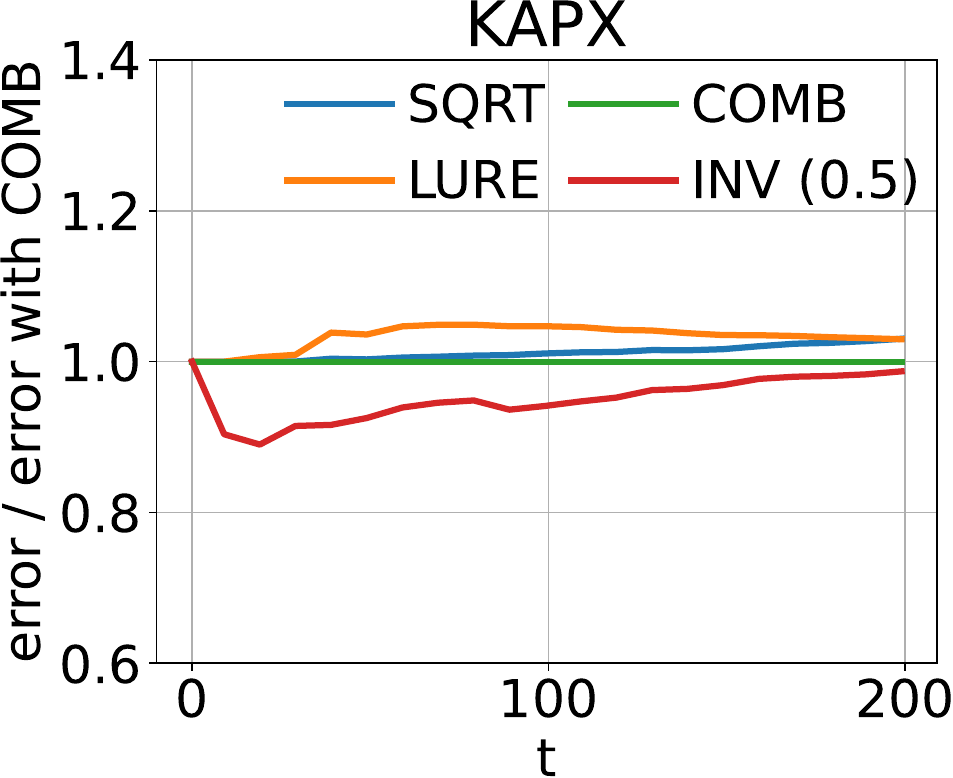} \hfill
    \includegraphics[height=\height]{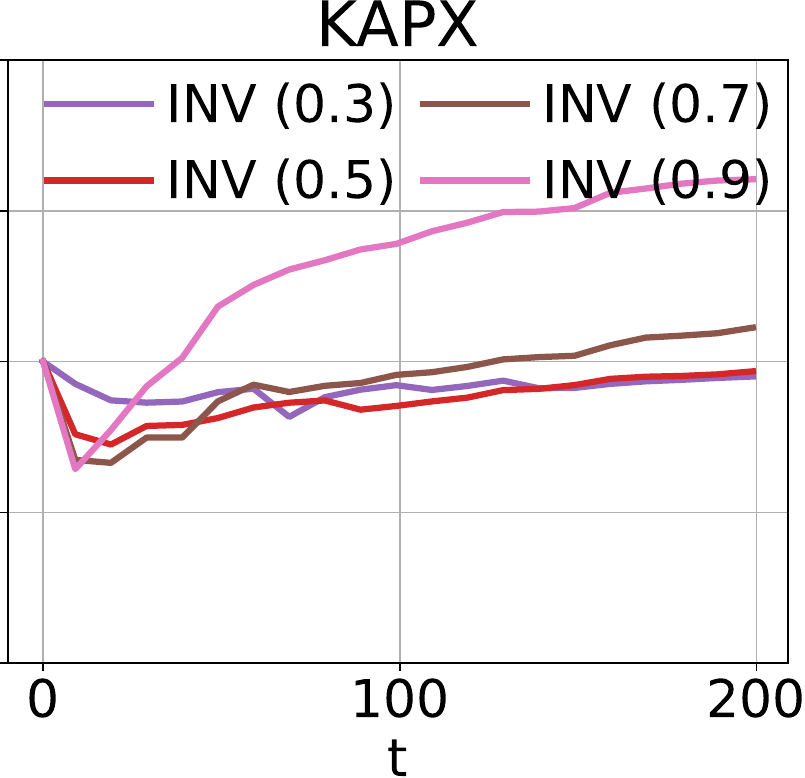} \hfill
    \includegraphics[height=\height]{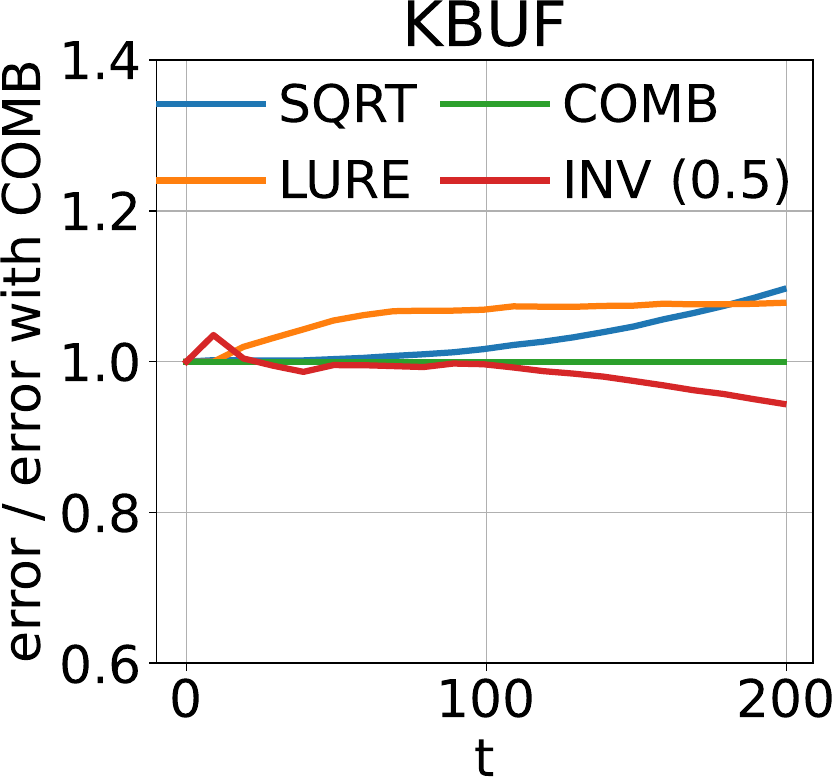} \hfill
    \includegraphics[height=\height]{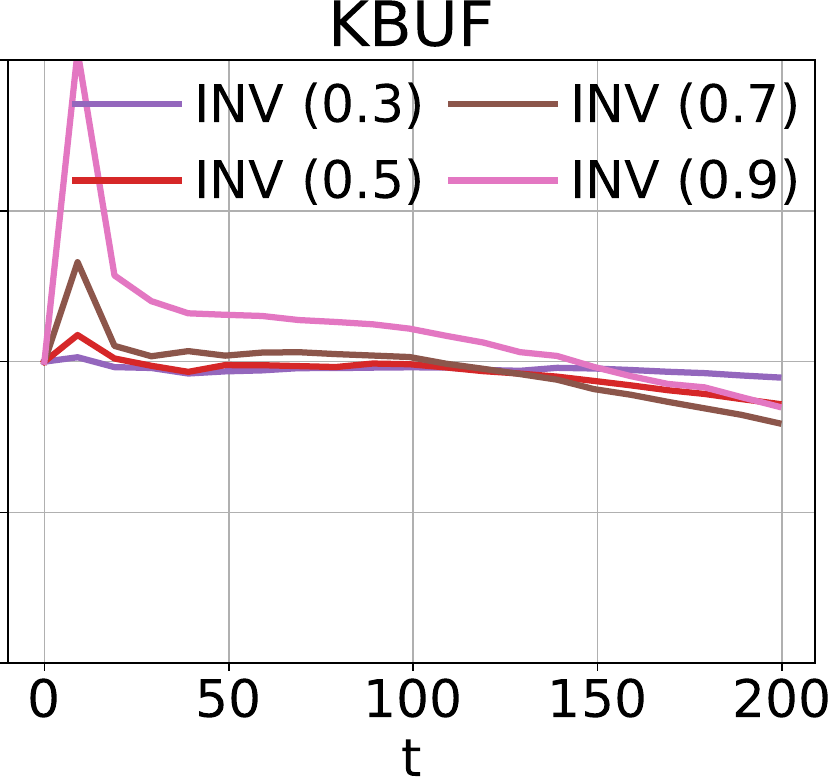}

        \includegraphics[height=\height]{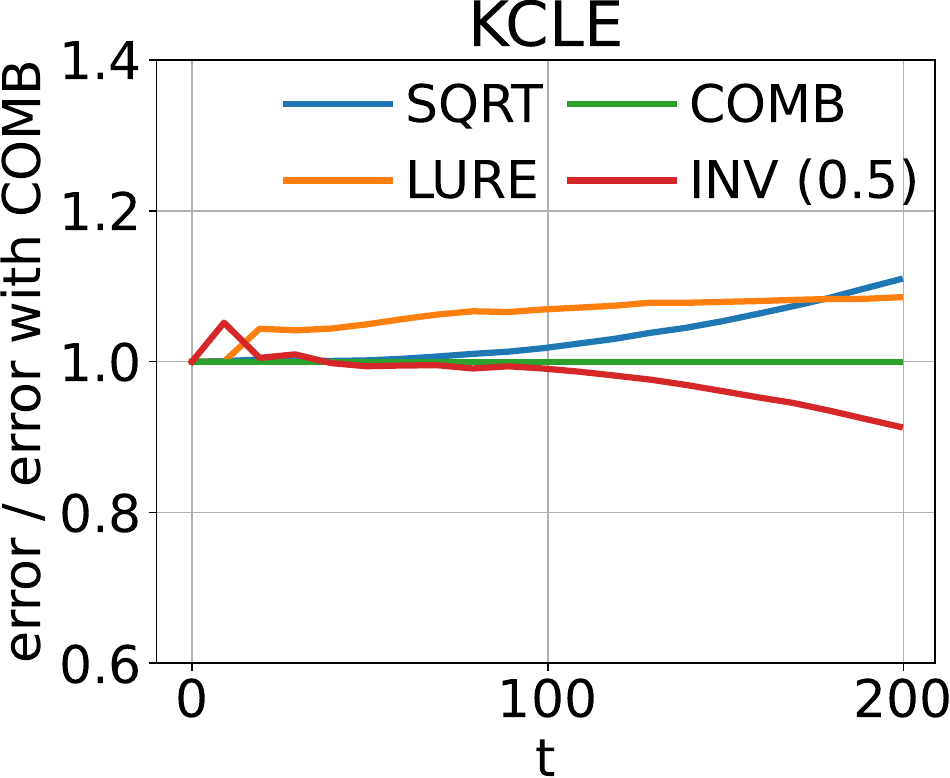} \hfill
    \includegraphics[height=\height]{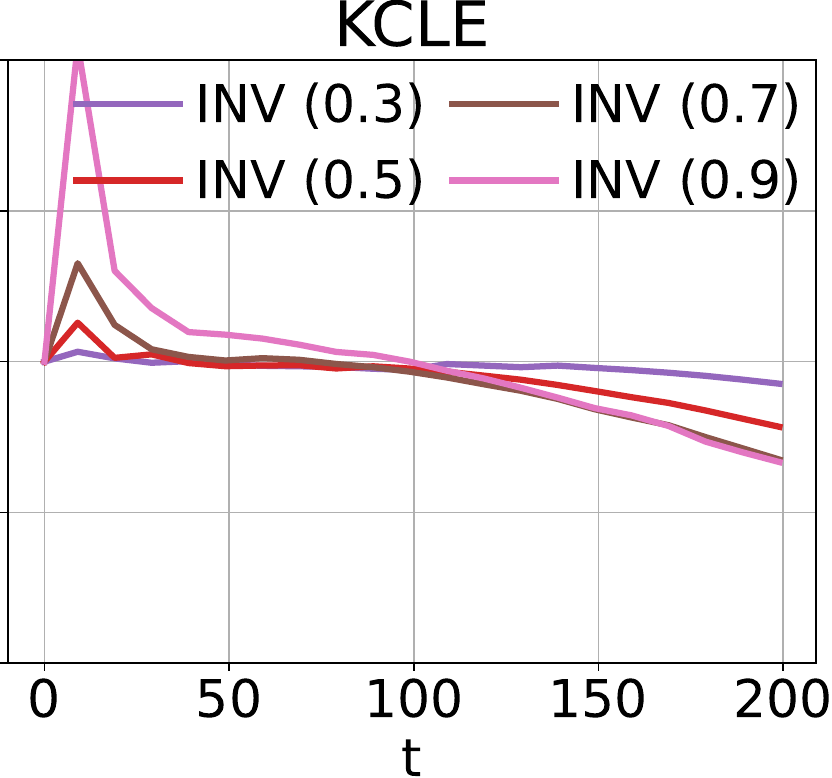} \hfill
    \includegraphics[height=\height]{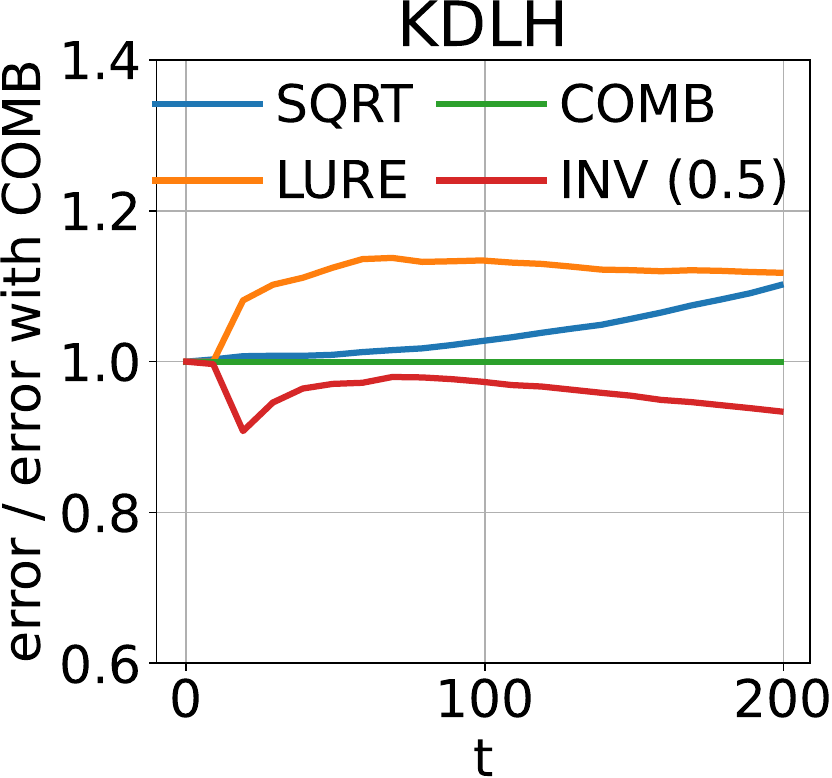} \hfill
    \includegraphics[height=\height]{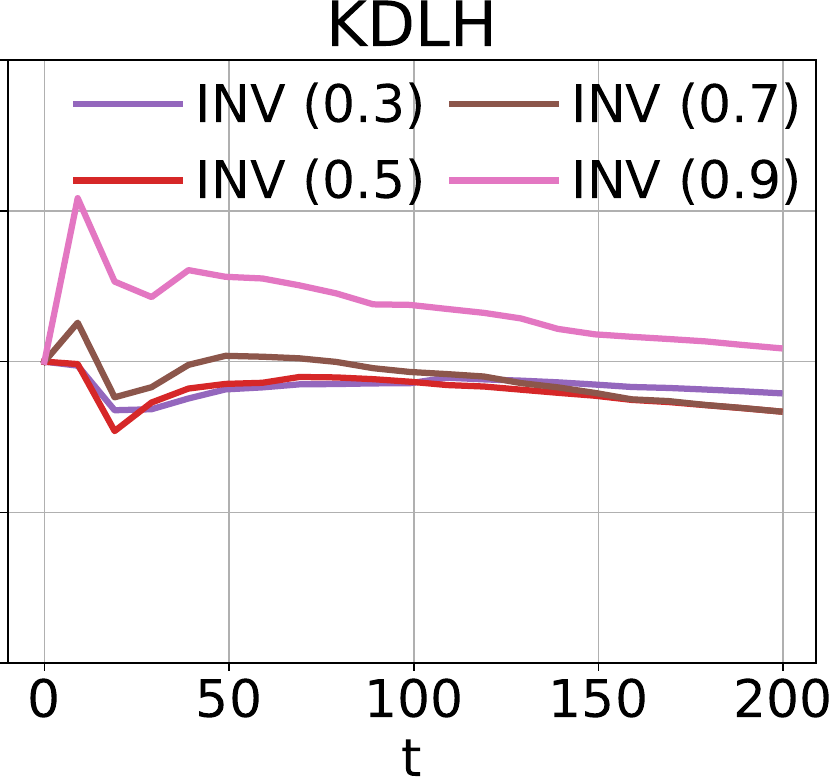}

        \includegraphics[height=\height]{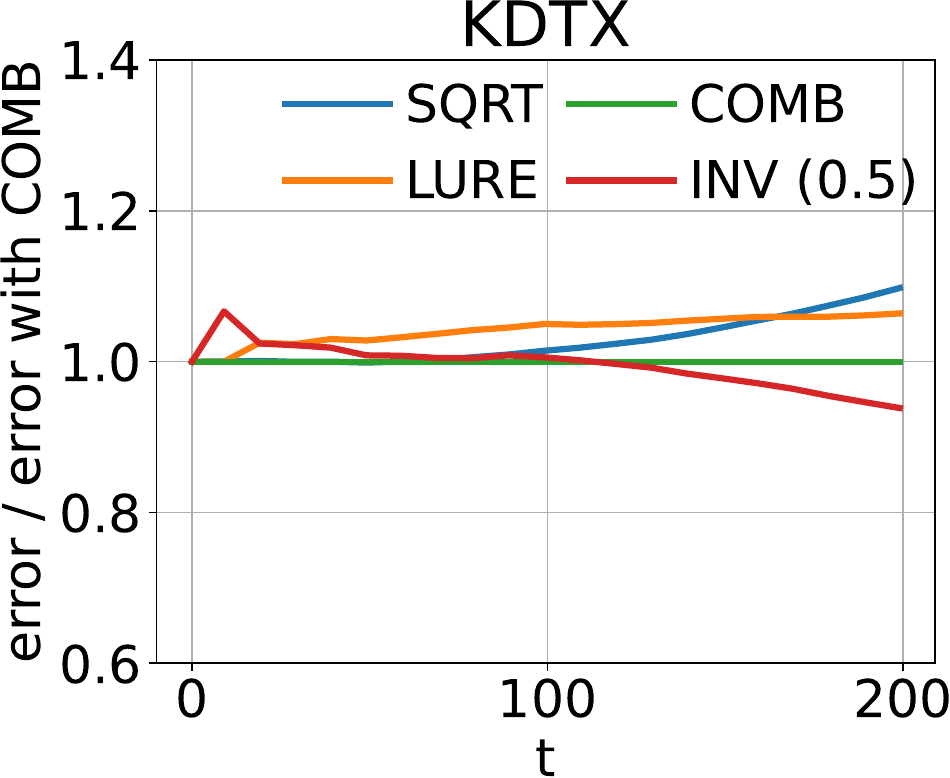} \hfill
    \includegraphics[height=\height]{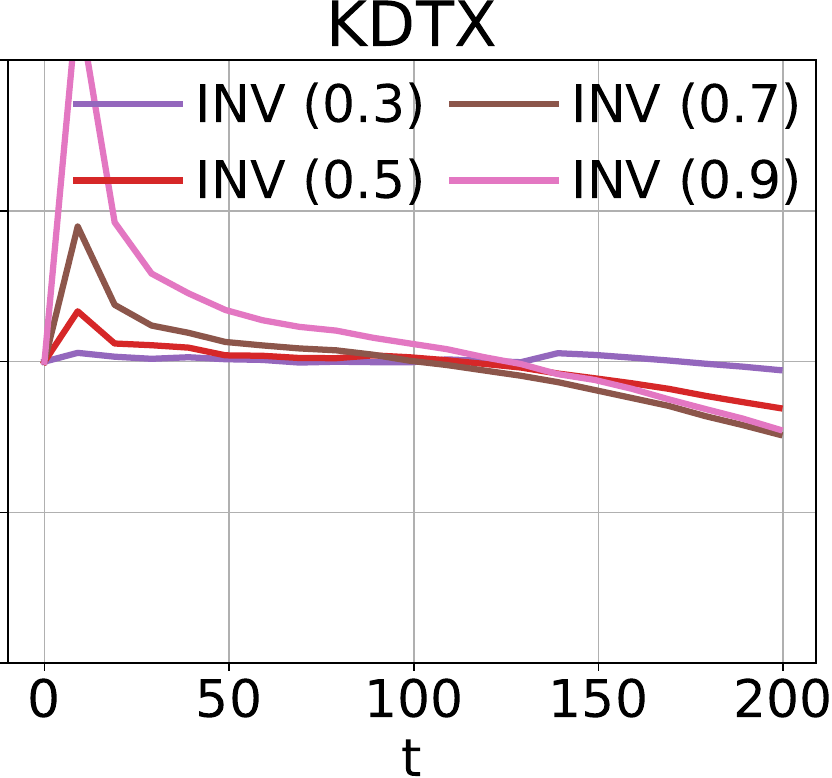} \hfill
    \includegraphics[height=\height]{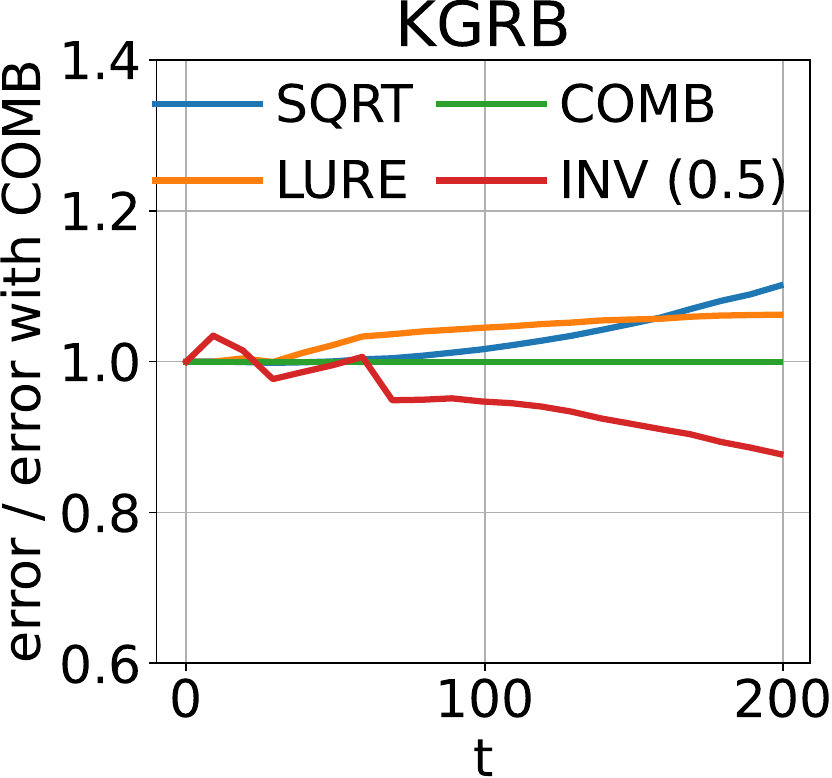} \hfill
    \includegraphics[height=\height]{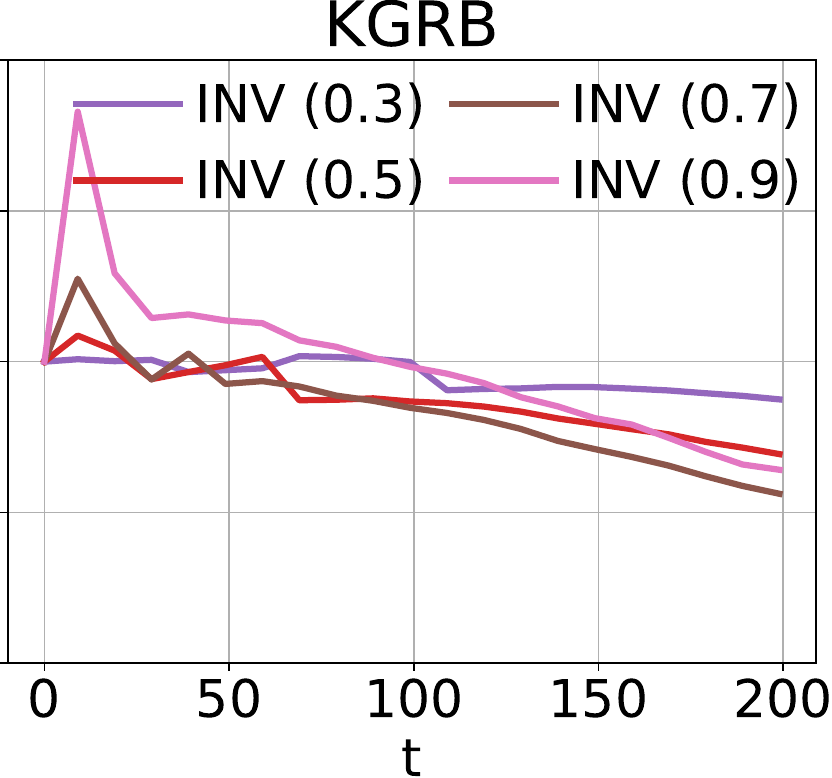}

        \includegraphics[height=\height]{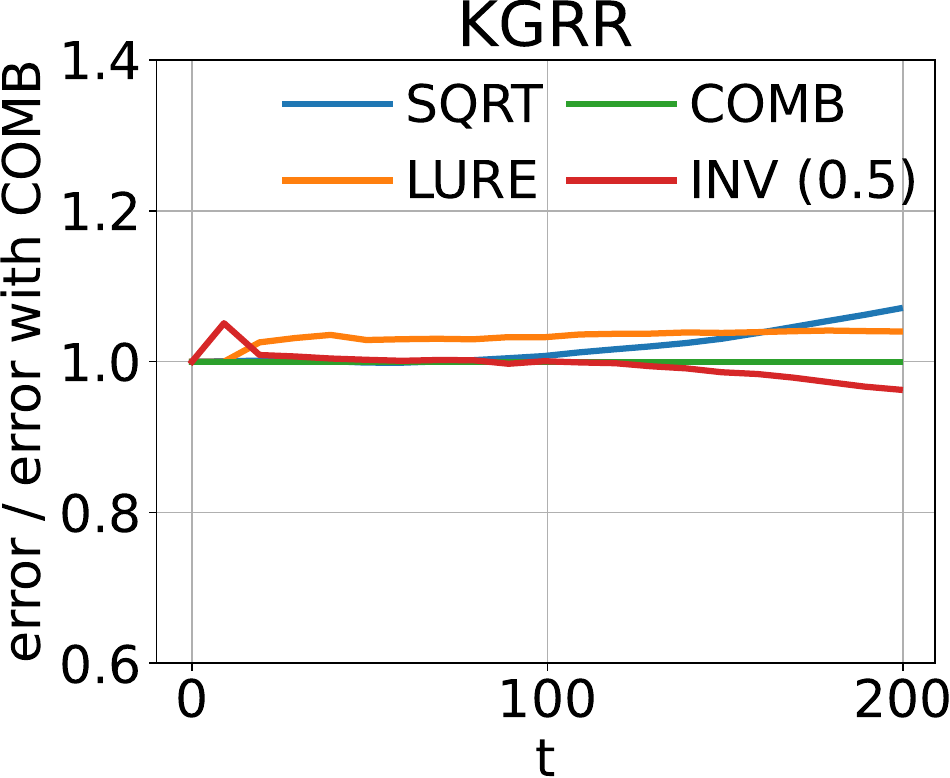} \hfill
    \includegraphics[height=\height]{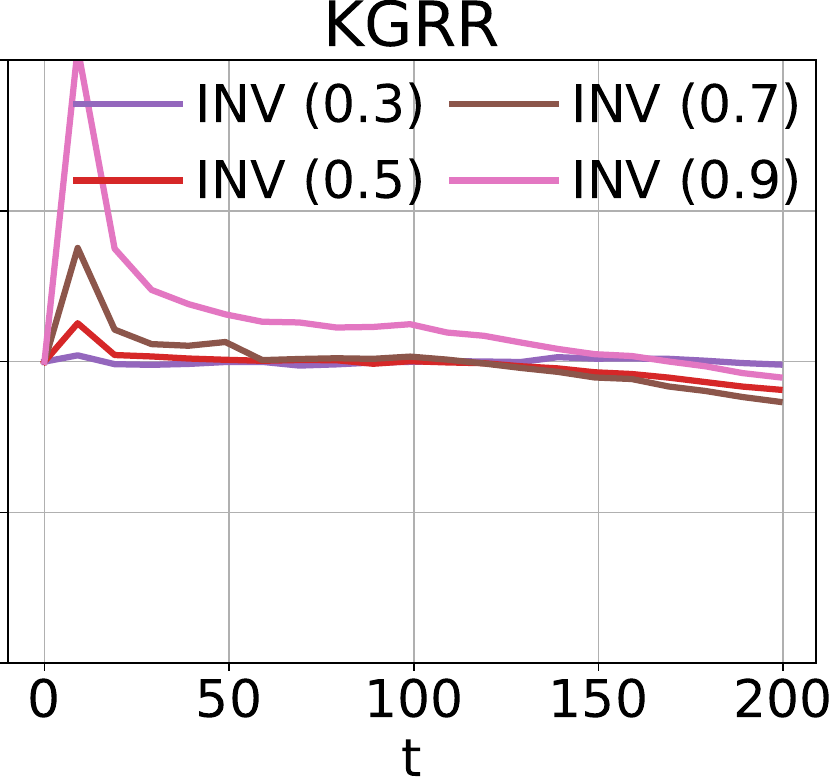} \hfill
    \includegraphics[height=\height]{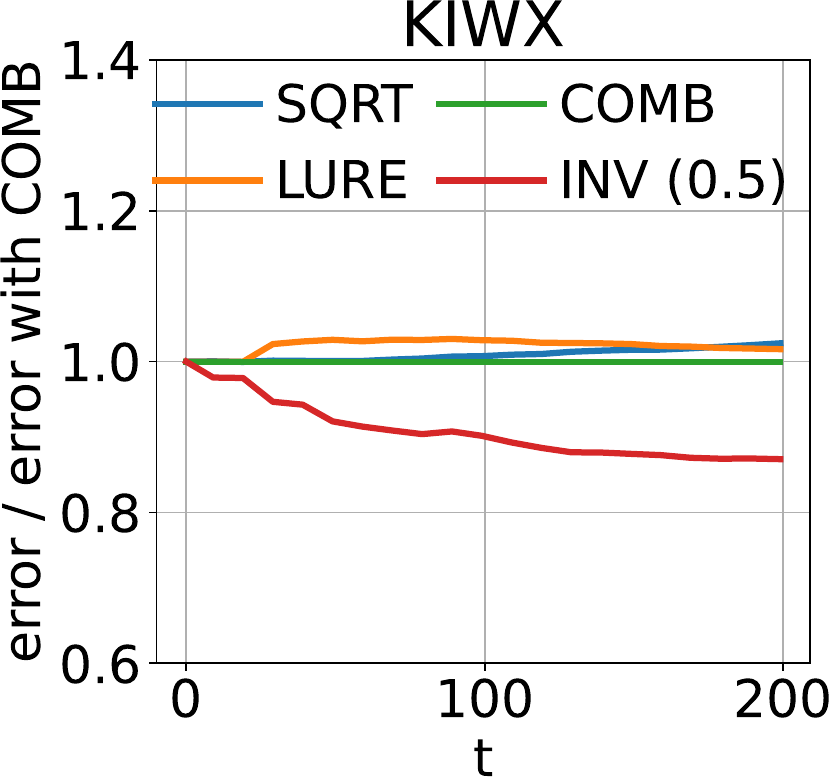} \hfill
    \includegraphics[height=\height]{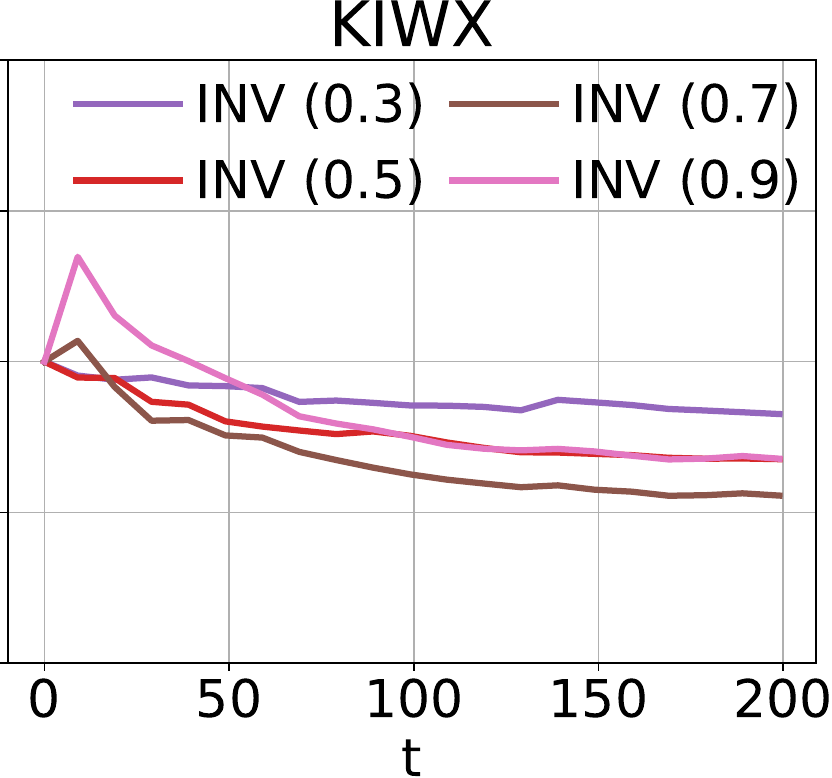}
    \includegraphics[height=\height]{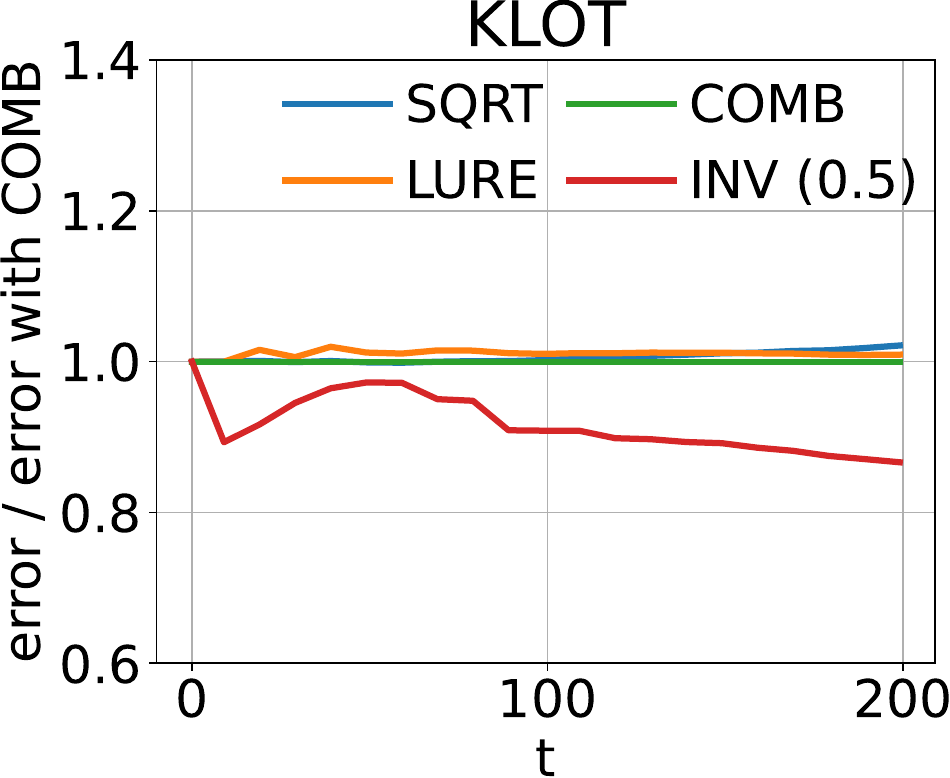} \hfill
    \includegraphics[height=\height]{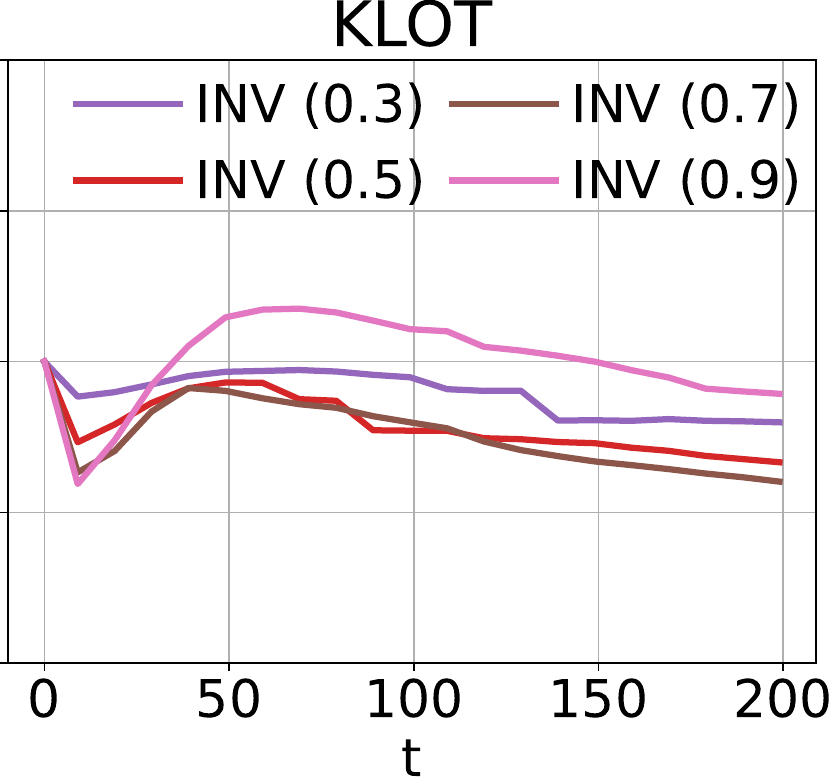}
    \hfill
    \includegraphics[height=\height]{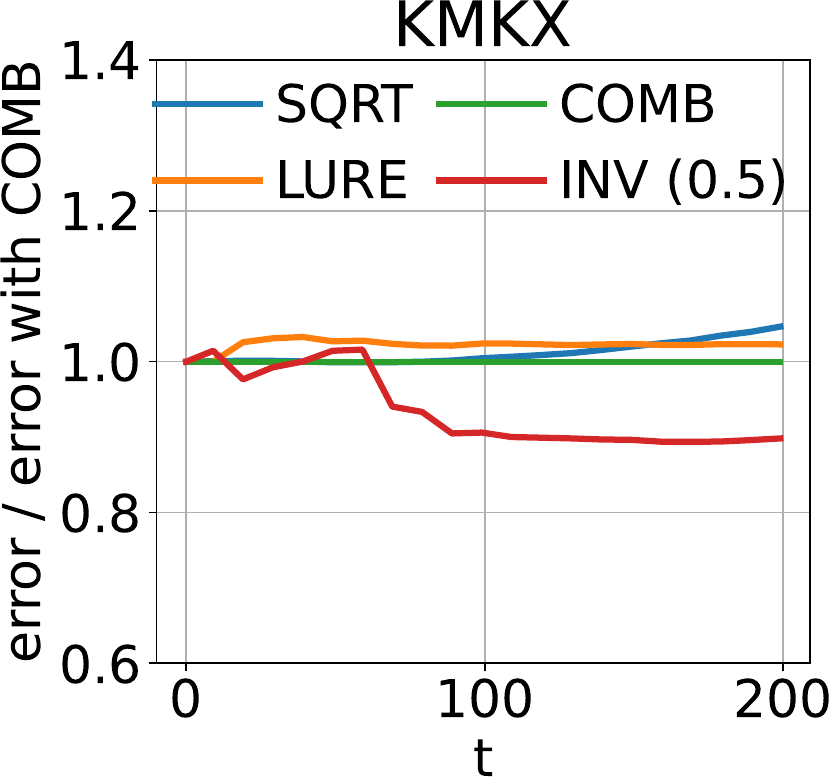} \hfill
    \includegraphics[height=\height]{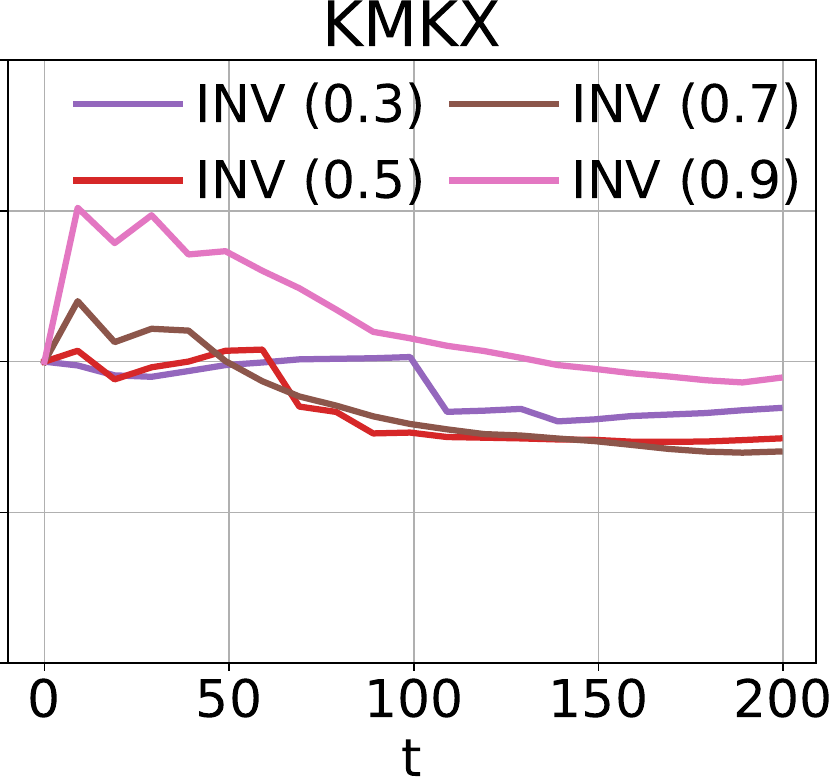}
    \raggedright
    \includegraphics[height=\height]{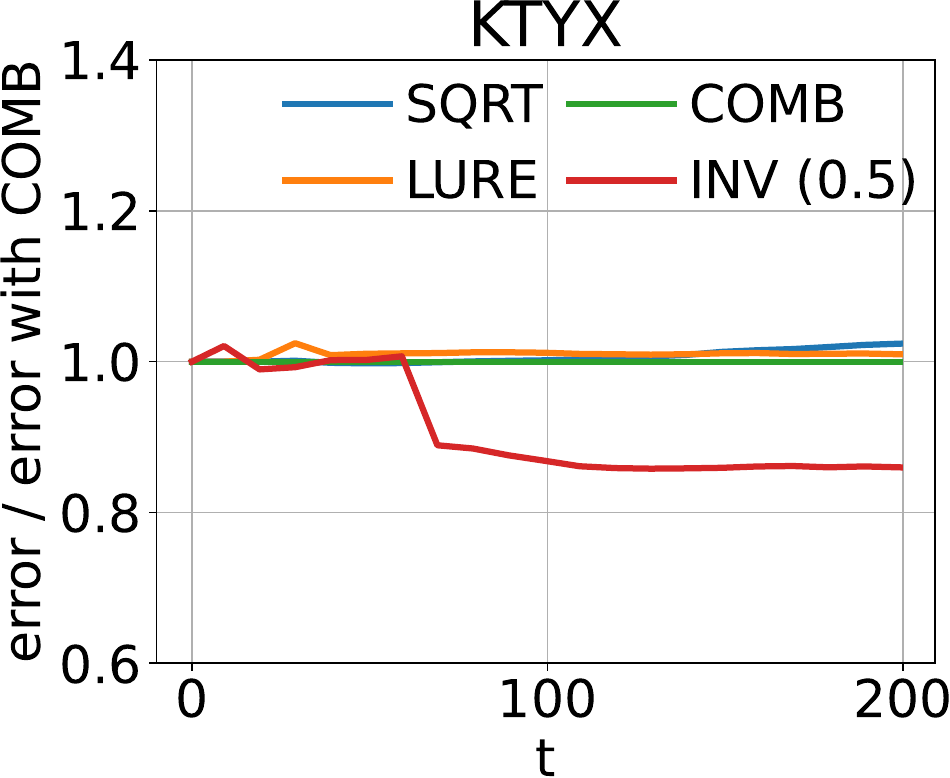}
    \hspace{1pt}
    \includegraphics[height=\height]{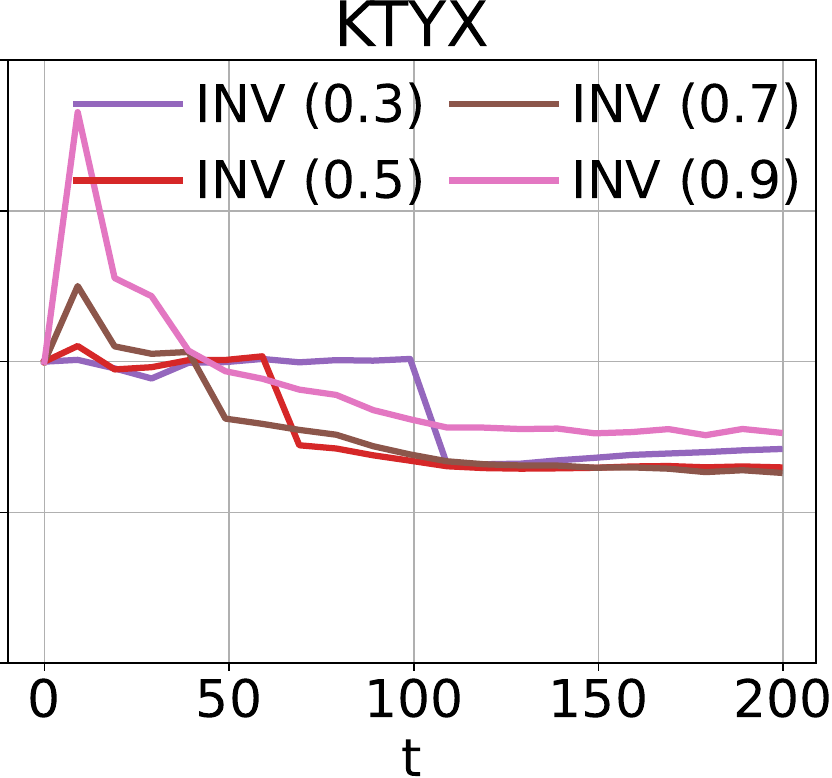}
    \caption{Relative errors compared to $\alpha^{\mathrm{COMB}}$ weighting, for the roost counting problem on all 11 radar stations.
    }
    \label{fig:weighting_station}
\end{figure}
\subsection{Confidence intervals for all experiments}
\begin{figure}
    \centering
        \includegraphics[width=0.75\linewidth]{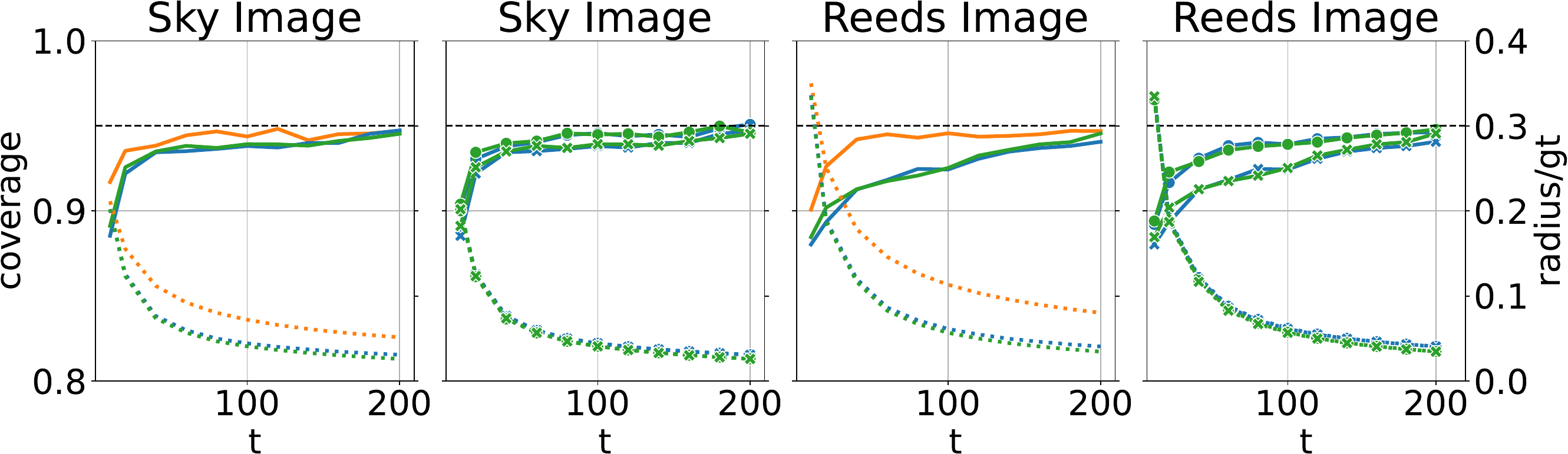}
    \includegraphics[width=0.75\linewidth]{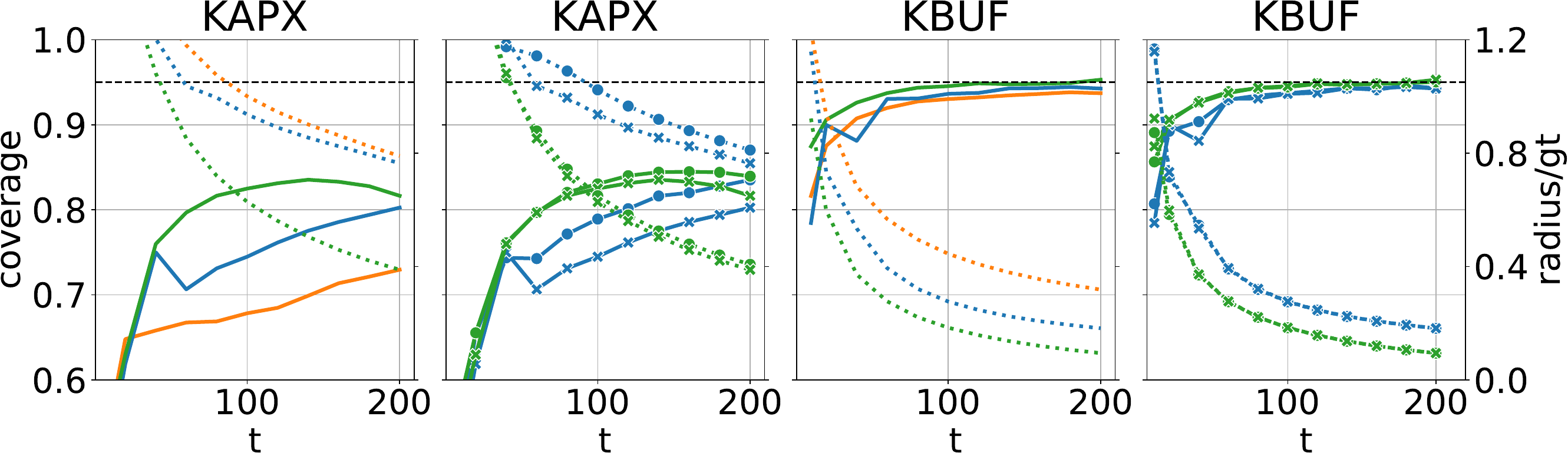}
        \includegraphics[width=0.75\linewidth]{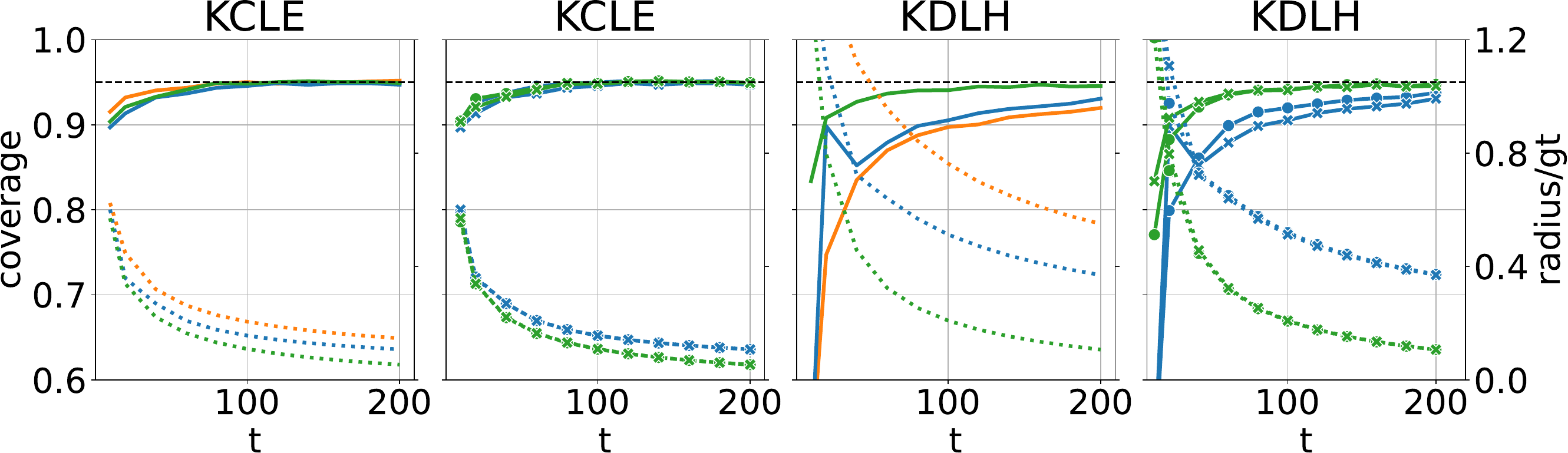}          
                \includegraphics[width=0.75\linewidth]{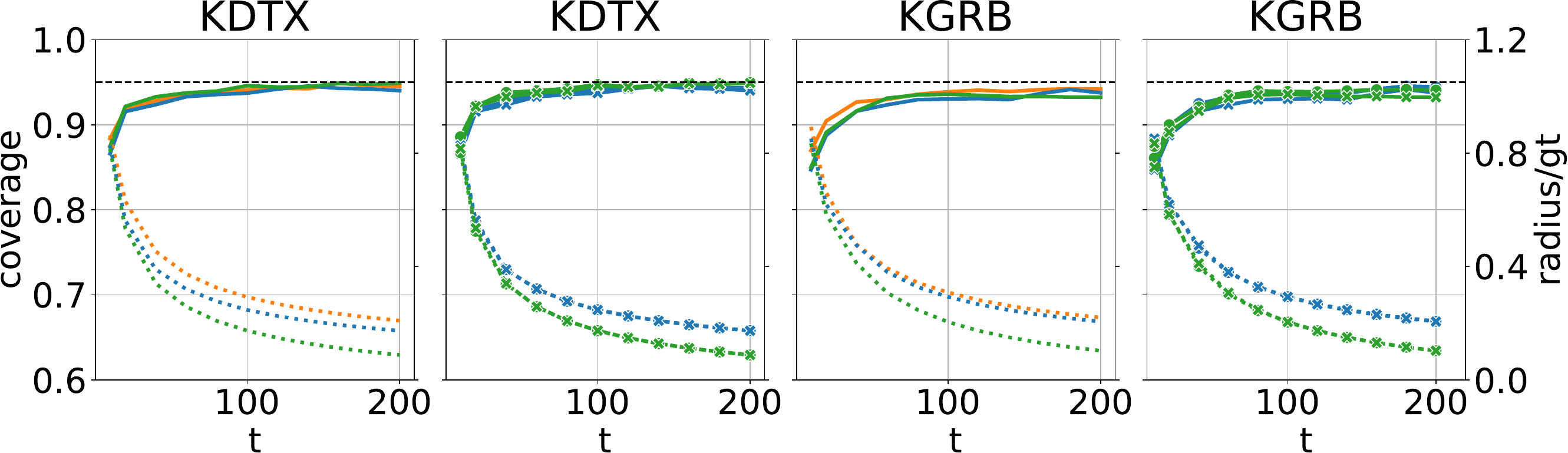}  
                        \includegraphics[width=0.75\linewidth]{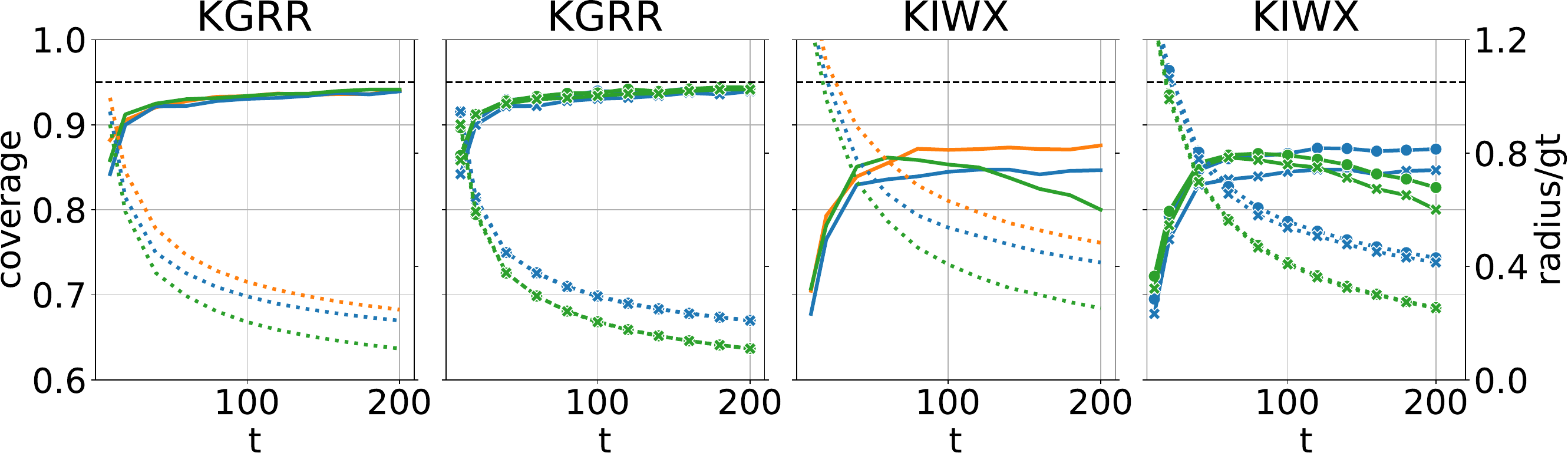} 
                                \includegraphics[width=0.75\linewidth]{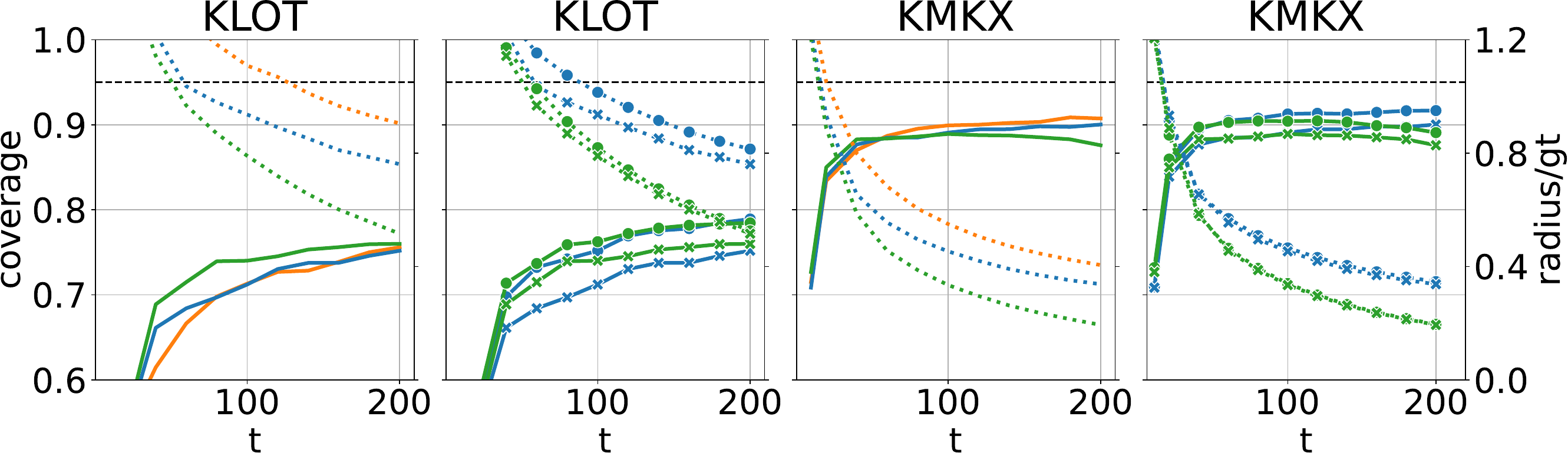} 
                    
        \hspace{-60pts} \includegraphics[width=0.61\linewidth]{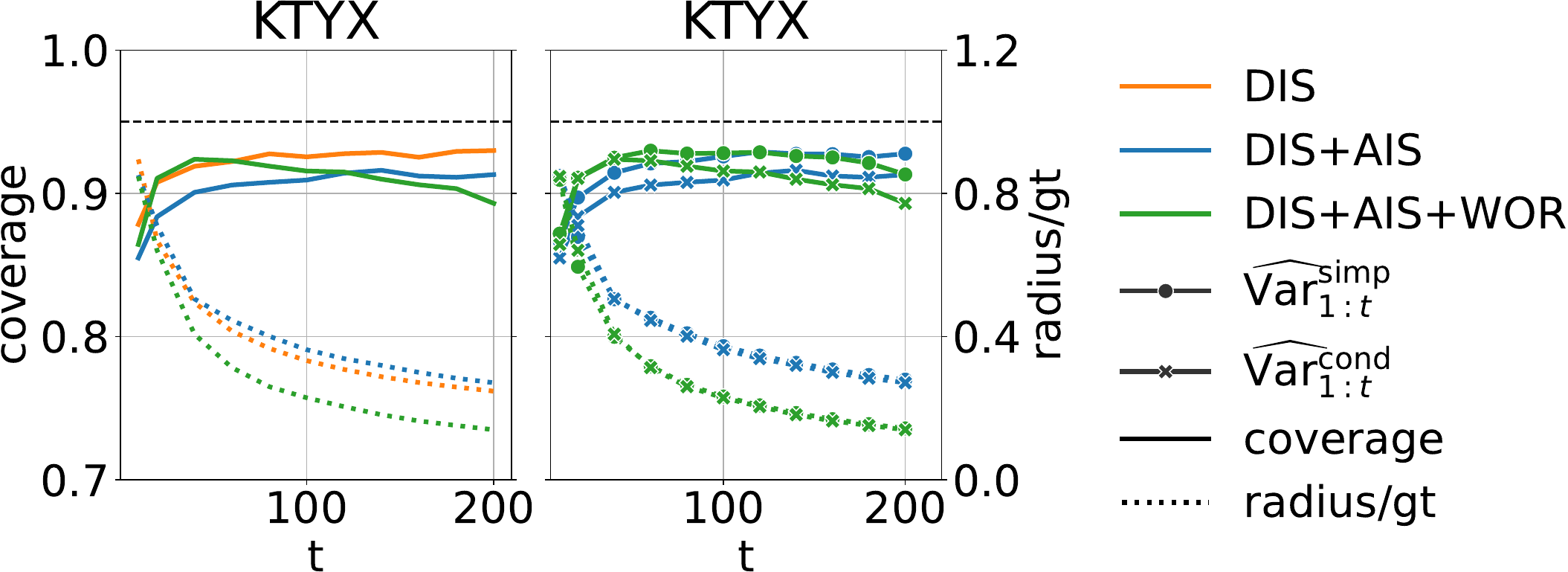}
    \caption{Full CI coverage results for counting on the two images and on all 11 radar stations.}
    \label{fig:coverage_station_all}
\end{figure}

In Fig. \ref{fig:coverage_station_all}, we compare the CI coverages for all experiments. In most of the experiments, the coverage for active measurement improves as the number of labeled samples increases (except on KIWX and KTYX). This behavior is typical in importance sampling-based methods. Out of the 11 radar stations, we achieve near perfect coverage (around 0.95) in 6, good coverage (around 0.9) in 2 and fair coverage (around 0.8) in 3. Comparing the two variance estimators $\Var_{1:t}^{\mathrm{simp}}$ and $\Var_{1:t}^{\mathrm{cond}}$, we find that the $\Var_{1:t}^{\mathrm{simp}}$ works better, especially on stations with bad coverage. We conclude that the CIs in active measurement usually have the desired properties for measuring the uncertainty. 
\end{document}